%% file: main.tex
\documentclass[10pt,twocolumn,letterpaper]{article}

\usepackage[pagenumbers]{iccv}
\input{preamble}

\usepackage{multirow}
\usepackage{colortbl}
\usepackage{version}
\usepackage{amsmath}
\usepackage{array}
\usepackage{xcolor}
\usepackage{graphicx}
\usepackage{cuted} 
\usepackage{float}
\graphicspath{{./}{img/}}

\newcommand{\GG}{\mathcal{G}}
\newcommand{\SSS}{\mathcal{S}}

\definecolor{lightgray}{gray}{0.9}
\definecolor{darkgray}{gray}{0.8}

\definecolor{iccvblue}{rgb}{0.21,0.49,0.74}
\usepackage[pagebackref,breaklinks,colorlinks,allcolors=iccvblue]{hyperref}

\def\paperID{8965} 
\def\confName{ICCV}
\def\confYear{2025}

\title{Superpixel Anything: A general object-based framework \\ for accurate yet regular superpixel segmentation}

\author{Julien Walther$^{\;1,\;2}$ \and Rémi Giraud$^{\;1}$ \and Michaël Clément$^{\;2}$ \\
\and $^{1\;}$Univ. Bordeaux, CNRS, Bordeaux INP, IMS, UMR 5218, France \\
${^2\;}$Univ. Bordeaux, CNRS, Bordeaux INP, LaBRI, UMR 5800, France}

\begin{document}

\maketitle

\begin{abstract}
Superpixels are widely used in 
computer vision
to simplify
image representation 
and reduce
computational complexity.
While traditional methods 
rely on low-level features,
deep learning-based approaches leverage high-level features 
but also tend to sacrifice regularity of superpixels to capture complex objects, leading to accurate but less interpretable segmentations.
In this work, we introduce SPAM (SuperPixel Anything Model), 
a versatile framework for segmenting images into accurate yet regular superpixels.
We train a model to extract image features for superpixel generation, and at inference, we leverage a large-scale pre-trained model for semantic-agnostic segmentation to 
ensure that superpixels align with 
object masks.
SPAM can handle any prior high-level segmentation, resolving uncertainty regions, 
and is able to interactively focus on specific objects. 
Comprehensive experiments demonstrate that SPAM qualitatively and quantitatively outperforms state-of-the-art methods on segmentation tasks,
making it a  
valuable
and robust tool for various applications.
Code and pre-trained models are available here:
\url{https://github.com/waldo-j/spam}.
\end{abstract}


\section{Introduction}
\label{sec:intro}

Superpixels are a relevant approach in many image processing and computer vision pipelines, primarily due to their ability to reduce computational complexity and provide an accurate under-representation for interactive or annotation purposes.
They are applied to various image types, each with its own specific characteristics, e.g.,
natural 2D images~\cite{achanta2012,xu2024learning}, 
3D medical images~\cite{tian2017supervoxel}, 
videos~\cite{chang2013video,wang2021supervoxel},
or hyperspectral images~\cite{massoudifar2014superpixel,zhao2021superpixel},
and are used in many applications such as 
stereo matching~\cite{li2016pmsc,yang2020superpixel},
optical flow estimation~\cite{liu2019selflow},
semantic segmentation \cite{kim2023adaptive,mei2025spformer},
object detection \cite{lee2022spsn},
representation learning using transformers \cite{ke2023learning},
implicit neural representation \cite{li2024superpixel} 
or used with neural radiance fields \cite{chen2023structnerf}.

From~\cite{ren2003} to the widely known SLIC algorithm~\cite{achanta2012}, superpixels define a form of image segmentation constrained by specific criteria, where
regions must have approximately the same size, be clearly identifiable and fit to the image objects.
Superpixels can be modeled using 
neighborhood ~\cite{sarkar2021non}
or graph-based methods~\cite{gould2008} and offer an interesting under-representation, especially for interactive tasks~\cite{long2018efficient,berg2019ilastik}.

However, as for general image segmentation, the problem is often considered ill-posed due to the implicit nature of its constraints, making it challenging to assess the interest of different approaches~\cite{giraud2024_tip}. 
For instance, some traditional, e.g.,~\cite{felzenszwalb2004,Uziel:ICCV:2019:BASS,zhou2023vine},
and deep learning-based methods~\cite{jampani2018superpixel,tu2018learning,wang2021ainet},
consistently relax the shape regularity constraint of superpixels to achieve higher accuracy.
This raises the questions: 
is accuracy more important than regularity
and
is it possible to achieve high segmentation accuracy
without generating irregular, noisy or almost non-connected segmentation?

Historically, superpixel clustering algorithms adhere to an implicit regular grid structure and rely on 
 computationally efficient low-level features \cite{achanta2012}.
Nevertheless, applying these methods to natural images poses challenges, as object boundaries 
are not always strongly correlated with color features,
and some objects may be thin or have irregular shapes.
The performance is thus limited by the regularity constraint and the reliance on low-level features.

With the rise of deep learning, 
and the large quantity of available semantic annotated images,
numerous methods now leverage such data to extract high-level features for local pixel clustering, e.g.,~\cite{jampani2018superpixel,wang2021ainet,peng2022hers}.
These high-level features, often derived from encoder architectures such as CNNs,
are trained on segmentation tasks using manually annotated data without semantic information, to cluster homogeneous pixels in any context.
Such deep features enable to follow object boundaries more accurately.

Nevertheless, most deep learning-based methods, {e.g}~\cite{tu2018learning,yang2020superpixel,wang2021ainet,xu2024learning} do not properly enforce the shape regularity of the superpixels, 
in order to increase segmentation accuracy, especially on stretched or thin image objects \cite{giraud2024_tip}.
Regularity is inherently part of the definition of superpixel decomposition, and constitutes an essential criterion for a fair comparison between algorithms. 
Moreover, regularity can largely benefit superpixel-based
graph-based structures \cite{gould2008},
superpatch construction \cite{giraud2017_spm}, 
video tracking \cite{chang2013video},
and interactive methods \cite{Uziel:ICCV:2019:BASS,kang2020dynamic,zhou2023vine}.
 Finally, even with very irregular superpixels, state-of-the-art
performance begins to plateau on natural image datasets \cite{giraud2024_tip}.
Recently, approaches like SAM~\cite{kirillov23sam} offered
significant progress in large-scale semantic-agnostic training, being able to 
segment almost any object with a certain semantic level.   

However, these methods only generate independent object mask proposals,
that may overlap, lack precise borders with potentially missing subparts.

This highlights the remaining need for complementary lower-level superpixel segmentation
that can be used to refine high-level proposals or as input of any superpixel-based pipeline.
Hence, we address how to efficiently leverage
high-level object proposals to reach unprecedented levels of superpixel accuracy.%

\medskip

\noindent\textbf{Contributions.}
In this work, we introduce the SuperPixel Anything Model (SPAM), a general superpixel segmentation framework leveraging large-scale pre-trained models such as SAM~\cite{kirillov23sam}.
Our main contributions are listed as follows: 
(i) Our method is the first to handle any prior segmentation, possibly containing uncertainty regions, ensuring that the 
superpixels are strictly contained within the mask objects.
Notably, we show how to efficiently use prior segmentations from 
SAM~\cite{kirillov23sam}.
(ii)  Thanks to the possibility of using high-level prior segmentation, we propose two new adaptive inference modes, based on  visual attention~\cite{caron2021emerging} or user interactions, to adjust superpixel scale within objects.
This provides a hierarchical decomposition, that can lighten the processing of superpixels or facilitate annotations.
(iii) We present an in-depth ablation study and extensive comparisons on segmentation tasks.
(iv) SPAM provides the most visually comprehensible superpixels, and significantly outperforms the accuracy of state-of-the-art approaches.


\section{Related work}

\noindent\textbf{Low-level feature-based methods.}
Superpixel methods gained popularity with 
SLIC 
\cite{achanta2012}, a simple yet efficient local $K$-means framework to cluster pixels 
in a combined low-level spatial and color space. 
Following methods use
augmented feature spaces~\cite{chen2017}, 
manifold learning 
\cite{liu2016manifold}, 
contour constraints 
\cite{zhang2016}, 
or a non-iterative approach~\cite{achanta2017superpixels}.
Beyond these, methods like 
Watershed~\cite{machairas2015},
coarse-to-fine~\cite{yao2015}
or Gaussian Mixture-based method~\cite{Ban18} 
represent interesting alternatives.
Hierarchical approaches, e.g.,~\cite{wei2018} have also been proposed to segment the image at different 
scales, using similar clustering features as~\cite{achanta2012}.
Finally, to address the lack of structure in the segmentation,
a lattice superpixel method has been proposed~\cite{pan2022fast}, preserving a 8-neighborhood connectivity, though this comes at the expense of segmentation accuracy. 
Recent approaches have moved away from the 
pseudo-grid layout constraint.
For instance, in~\cite{Uziel:ICCV:2019:BASS,kang2020dynamic,xu2022high,zhou2023vine}
methods 
generate fewer (so larger) or more regular superpixels in low-variance
areas, increasing performance for a given number 
of requested superpixels.

\begin{figure*}[ht]
\centering
\includegraphics[width=0.975\textwidth, height=0.35\textwidth]{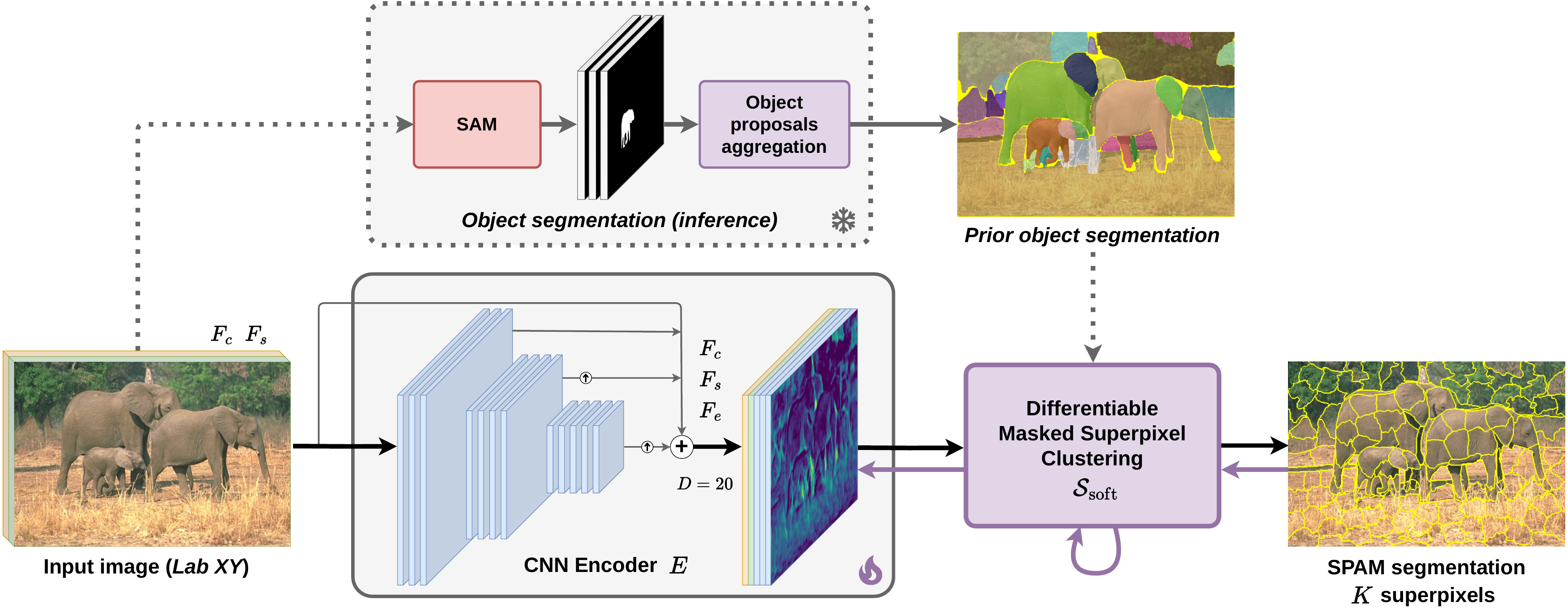}
\caption{\textbf{SPAM superpixel segmentation pipeline.}
A CNN encoder is trained to generate deep superpixel features via a differentiable clustering algorithm.
At test time, this clustering algorithm can be constrained by a prior object segmentation, coming from a pre-trained large-scale segmentation model such as SAM~\cite{kirillov23sam}.
Since this prior object segmentation may present overlapping
objects, a processing step is applied to convert it into a proper image partition.  
The final deep learning-based superpixels are therefore, by construction, strictly contained within the objects of the prior segmentation.
See text for details.
}
\label{fig:spam_archi}
\end{figure*}

\noindent\textbf{Deep learning-based methods.}
In recent years, deep learning-based superpixel methods have emerged 
to address the limitations of low-level feature-based approaches.
The Superpixel Sampling Network (SSN)~\cite{jampani2018superpixel} introduces
a convolutional neural network to compute high-level features.
SSN is trained using a differentiable version of SLIC, 
performing a local $K$-means clustering on the extracted features, and is optimized with both segmentation and shape compactness losses.
Several methods have built on this approach by, e.g.,
introducing a contour loss to enhance boundary adherence~\cite{tu2018learning},
using an encoder-decoder network to
learn the association between pixels and superpixels~\cite{yang2020superpixel},
or improving~\cite{yang2020superpixel}
by adding to the pixel features, the information of neighboring grid superpixel embeddings~\cite{wang2021ainet}. 
Note that all these methods are trained on 
moderate-sized
dataset like BSD~\cite{martin2001} (200 train images),
and report marginal improvements when trained or fine-tuned on
larger ones. 
Other approaches like~\cite{zhu2021learning} 
try to increase the generalizability by projecting the
features into a cluster-friendly space, 
or propose a hierarchical segmentation~\cite{peng2022hers},
both at the expense of
lower segmentation accuracy. 
Deep learning-based methods have focused heavily on segmentation performance or computational efficiency
and offer limited control over regularity, often requiring to retrain the network, while producing irregular regions
(see results in Sec.~\ref{subsec:results}). 
Nevertheless, the overall segmentation performance 
begins to plateau on natural image datasets.

\noindent\textbf{Evaluation framework.}
Superpixel methods may be evaluated on several properties
with dedicated metrics to assess the respect of the segmentation constraints. 
Key evaluation metrics include segmentation accuracy and contour detection performance relative to a ground truth, shape regularity, color homogeneity, or respect of the target number of superpixels.

However, superpixel evaluation is often as biased as the segmentation problem itself, given its ill-posed nature~\cite{giraud2024_tip}. 
For instance, boundary recall~\cite{martin2004} is frequently reported without precision, favoring noisier methods, or strongly correlated metrics are reported together. 
More importantly, many works omit superpixel regularity in their comparisons,
although it may have a significant impact on segmentation performance.
Therefore, many recent deep learning methods have often prioritized segmentation accuracy over regularity.
Studies such as~\cite{giraud2017_jei,stutz2018superpixels,giraud2024_tip} have highlighted these biases, recommending one metric per property for balanced evaluation.
In our experiments, we use these metrics alongside qualitative results.


\section{Superpixel Anything Model}
\label{sec:spam}

We introduce the proposed SuperPixel Anything Model (SPAM), 
a deep learning-based superpixel segmentation pipeline which can generate accurate and regular superpixels.
The key idea of SPAM is to allow the superpixel segmentation algorithm to be guided by a prior object segmentation.
This approach enables SPAM to leverage robust, large-scale pre-trained segmentation models, such as SAM~\cite{kirillov23sam}, or to utilize any user-provided prior segmentation.
The overall pipeline of SPAM is shown in Fig~\ref{fig:spam_archi}.

\subsection{Masked deep superpixel clustering}
\label{subsec:maskssn}

First, we introduce the main superpixel segmentation pipeline.
Then, we explain how to incorporate a prior object segmentation to constrain the superpixels to align with it.

\smallskip
\noindent\textbf{Deep superpixel clustering.}

We train a convolutional neural network (CNN) encoder $E$ to generate 
$D$ deep features which are then fed into a differentiable superpixel clustering algorithm 
to produce the output superpixel segmentations.
The core of this approach is based on the SSN method proposed by~\cite{jampani2018superpixel}.
The differentiable clustering algorithm takes as input image color features $F_c$ (\textit{Lab} channels), spatial features $F_s$ (\textit{xy} coordinates of each pixel), and the higher-level deep features $F_e$ from the CNN encoder $E$.

Starting from initial set of seed points (superpixel candidates),
the clustering algorithm iteratively computes a \textit{soft mapping} $\SSS_\mathrm{soft}$ of pixels to superpixels.
The procedure is similar to SLIC, but adapted to incorporate features $F_e$ from the encoder in a differentiable way.
Specifically, the clustering distance is balanced between different features with parameters $\lambda_c$ and $\lambda_s$ weighting color $F_c$ and spatial $F_s$ features, respectively.
These parameters are fixed during training, but can be modified at inference, 
allowing
to control superpixel regularity.

The model is trained in a supervised way using ground-truth 
high-level segmentations $\GG$.
The training loss is composed of two terms: 
(i) a pixel-wise categorical cross-entropy term $\mathcal{L}_\mathrm{seg}$,
comparing the predicted soft superpixels assignements $\SSS_\mathrm{soft}$ to ground-truth
$\GG$, and
(ii) a compactness term $\mathcal{L}_\mathrm{compact}$, penalizing superpixels with high spatial variance, such as: \vspace{-0.1cm}
\begin{equation}
    \mathcal{L} = \mathcal{L}_\mathrm{seg}(\GG,\SSS_\mathrm{soft}) + 
    \lambda \, \mathcal{L}_\mathrm{compact}(F_s,\SSS_\mathrm{soft}) .
    \label{eq:loss}
\end{equation}
More details on these losses are given in the supp. mat.

The clustering algorithm being differentiable, 
the model is end-to-end trainable so the encoder $E$ can learn deep pixel-wise features $F_e$ designed for superpixel clustering.
At inference time, a \textit{hard mapping} of every pixel to a unique superpixel is computed to generate the final segmentation. 
In principle, any deep learning architecture could be used as the encoder to train the model.
In this work, we use the same CNN encoder as in~\cite{jampani2018superpixel}.

This encoder provides a lightweight architecture, 
easily trainable, and ensures fair comparison with previous methods.
For more details about the architecture, 
we refer the reader to~\cite{jampani2018superpixel}.

\begin{figure}[t]
\centering
{\footnotesize
\begin{tabular}{@{\hspace{0mm}}c@{\hspace{1mm}}c@{\hspace{0mm}}}
\includegraphics[width=0.22\textwidth]{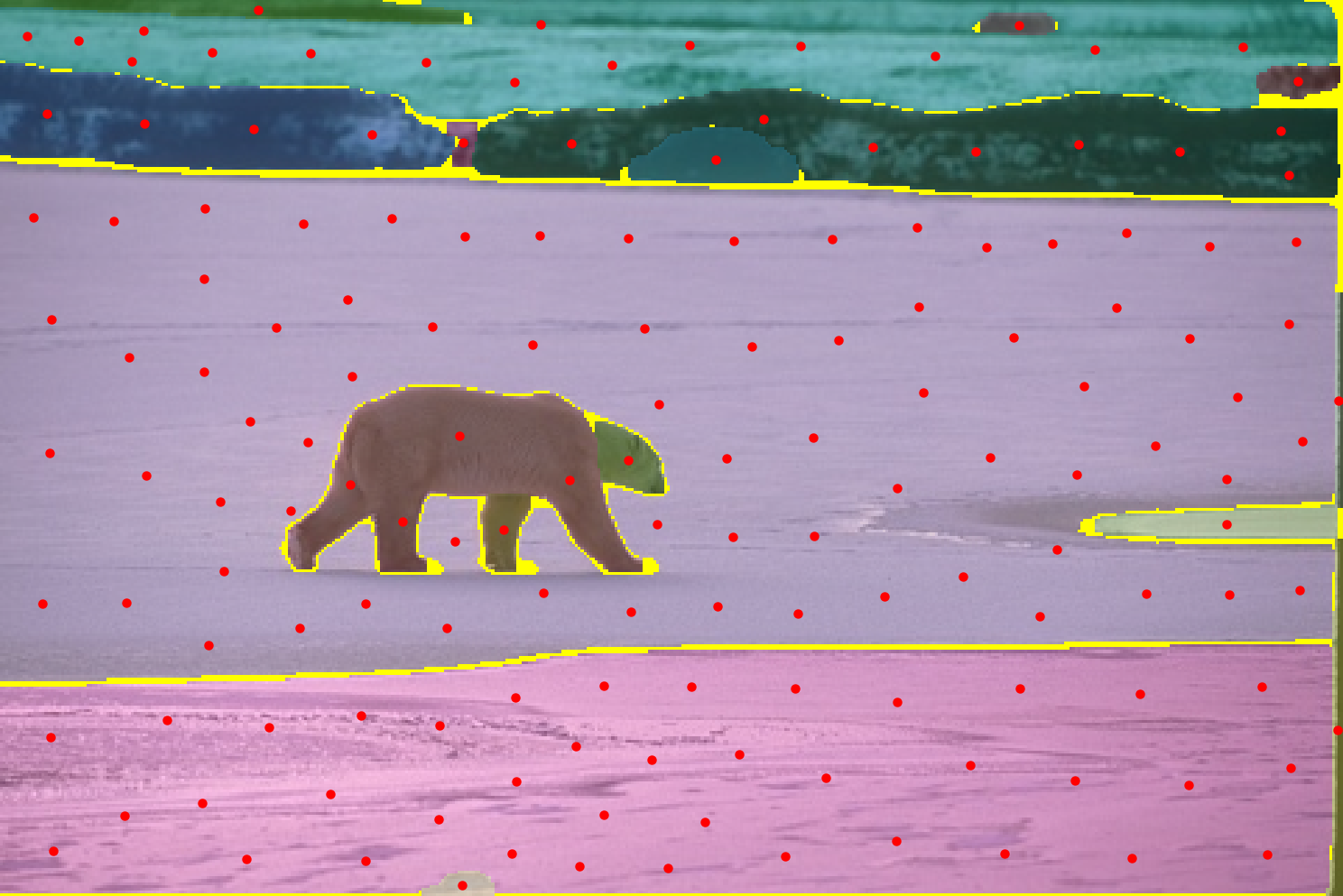} & 
\includegraphics[width=0.22\textwidth]{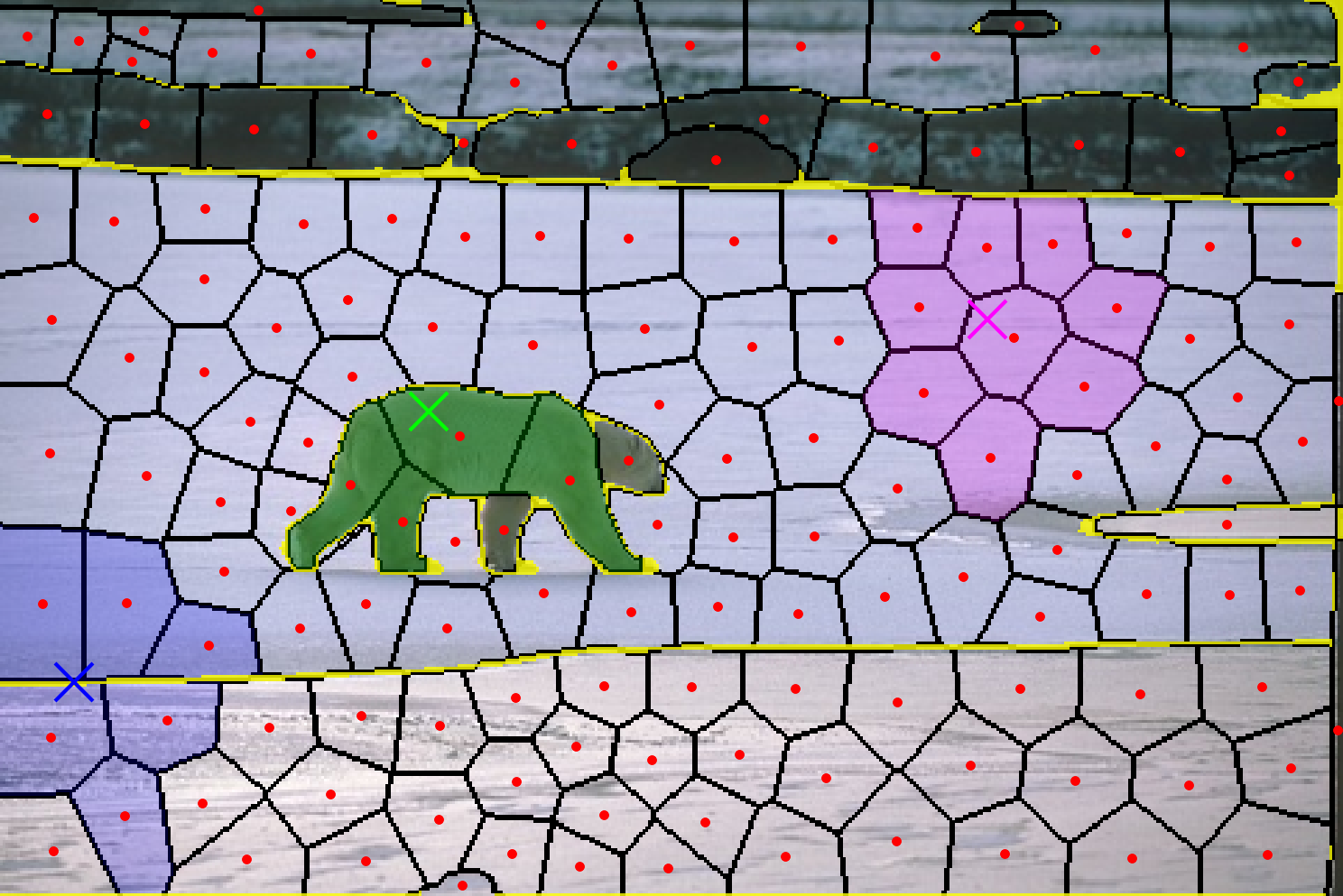} \\[-0.5ex]
(a) Seeds placement within objects & (b) Superpixels candidates
\end{tabular}}
\caption{\textbf{Seeds initialization with a prior object segmentation.}
(a) In each object $O_i$, a proportional number of seeds $K_i$ is randomly set. 
(b) The initial clustering is generated using spatial nearest neighbors.
For each pixel (crosses), the at most 9 closest superpixels within a distance threshold and in the same object are selected as candidates for association.
Pixels in uncertainty regions (yellow) can be associated to any close superpixels (see blue cross).}%
\vspace{-0.2cm}
 \label{fig:neighborhood}
\end{figure}

\smallskip
\noindent\textbf{Masked clustering based on prior segmentation.}
In SPAM, we
adapt the differentiable clustering 
such that it can
be constrained by a prior object segmentation.
SPAM guarantees that the generated superpixels 
are contained within the objects of this prior segmentation.
To achieve this, we propose several key modifications to the algorithm.

First, to 
initialize the constrained clustering, we assign seed points within each object based on its area, rather than using a rectangular sampling grid.
Specifically, we set $K_i=K|O_i|/|I|$ seeds in each object $O_i$,
 with $K$ the total number of superpixels, $|O_i|$ the object area and $|I|$ the image area.
A subset of pixels within each object is selected to draw $K_i$ seeds, and their positions are refined through 5 iterations of $K$-means clustering.
This approach ensures a balanced distribution of seeds across the image objects (see Fig.~\ref{fig:neighborhood}(b)).

Next, to prevent pixels from being assigned to superpixels outside their original object, their neighborhood (i.e., superpixel candidates) is constrained according to the object map.

Each pixel is thus associated with up to 9 of the closest superpixels within the same object and within a certain distance (see Fig.~\ref{fig:neighborhood}(c)).
Note how the pixel identified by the green cross cannot consider as candidates superpixels outside the same object in the prior segmentation.

Additionally, the prior object segmentation may also include thin uncertainty regions (in yellow) where pixels are not assigned to any object.
For these pixels, the neighborhood is not constrained by objects, allowing them to be associated with any nearby superpixels. 

Finally, as in most superpixel segmentation methods, a final step ensures superpixel spatial connectivity by merging small disconnected regions with adjacent ones.

In SPAM, this process is also constrained so that pixels can only be assigned to a superpixel contained in the same object of the prior segmentation.
Hence, only pixels of uncertainty regions are free to be assigned to any close superpixel.

\subsection{Prior object segmentation}
\label{subsec:sam}

We present an efficient automatic way to obtain a prior object segmentation to be used with SPAM at test time.
Note that any other prior segmentation could be used.

\smallskip
\noindent\textbf{Object segmentation.}
\label{subsubsec:object_seg}
The first step is to use the pre-trained SAM model~\cite{kirillov23sam} (with ViT-H weights) to generate a set of object proposals.
In this work,
we use a uniform sampling grid of $32 \times 32$ foreground points
as input prompt for the model.
Note that other input prompts could be considered, such as user-provided interactive clicks.
However, even after applying non-maximum suppression, the set of objects segmented by SAM does not form a proper partition of the image because (i) the objects often overlap, and (ii) the union of objects does not cover all the pixels in the image.
Thus, along the overlapping main objects, we 
obtain unlabeled pixels that can be 
categorized into: 
large areas, which can be viewed as background, and smaller regions, often corresponding to thin object boundaries.

\smallskip

\begin{figure}[t]
\centering
\includegraphics[width=0.98\linewidth,height=0.275\textwidth]{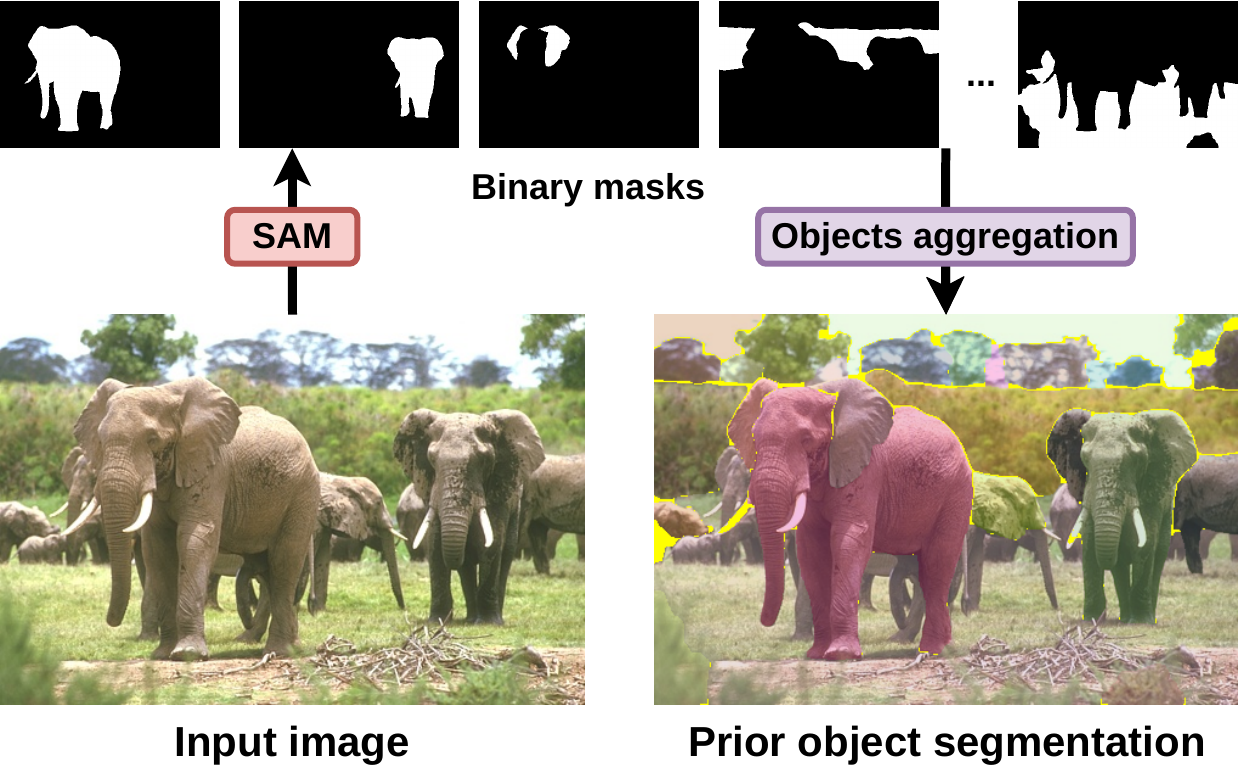}%
\caption{\textbf{Object proposals aggregation.}
Object masks provided by SAM~\cite{kirillov23sam} can overlap and do no cover the whole image. We aggregate these object proposals by iteratively removing overlaps and treating large unlabeled regions as additional (background) prior objects. Remaining uncertainty pixels (thin yellow regions) are handled without being constrained by prior objects. }%
\label{fig:objects_aggregation}
\end{figure}

\noindent\textbf{Object proposals aggregation.}
To ensure that we get a proper image partition from the initial object proposals, 
we propose the following aggregation method (see Fig.~\ref{fig:objects_aggregation}).
First, we retain only objects larger than a specified minimum area threshold. 

\begin{figure*}[t]
 \centering
 \newcommand{\www}{0.19\textwidth}
 {\footnotesize
 \begin{tabular}{@{\hspace{0mm}}c@{\hspace{1mm}}c@{\hspace{1mm}}c@{\hspace{2mm}}c@{\hspace{1mm}}c@{\hspace{1mm}}c@{\hspace{0mm}}}
 \includegraphics[width=\www]{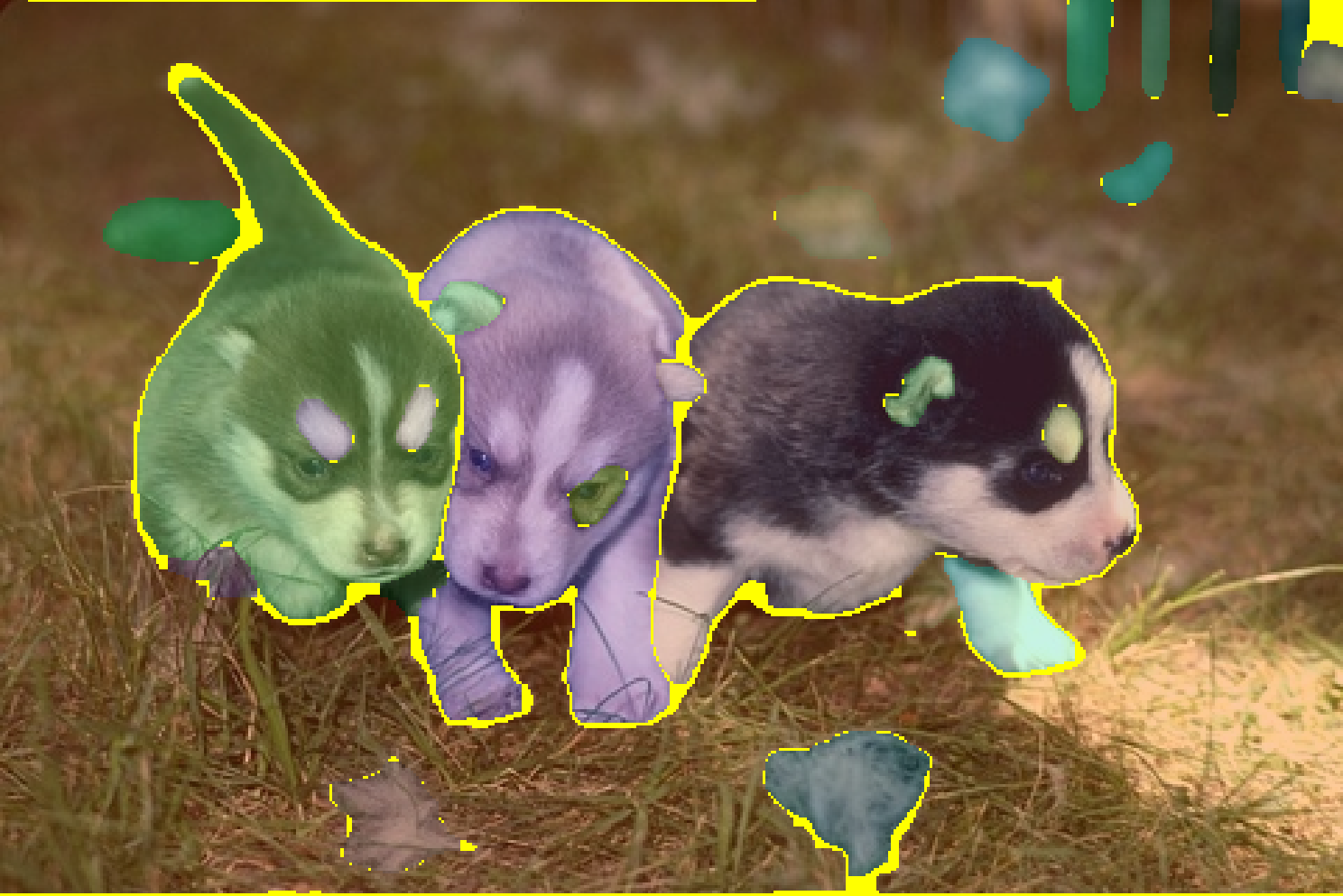} & 
 \includegraphics[width=\www]{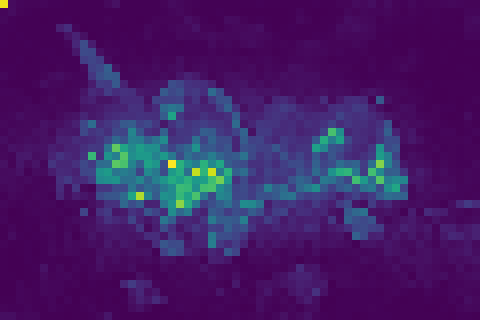} &
 \includegraphics[width=\www]{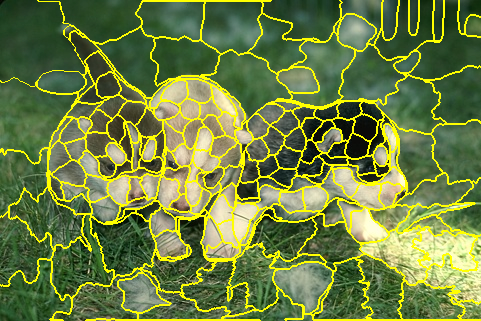} & 
 \includegraphics[width=\www]{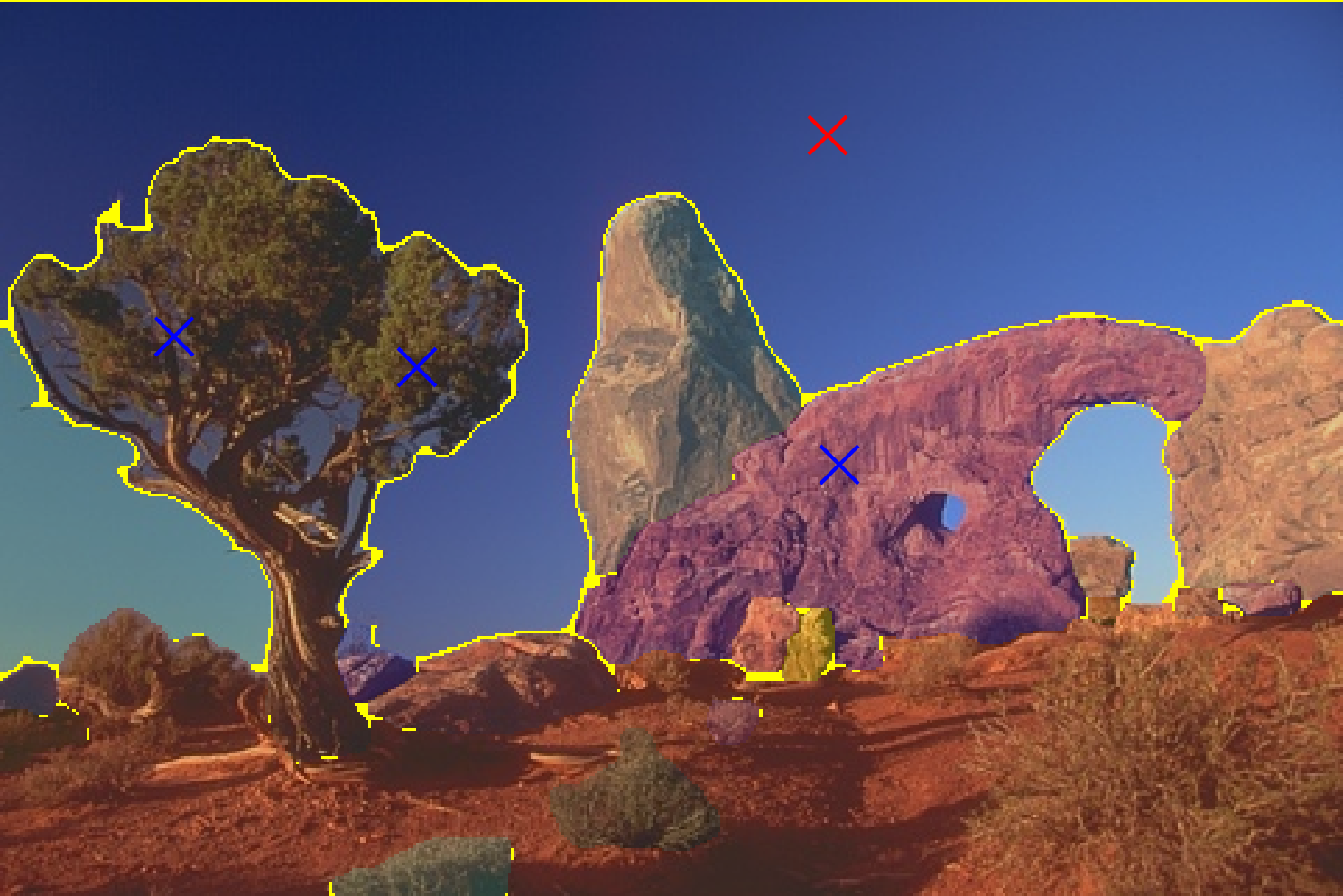} &
 \includegraphics[width=\www]{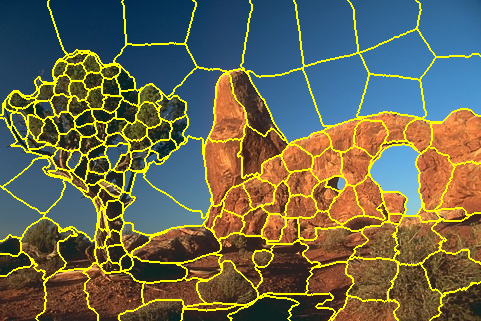} \\[-0.25ex] 
 \multicolumn{3}{@{\hspace{0mm}}c@{\hspace{1mm}}}{(a) \textit{Visual attention} (VA) mode using saliency map~\cite{caron2021emerging}} & \multicolumn{2}{@{\hspace{0mm}}c}{(b) \textit{User-driven attention} mode using clicks}  \\[1ex]
 \end{tabular}}
 \caption{\textbf{SPAM adaptive segmentation modes.}
 We propose two modes to focus on specific objects of interest.
 (a) For a given object map, we increase the number of superpixels generated in objects of interest by a factor $r$ (here $r=2.5$), providing a finer segmentation.
 These objects can be automatically selected using a visual attention map~\cite{caron2021emerging}.
 The number of superpixels is reduced outside the object(s) of interest to achieve $K$ requested superpixels.
 (b) The user can also manually select different factors for each object (blue=x2, red=x0.5).
 }
 \label{fig:adaptive_modes}
 \end{figure*}

\begin{figure*}[t!]
\centering
\newcommand{\heee}{0.15\textwidth}
{\footnotesize
\begin{tabular}{@{\hspace{0mm}}c@{\hspace{2mm}}c@{\hspace{5mm}}c@{\hspace{2mm}}c@{\hspace{2mm}}c@{\hspace{0mm}}}
\includegraphics[width=0.23\textwidth,height=\heee]{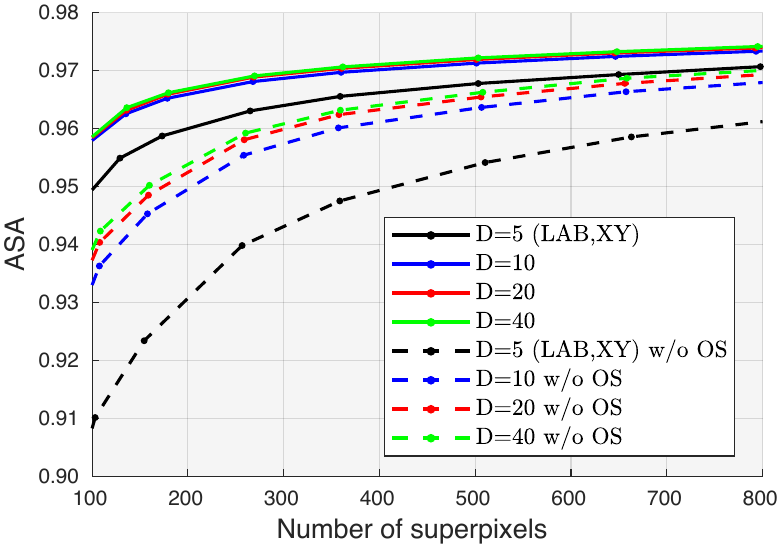}&
\includegraphics[width=0.23\textwidth,height=\heee]{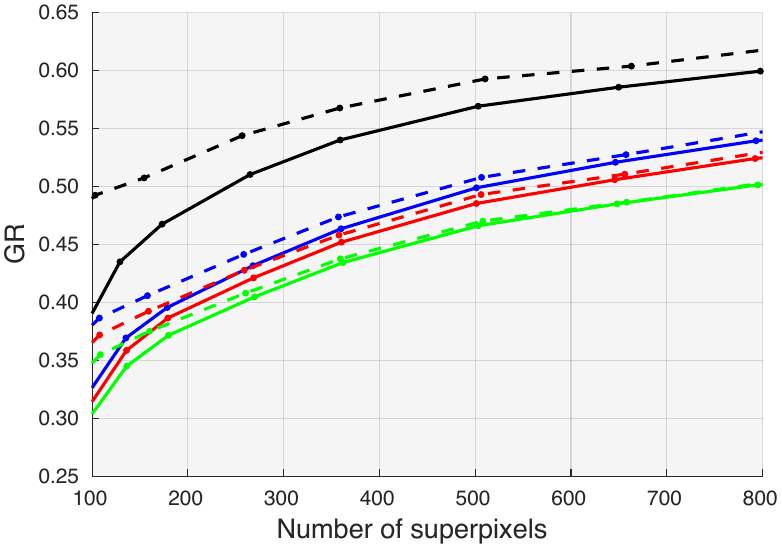}&
\includegraphics[width=0.23\textwidth,height=\heee]{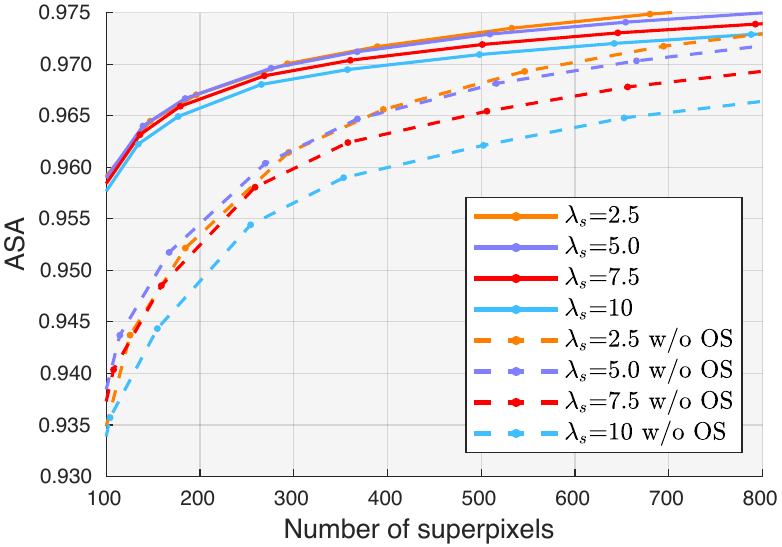}&
\includegraphics[width=0.23\textwidth,height=\heee]{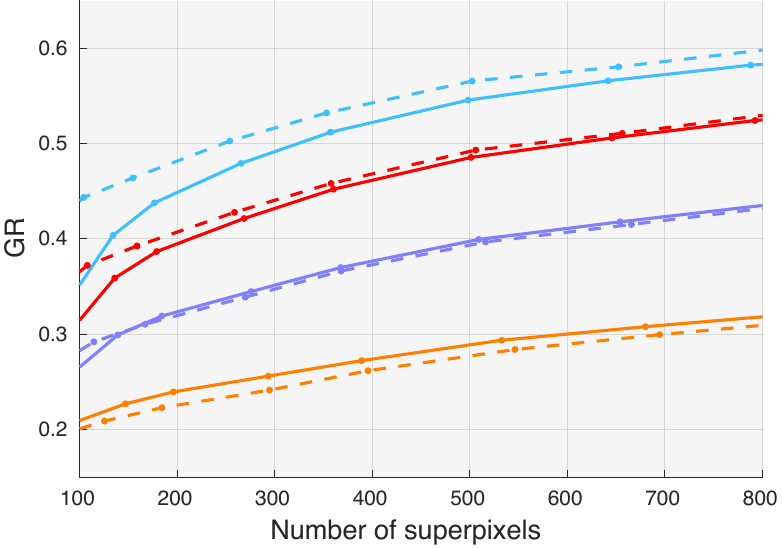}\\
\multicolumn{2}{c}{(a) Impact of the number of deep superpixel features $D$}  & 
\multicolumn{2}{c}{(b) Impact of spatial regularity as test time, set by $\lambda_s$} \\[1ex]
\end{tabular}}
\caption{\textbf{Ablation and parameters study for SPAM.}
Using prior object segmentation (solid lines) significantly improves the accuracy (ASA) of the models without (dashed lines, w/o OS).
(a) Increasing  the number of features generated by the encoder $D$ improves segmentation performance (ASA) and reaches a plateau. In contrast, a smaller $D$ gives more weight to low-level features, resulting in higher regularity (GR).
(b) 
With prior object segmentation, enforcing spatial regularity by increasing $\lambda_s$ yields significantly more regular superpixels (GR), without compromising accuracy (ASA).
}
\label{fig:abla_SSN}
\end{figure*}

Next, we remove overlaps between these objects by iteratively subtracting smaller objects from larger ones (i.e., the smallest objects are removed from the larger objects they overlap with, until no overlaps remain).
Remaining unlabeled pixels are then divided into connected components, with only those larger than the minimum area threshold being kept.
Additionally, we apply morphological opening to distinguish between large background areas and thin boundaries between objects.
The remaining large regions (non-overlapping objects from SAM and large connected background components) are treated as prior objects to constrain our superpixel algorithm.
During clustering, unlabeled pixels (yellow pixels in Fig.~\ref{fig:objects_aggregation}) are 
locally associated
without being constrained by prior objects, while 
ensuring spatial connectivity.

 \subsection{Adaptive segmentation modes}
 \label{subsubsec:adaptive}

For annotation purposes or to reduce computation time, we may aim to concentrate superpixels in regions of interest while decreasing their number in smoother or less salient areas, such as the background.
Several works have explored such approaches~\cite{Uziel:ICCV:2019:BASS,kang2020dynamic,zhou2023vine}, producing superpixels of varying sizes that do not follow an implicit regular grid.

In SPAM, we propose a \textit{visual attention} (VA) mode that automatically 
adjusts the scale to produce more superpixels in object of interests.
The main motivation for increasing superpixel density within regions of interest is to boost the chance of accurately segment pertinent sub-objects embedded in those regions.
Note that even by increasing the number of points for prompting SAM~\cite{kirillov23sam} do not guarantee to efficiently capture such objects.

Since~\cite{kirillov23sam} does not directly provide visual attention or saliency information (every region is considered as an object), we use DINO~\cite{caron2021emerging} to compute a saliency map of the input image.
Within each object of the prior map, if the average saliency score is over 0.1, the object is considered as of interest, or \textit{foreground} and otherwise as \textit{background}.  
The adjustment is set by a ratio $r$ so that for an image $|I|$,
a foreground object $O_i^f$ is initialized with 
$K_i^f = \frac{|O_i^f|}{|I|}{K}{r}$
superpixels, with $r>1$.
To generate $K$ superpixels overall,  
a background object $O_j^b$ is initialized with  
$K_j^b = \frac{|O_j^b|}{\sum\limits_p|O_p^b|}(K - \sum\limits_i K^f_i)$.
See examples of saliency and adapted segmentation in Fig.~\ref{fig:adaptive_modes}(a).

In the provided code, we also feature 
an \textit{user-driven attention} mode, where the user can easily refine the scale of the segmentation by selecting objects where to increase or decrease the number of superpixels (see Fig.~\ref{fig:adaptive_modes}(b)). 
 Note that SAM alone, in its click-guided variant, might fails to detect small and ambiguous objects.
The user-driven mode therefore becomes especially valuable to force the capture of such objects.

Additional examples are presented in supp. mat.


\begin{figure*}[t!]
\centering
\newcommand{\heee}{0.15\textwidth}
{\scriptsize
 \begin{tabular}{@{\hspace{0mm}}c@{\hspace{1.5mm}}c@{\hspace{1.5mm}}c@{\hspace{1.5mm}}c@{\hspace{1.5mm}}c@{\hspace{0mm}}}
 \rotatebox{90}{\hspace{0.9cm}{BSD}}&
\includegraphics[width=0.24\textwidth,height=\heee]{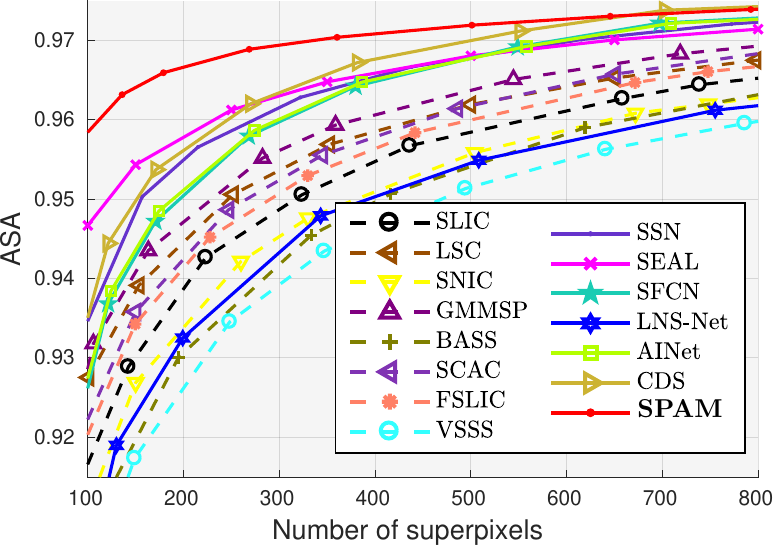}&
\includegraphics[width=0.24\textwidth,height=\heee]{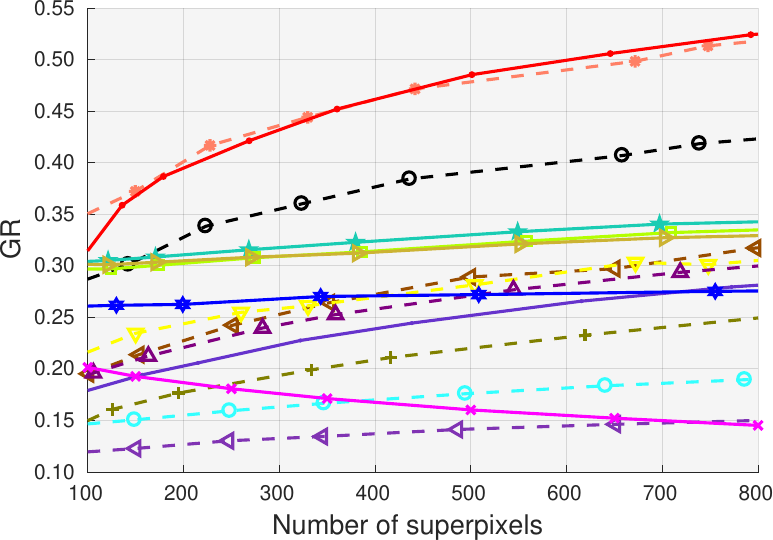}&
\includegraphics[width=0.21\textwidth,height=\heee]{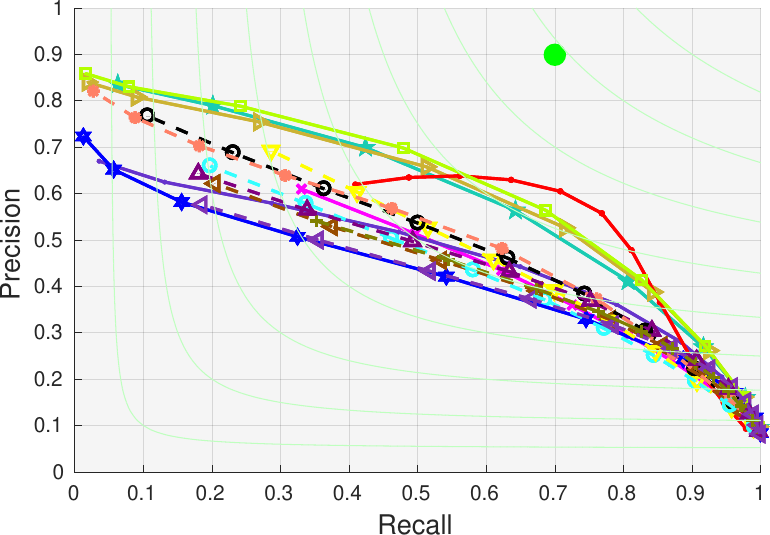}&
\includegraphics[width=0.24\textwidth,height=\heee]{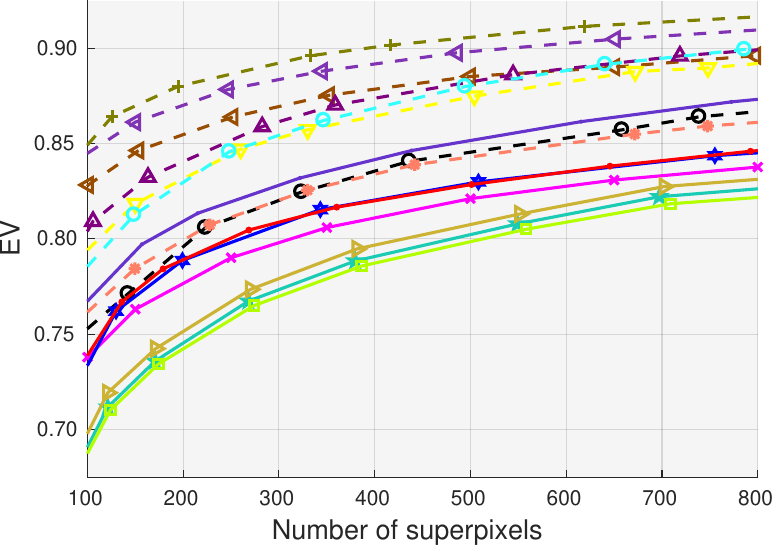}\\ 
 \rotatebox{90}{\hspace{0.75cm}{NYUv2}}&
\includegraphics[width=0.24\textwidth,height=\heee]{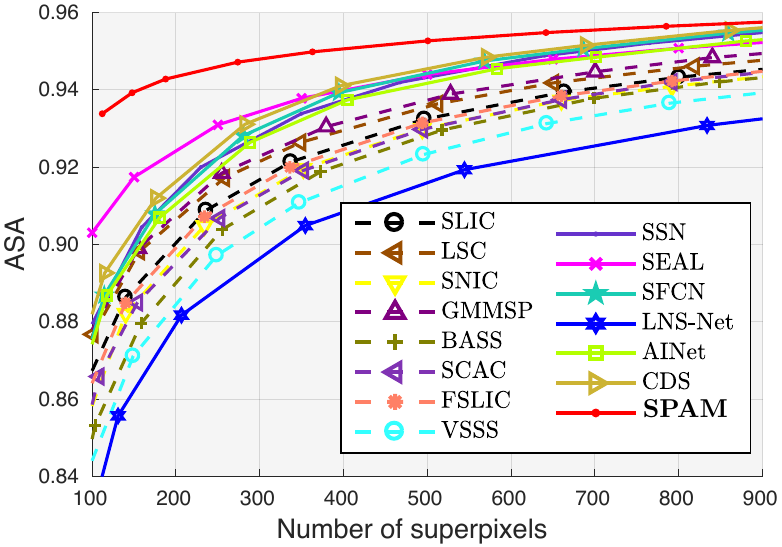}&
\includegraphics[width=0.24\textwidth,height=\heee]{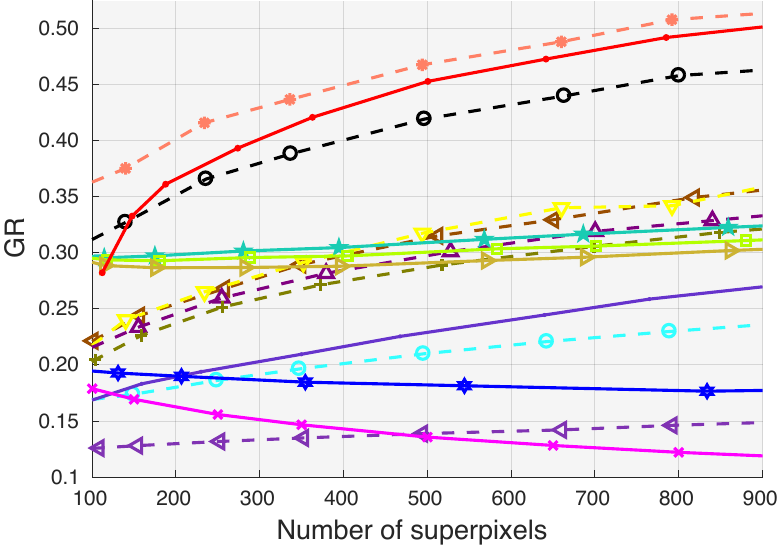}&
\includegraphics[width=0.21\textwidth,height=\heee]{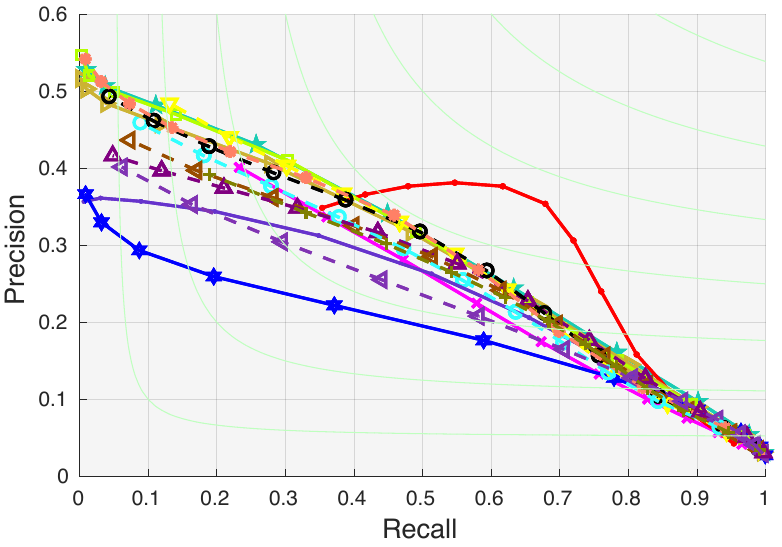}&
\includegraphics[width=0.24\textwidth,height=\heee]{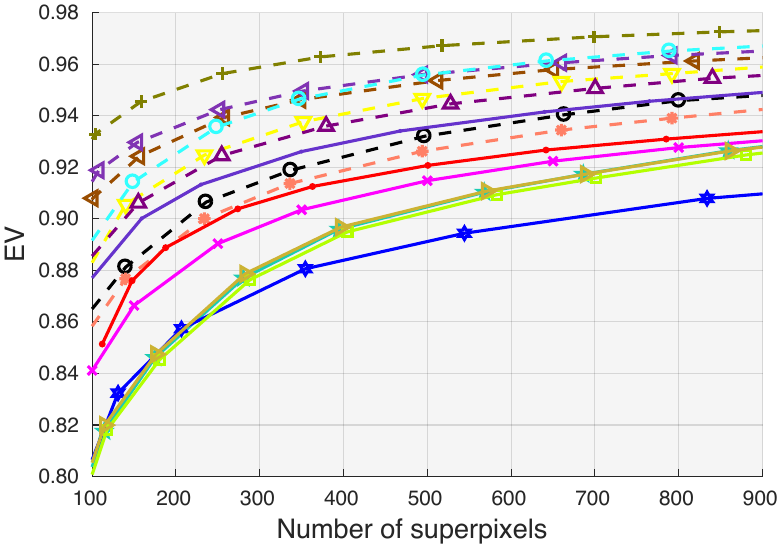}\\ 
 \rotatebox{90}{\hspace{0.9cm}{SBD}}&
\includegraphics[width=0.24\textwidth,height=\heee]{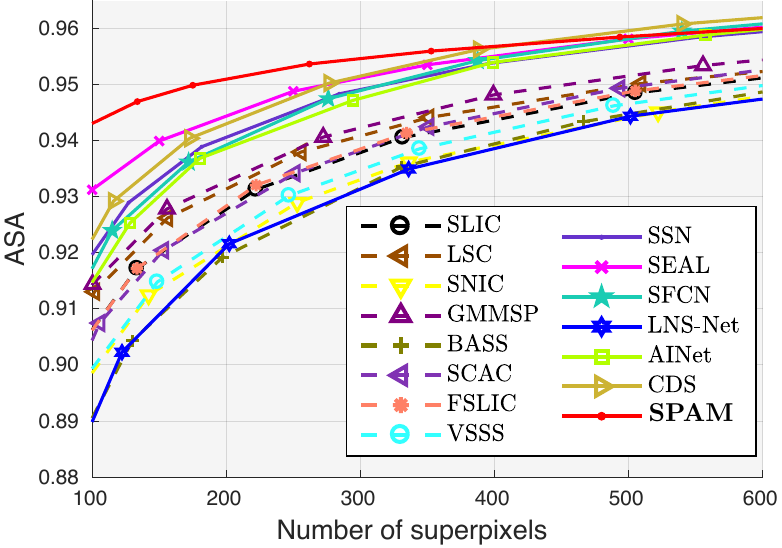}&
\includegraphics[width=0.24\textwidth,height=\heee]{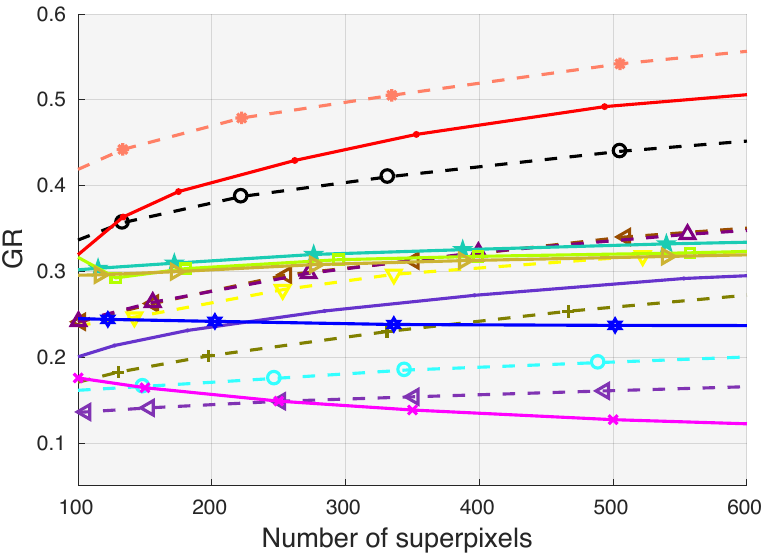}&
\includegraphics[width=0.21\textwidth,height=\heee]{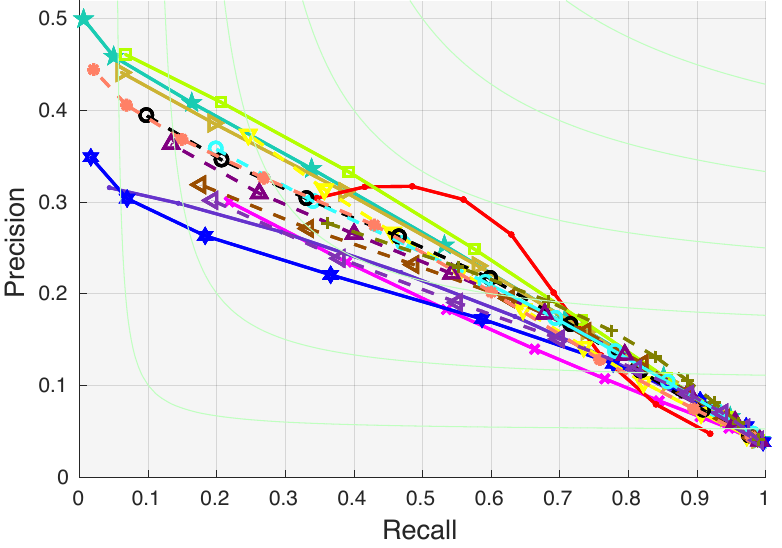}&
\includegraphics[width=0.24\textwidth,height=\heee]{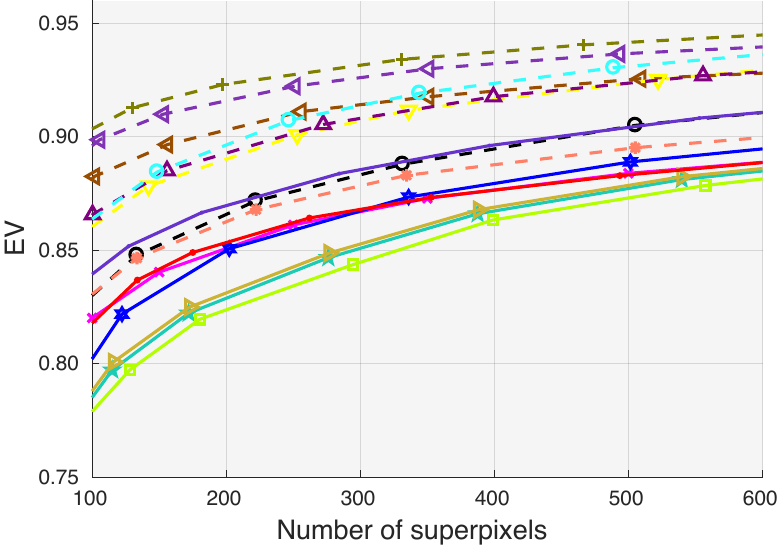}\\[1ex]
\end{tabular}}%
\caption{\textbf{Quantitative evaluation of superpixel methods.} 
SPAM (red lines) obtains the best object segmentation accuracy (ASA) compared to traditional (dashed lines) and DL-based (solid lines) methods, with a regularity (GR) among the highest, and the best of DL-based methods. 
SPAM also has accurate contour detection (Recall vs Precision) and EV scores consistent with other DL-based methods.}%
\label{fig:res_quanti}
\end{figure*}

\section{Experimental results}

\subsection{Validation framework}
\label{subsec:validation}

\textbf{Datasets.}
We train and evaluate our method on the BSD~\cite{martin2001}, 
the mainly used dataset in superpixel segmentation.
The BSD contains 200 train, 100 validation, and 200 test images, of size 321x481 pixels, with fine segmentation annotations (up to 173 objects per image).
We also report segmentation results on the NYUv2~\cite{silberman2012indoor} 
containing 399 images of size 448x608 pixels and the
SBD~\cite{gould2009decomposing} containing 477 images of size 240x320 pixels.

\smallskip
\noindent\textbf{Evaluation metrics.}
We use the recommended metrics and framework described
in~\cite{stutz2018superpixels,giraud2024_tip} to measure the main superpixel properties:
(i) Object segmentation with Achievable Segmentation Accuracy (ASA)~\cite{liu2011},
measuring the compliance of superpixels with the groundtruth objects; 
(ii)
Regularity with the Global Regularity (GR)~\cite{giraud2017_jei},
evaluating the convexity, balance, border smoothness of the superpixel shapes; 
(iii)
Contour detection with Boundary Recall (Rec.) vs Precision (Pre.)~\cite{martin2004},
comparing the superpixel and groundtruth object border 
following the protocol of~\cite{vandenbergh2012};
(iv)
Color homogeneity with Explained Variation (EV)~\cite{moore2008},
considering the average color variance in each superpixel;
(v) The average error percentage between the number of generated superpixels $\hat{K}$ compared to requested $K$,
$\Delta K = | \hat{K} - K |/ K$.
Detailed descriptions of these metrics are provided in the supp. mat.

\smallskip
\noindent\textbf{Training settings.}
SPAM is trained on the BSD dataset, 
allowing fair comparison with other  
methods that also train on this dataset. 
We train the model for $500$k iterations with a batch size of $6$ and a learning rate of $10^{-4}$.
The compactness parameter was set to $\lambda = 10^{-5}$.
The color and spatial feature weights are set to $\lambda_c = 0.26$ and $\lambda_s = 7.5$ for training and testing.
We keep the model achieving the best ASA score on the BSD validation set.
Note that only the CNN encoder of SPAM is trained, as SAM prior segmentations are only used during inference.
All models were trained and tested on a Nvidia RTX A5000 GPU.

\subsection{Ablation and parameters study}
\label{subsec:ablation}

We present the results of SPAM on the test set of the BSD in Fig.~\ref{fig:abla_SSN}.
We report the performance of the model for two test-time configurations: without prior object segmentation (dashed lines, w/o OS) and with prior object segmentation from SAM (solid lines).

\smallskip
\noindent\textbf{Deep features and prior segmentation.}
In Fig.~\ref{fig:abla_SSN}(a), we examine the influence of $D$, the number of deep features 

generated by the CNN encoder $E$ and used as input by the clustering algorithm.
First, we see that using prior object segmentation consistently enhances segmentation accuracy, highlighting the interest of our proposed prior segmentation approach.
As $D$ increases, segmentation accuracy (ASA) generally improves, but possibly at a slight cost to regularity (GR).
Note that setting $D=5$ restricts the model to color and spatial features only, resulting in a classical SLIC clustering.
The improvement with $D>5$, even with object segmentation, demonstrates the added value of deep superpixel features, capturing finer object details beyond what the large-scale model achieves.
In the following, we use the model trained with $D=20$, prior object segmentation and $\lambda_s = 7.5$, achieving a good compromise between segmentation accuracy and superpixels regularity. More details on the influence of $\lambda_s$ are given in the supp. mat.

\begin{figure*}[t!]
\centering
\newcommand{\hhh}{0.16\textwidth}
\newcommand{\hee}{0.1065\textwidth}
{\tiny \begin{tabular}{@{\hspace{0mm}}c@{\hspace{1mm}}c@{\hspace{1mm}}c@{\hspace{1mm}}c@{\hspace{1mm}}c@{\hspace{1mm}}c@{\hspace{1mm}}@{\hspace{0mm}}}
\includegraphics[width=\hhh,height=\hee]{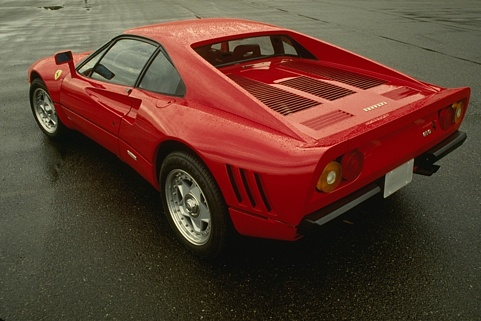}&
\includegraphics[width=\hhh,height=\hee]{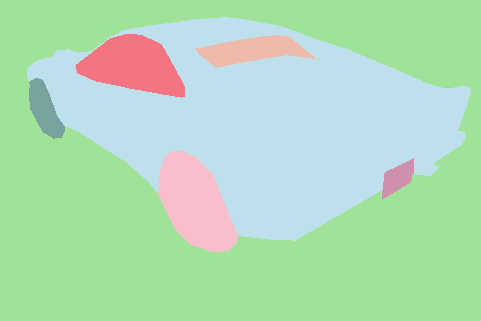}&
\includegraphics[width=\hhh,height=\hee]{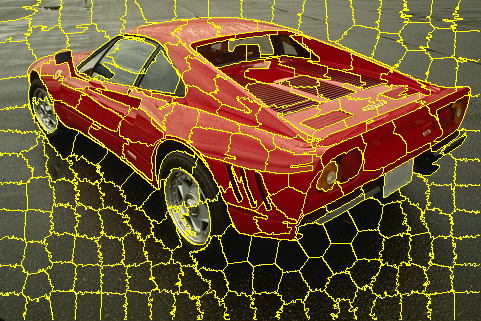}&
\includegraphics[width=\hhh,height=\hee]{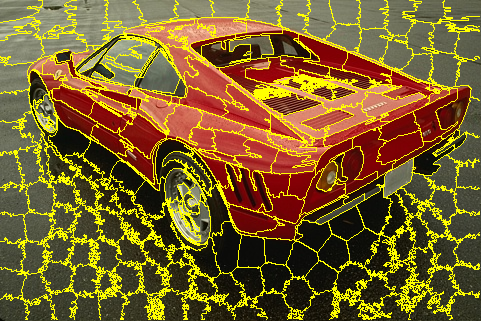}&
\includegraphics[width=\hhh,height=\hee]{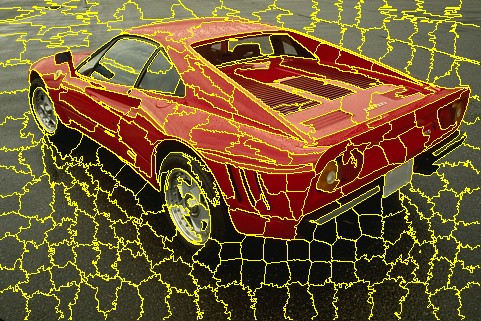}&
\includegraphics[width=\hhh,height=\hee]{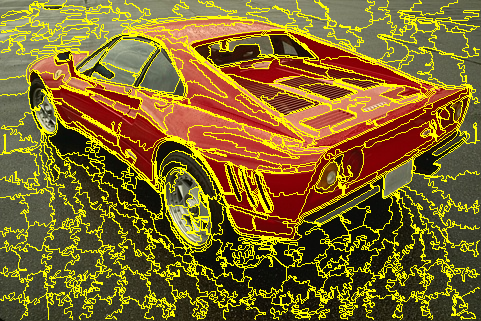}\\
Image & Groundtruth & SLIC \cite{achanta2012} & LSC \cite{li2015} & SNIC \cite{achanta2017superpixels}&SCAC \cite{yuan2021superpixels}\\[0.5ex]
\includegraphics[width=\hhh,height=\hee]{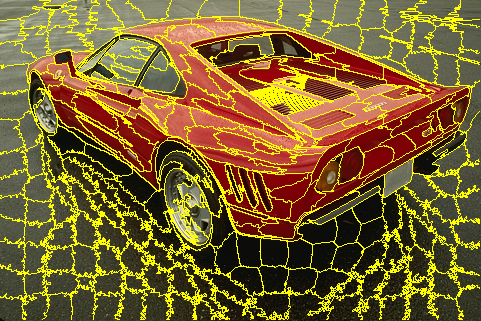}&
\includegraphics[width=\hhh,height=\hee]{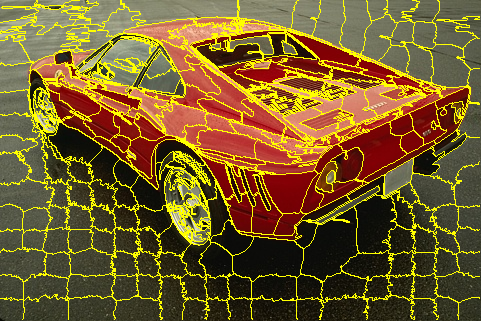}&
\includegraphics[width=\hhh,height=\hee]{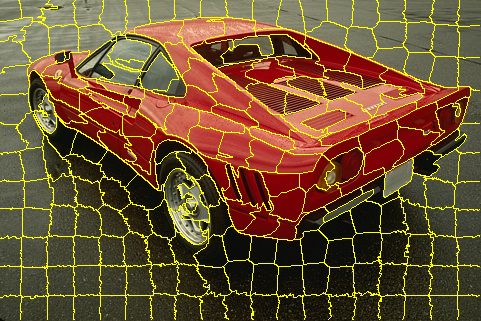}&
\includegraphics[width=\hhh,height=\hee]{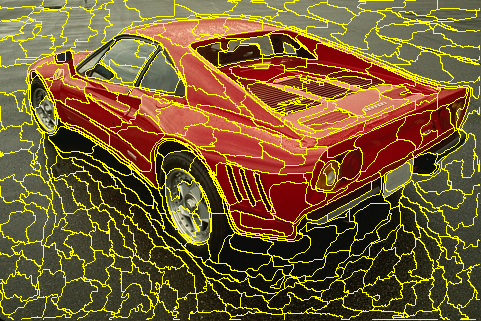}&
\includegraphics[width=\hhh,height=\hee]{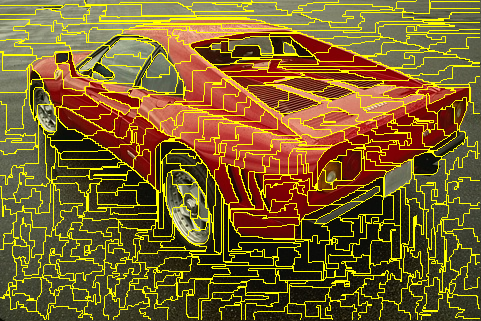}&
\includegraphics[width=\hhh,height=\hee]{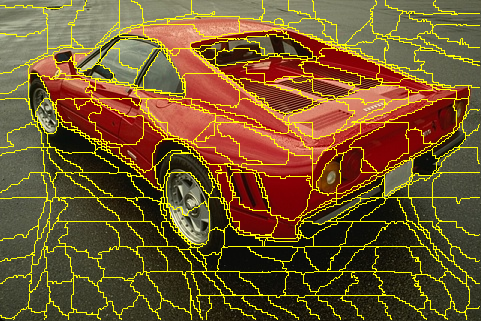}\\
 GMMSP \cite{Ban18} & BASS \cite{Uziel:ICCV:2019:BASS} & FSLIC \cite{wu2020fuzzy} & SSN \cite{jampani2018superpixel}&
 SEAL \cite{tu2018learning} & SFCN \cite{yang2020superpixel}\\[0.5ex]
\includegraphics[width=\hhh,height=\hee]{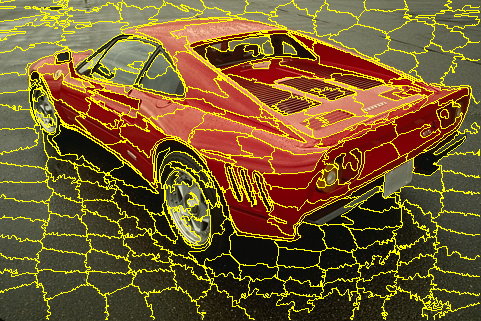}&
\includegraphics[width=\hhh,height=\hee]{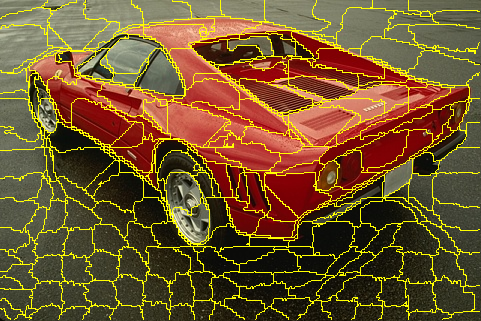}&
\includegraphics[width=\hhh,height=\hee]{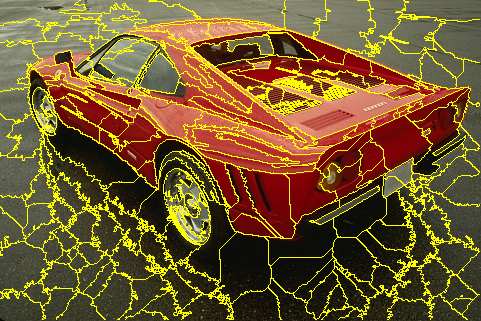}&
\includegraphics[width=\hhh,height=\hee]{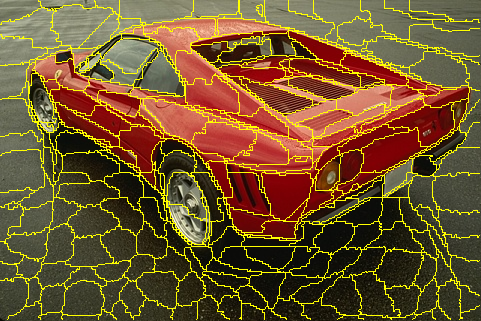}&
\includegraphics[width=\hhh,height=\hee]{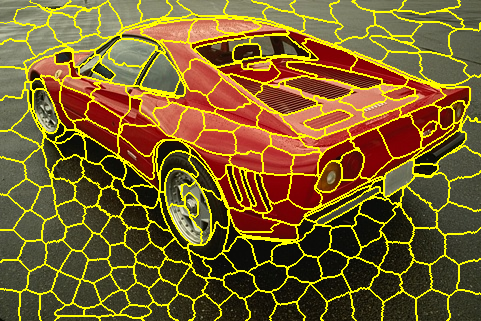}&
\includegraphics[width=\hhh,height=\hee]{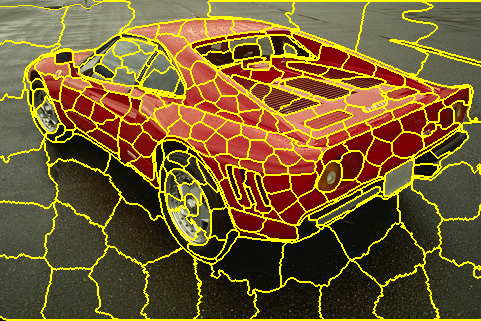}\\
 LNS-Net \cite{zhu2021learning} & AINet \cite{wang2021ainet} & VSSS \cite{zhou2023vine} & CDS \cite{xu2024learning} & SPAM & SPAM (VA $r$=2) \\[1ex]
\end{tabular}}%
\caption{\textbf{Qualitative evaluation of superpixel methods on the BSD test set.}
The number of requested superpixels is set to $K=250$. 
SPAM generates superpixels that are both more accurate and considerably more regular than other state-of-the-art 
DL-based methods.} 
  \label{fig:res_quali}
  \end{figure*}

 \subsection{Comparison to state-of-the-art}
\label{subsec:results}

We present an extensive performance comparison with state-of-the-art superpixel methods, categorizing them into traditional and deep learning (DL)-based methods.
All DL-based methods were trained on the BSD train set. 
Additional results are reported in the supp. mat.

\smallskip
\noindent\textbf{Quantitative evaluation.}
Evaluation results are shown in Fig.~\ref{fig:res_quanti}  
for the three test sets of BSD, NYUv2 and SBD.
With SPAM, we propose a very precise and robust superpixel method, significantly outperforming the state-of-the-art approaches, in terms of segmentation accuracy (ASA) and contour detection (Pre./Rec.),
yet producing very regular superpixels (GR)
contrary to all DL-based methods.
We also observe that no 
other method achieves high performance across multiple metrics simultaneously.
This highlights the contradictory nature of these metrics, especially color homogeneity (low-level) vs object segmentation (high-level features).
Nevertheless, SPAM achieves some of the highest EV scores among DL-based methods.
Finally, 
as reported in the supp. mat., 

SPAM provides the best respect of the requested superpixel number $K$ among DL-based methods after SEAL.
For $K=400$ on the BSD, SPAM gets a $\Delta K=0.036$ while the average $\Delta K$ of compared  methods is $0.108$.

SPAM inference takes
less than 200ms for a BSD image,
independently of the superpixel number and for any prior object segmentation.
When using SAM as prior segmentation model, it takes an additional 2s for a fine sampling grid of $32\times32$.
This prior computation can be largely reduced by using either a coarser grid or a more lightweight model like FastSAM~\cite{zhao2023fast} (see results in supp. mat.), or is completely neglected with user-provided masks.

\smallskip
\noindent\textbf{Qualitative evaluation.}
Examples of superpixel segmentations are shown in Fig.~\ref{fig:res_quali} and in the supp. mat.
See for instance how irregular can be the superpixels of DL-based methods such as SEAL, SFCN, AINet or CDS,
that produce thin superpixels around object contours or even unnatural straight superpixel borders in low contrast areas.
Such non easily identifiable superpixels may be an issue for interactive tasks~\cite{long2018efficient,berg2019ilastik}.
SPAM demonstrates accurate and significantly more regular superpixels than other methods.

Finally, as with other state-of-the-art segmentation algorithms, the CNN encoder of SPAM may still struggle to extract discriminative features in the presence of very low-contrast regions, extremely small target objects, or objects with ambiguous boundaries that would not be captured by the prior segmentation mask.
SPAM performance may also be sensitive to the quality of this input mask. 
When using a low quality mask, uncertainty regions can be added at the borders to
give SPAM more flexibility in the clustering process.

\subsection{Semantic segmentation refinement}
As additional downstream evaluation, we show that SPAM can enhance the performance of semantic segmentation methods.

To do so, we use DeepLabV3 \cite{chen2017rethinking} segmentations as  prior object maps.
Object borders are dilated by a 5x5 kernel to define uncertainty pixels, that are locally reassigned to objects by SPAM superpixels.
Such process improves the accuracy, {e.g.} from $79.48\%$ to $79.90\%$ mIoU on PACAL VOC2012 \cite{pascalvoc} using a ResNet101 as backbone.
Details are provided in the supp. mat.

\section{Conclusion}

In this work, we introduce the SuperPixel Anything Model (SPAM), which redefines the traditional superpixel segmentation paradigm by
leveraging the accuracy and generalization capacity of modern large-scale object segmentation models.
SPAM significantly outperforms state-of-the-art methods in accuracy while providing highly regular and easily identifiable superpixels
— two objectives widely regarded as contradictory.

Our clustering strategy enables to incorporate 
any prior object segmentation
and effectively resolves uncertainty regions.
Such accurate finer-scale segmentation beyond high-level objects
is crucial for many vision and editing pipelines.

SPAM performance and its 
new interactive modes, that further
enhance its versatility, 
open new possibilities for the community.
Future work will naturally explore the extension of SPAM to video processing.
{
\small
\bibliographystyle{ieeenat_fullname}
\bibliography{main}
}

\clearpage

\appendix
\clearpage
\begin{strip}
	\centering
	\parbox{0.7\linewidth}{\centering
	{\Large\bfseries
		-- Supplementary Material --\\
		Superpixel Anything: A general object-based framework for accurate yet regular
		superpixel segmentation\par}}
	\vspace{0.6em}
\end{strip}

In this supplementary material, 
we report: 
details on the used superpixel evaluation metrics in Section \ref{sec:metrics};
details on the used training losses, dataset,
visualization of deep superpixel features,
impact of parameters during inference and 
impact of prompt in SAM segmentation in Section \ref{sec:spam_details};
and additional results of SPAM segmentation including 
its adaptive modes, 
comparison to the state-of-the-art methods on the three datasets (BSD, NYUv2, SBD), 
and the semantic segmentation refinement process in Section \ref{sec:qualitative}.

\section{Superpixel evaluation metrics\label{sec:metrics}}

We use the recommended metrics and framework described
in~\cite{stutz2018superpixels,giraud2024_tip} to measure the main following superpixel properties, 
defined for a superpixel segmentation $\SSS=\{S_k\}$ of an image $I$ containing $|I|$ pixels and a set of groundtruth objects $\GG=\{G_j\}$.

\smallskip
\noindent\textbf{Object segmentation (ASA)}.
The main property when evaluating superpixel segmentation is the 
compliance of superpixels with image objects.
Hence, object segmentation metrics evaluate the average overlap of superpixels with the objects of a groundtruth segmentation.
Among existing metrics, the most commonly used is the
Achievable Segmentation Accuracy (ASA) \cite{liu2011},
which measures the maximum overlap of each superpixel with an object of the groundtruth segmentation, and is defined as:

{
	\begin{equation}
		\text{ASA}(\SSS,\GG) = \frac{1}{|I|}\sum_{S_k}\underset{G_j}{\max}|S_k\cap G_j|.  \label{asa}
	\end{equation}
}

\smallskip
\noindent\textbf{Regularity (GR)}.
The regularity is a very important aspect of superpixel segmentation.
As shown in \cite{giraud2017_jei} and in our parameter study,
this setting may have a significant impact on segmentation
performance.
Therefore, a method that enforces this aspect to generate regular
superpixels may perform worse than another one allowing stretched, thin or sometimes visually unidentifiable superpixels.

To evaluate this aspect, the GR metric \cite{giraud2017_jei}
was introduced to address the
lack of robustness of the circularity or compactness (CO) metric \cite{schick2014evaluation}.
GR is composed of two terms: a Shape Regularity Criteria (SRC) 
that independently evaluates each superpixel shape 
based on convexity, balance and contour smoothness,
and 
a Smooth Matching Factor (SMF) that evaluates the consistency of the shapes across the segmentation.
GR combines these two terms to be defined as: \vspace{-0.2cm} 

{
	\begin{equation}
		\text{GR}(\SSS) = \frac{1}{\sum\limits_{S_k\in \mathcal{S}} |S_k|}
		\sum\limits_{S_k\in \mathcal{S}} |S_k| \, \text{SRC}(S_k) \, \text{SMF}(S_k) . \label{gr} 
	\end{equation}
}

\smallskip
\noindent\textbf{Contour detection (F-measure)}.
Results for object contour detection may differ from object segmentation ones, 
for instance because of thin objects having large contours, or large objects well captured in terms of area but with superpixel borders misaligned with the ones of the groundtruth.  
The most commonly used metric is the standard Boundary Recall (BR) 
measuring the detection of groundtruth contours $\mathcal{B(\GG)}$ 
by the superpixel boundaries $\mathcal{B}(\SSS)$: \vspace{-0.2cm}

\begin{equation}
	\text{BR}(\SSS,\GG) = \frac{1}{|\mathcal{B}(\GG)|}\sum_{p\in\mathcal{B}(\GG)}\delta[\min_{q\in\mathcal{B}(\SSS)}\|p-q\|< \epsilon]  ,   \label{br}
\end{equation}
\noindent with 
$\delta[a]=1$ when $a$ is true and $0$ otherwise,
and $\epsilon$, a threshold distance set to $2$ pixels \cite{liu2011,vandenbergh2012}. 

Following \cite{stutz2018superpixels,giraud2017_jei}, 
Boundary Recall (BR) results should be reported along with Precision (P), which measures the percentage of true detections of superpixel boundaries, without favoring noisy superpixel borders.
In the results presented in Fig. 7 of the main paper, 
we use the following the framework introduced in 
\cite{vandenbergh2012}.
Superpixel boundaries obtained at different scales,
ranging from 50 to 1500 superpixels, are averaged to provide a contour detection map.
Precision and recall are computed for different thresholds of 
this average contour map and the $F$-measure reported in Tab. 1 of the main paper corresponds to the maximum value, calculated as:

\begin{equation}
	\text{F} = \frac{2.\text{P}.\text{BR}}{\text{P}+\text{BR}} . \label{fmeasure}
\end{equation}

\smallskip
\noindent
\textbf{Color homogeneity (EV)}.
Although color homogeneity within superpixels may conflict with object segmentation performance—since object borders do not always align naturally with color gradients—it remains an essential property to evaluate.
Hence, it should be noted that it is impossible to achieve both optimal object segmentation performance and maximum color homogeneity simultaneously.
The most used metric for this property is the Explained Variation (EV) defined as:  \vspace{-0.2cm}

\begin{align}
	\text{EV}(\SSS)   
	&= \frac{\sum_{S_k}{|S_k|\left(\mu(S_k) - \mu(I)\right)^2}} 
	{\sum_{p\in I}{\left(I(p)-\mu(I)\right)^2}} . \label{ev}
\end{align}
\smallskip

In the comparison to the state-of-the-art methods (Tab.~1 and Fig.~7 in the main paper),
SPAM obtains similar EV scores compared to the ones of the deep learning-based methods, while most traditional approaches achieve higher EV but lower segmentation accuracy.

\smallskip
\noindent
\textbf{Number of requested superpixels}
In Tab.~1 of the main paper, we report the 
average error percentage between the number of generated superpixels $\hat{K}$ compared to requested $K$ as: \vspace{-0.2cm}

\begin{equation}
	\Delta K = \frac{| \hat{K} - K |}{K}.
\end{equation}
Finally, note that in Figs.~6 and~7 of the main paper, results are reported for each method according to the actual average number of generated superpixels $\hat{K}$.

\section{Details on SPAM method\label{sec:spam_details}}

\noindent\textbf{Training loss.}
The loss function used to train SPAM is the same as the one of SSN~\cite{jampani2018superpixel}:
\begin{equation}
	\mathcal{L} = \mathcal{L}_\mathrm{seg}(\GG,\SSS_\mathrm{soft}) + 
	\lambda \, \mathcal{L}_\mathrm{compact}(F_s,\SSS_\mathrm{soft}) .
	\label{eq:loss}
\end{equation}
where, for $N$ pixels, $\GG \in \mathbb{R}^{N \times L}$ is the ground-truth segmentation representing hard pixel assignments to $L$ high-level regions, and $\SSS_\mathrm{soft} \in \mathbb{R}^{N \times N_{SP}}$ is the soft pixel assignments to $N_{SP}$ superpixels produced by SPAM, and $F_s  \in \mathbb{R}^{N \times 2}$ are the spatial features (2D coordinates) of the image pixels.

To compute the pixelwise categorical cross-entropy $\mathcal{L}_\mathrm{seg}$ between $\SSS_\mathrm{soft}$ and $\GG$, we first need to compute the projection $\hat{\GG} \in \mathbb{R}^{N \times L}$ of the soft assignments of superpixels onto the hard high-level regions, as:

\begin{equation}
	\hat{\GG} = 
	\tilde{\SSS}_\mathrm{soft} \, \hat{\SSS}_\mathrm{soft}^\mathsf{T} \, \GG 
\end{equation}
where $\hat{\SSS}_\mathrm{soft}$ and $\tilde{\SSS}_\mathrm{soft}$ the column- and row-normalized soft assigment matrices respectively.
Then, the pixelwise categorical cross-entropy at the ground-truth level is computed as:
\begin{equation}
	\mathcal{L}_\mathrm{seg}(\GG,\SSS_\mathrm{soft}) = 
	-\frac{1}{N} \sum_{i=1}^{N} \sum_{j=1}^{L} \GG_{ij} \log \hat{\GG}_{ij}.
\end{equation}

The compactness $\mathcal{L}_\mathrm{compact}$ is a regularization term designed to minimize the spatial variance of superpixels, ensuring they remain regular.
It is computed by first computing the spatial superpixel centers from spatial features $F_s$ and soft assignments $\SSS_\mathrm{soft}$, as:
\begin{equation}
	\bar{F}_s = \SSS_\mathrm{soft}^\mathsf{T} \, F_{s}. 
\end{equation}

Then, the compactness term is obtained by computing the sum of distances between pixelwise spatial features and their corresponding superpixel center picked from the hard assignment $\SSS_\mathrm{hard}$ such as:
\begin{equation}
	\hspace{-0.2cm}\mathcal{L}_\mathrm{compact}(F_s,\SSS_\mathrm{soft}) = \sum_{i=1}^N \left\| F_s(i) - \bar{F}_s(\mathcal{S}_\mathrm{hard}(i)) \right\|_2.
\end{equation}
The compactness parameter was set to $\lambda = 10^{-5}$ as in~\cite{jampani2018superpixel}.

\smallskip
\noindent\textbf{Training dataset.}
As for other deep learning-based superpixel methods, SPAM 
is trained on the train set of the BSD dataset, which contains 200 images.
These images were manually annotated up to 8 times per image, resulting
in 1063 groundtruth object segmentation maps.
During training, data augmentation is randomly applied 
on each image of the minibatch of size 6,
with horizontal flips, rescale, and crops to 200x200 pixels.

\smallskip
\noindent\textbf{Superpixel feature maps.}
In Fig. \ref{fig:deep_features} we visualize examples of the deep superpixel feature maps $F_e$ learned by our CNN encoder.
We see how the feature maps are activated for different image regions.

\begin{figure*}[h]
	\centering
	\newcommand{\hhh}{0.305\textwidth}
	{\scriptsize \begin{tabular}{@{\hspace{0mm}}c@{\hspace{1mm}}c@{\hspace{1mm}}c@{\hspace{1mm}}c@{\hspace{0mm}}}
			&\includegraphics[width=\hhh]{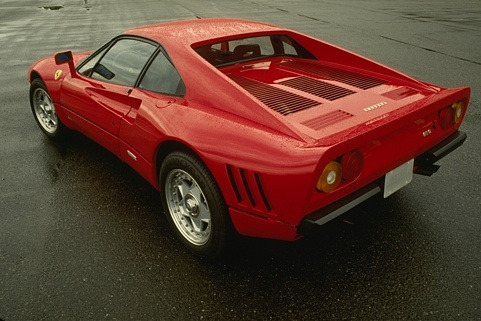}&
			\includegraphics[width=\hhh]{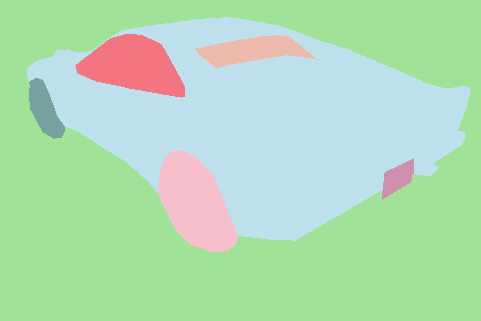}&
			\includegraphics[width=\hhh]{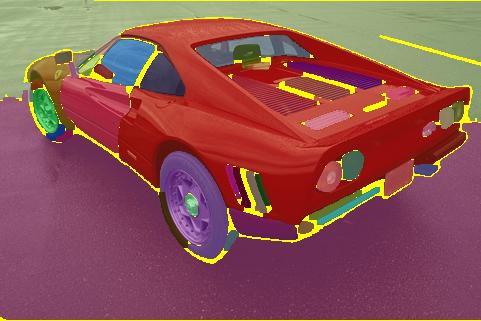}\\
			&(a) Image & (b) Groundtruth & (c) Object segmentation  \\[1ex]
			\rotatebox{90}{(d) Superpixel number $K$}&
			\includegraphics[width=\hhh]{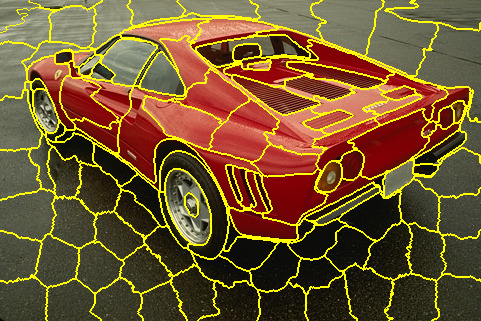}&
			\includegraphics[width=\hhh]{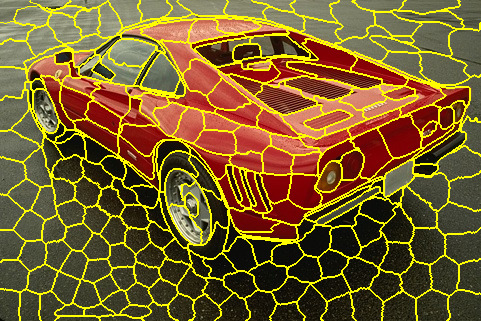}&
			\includegraphics[width=\hhh]{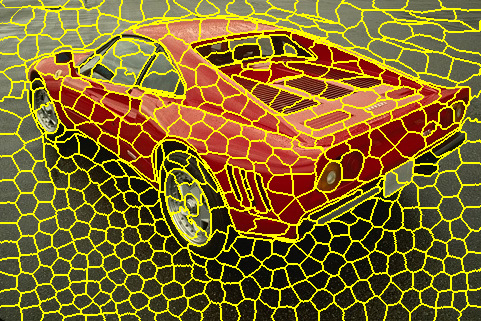}\\
			&$\mathbf{K=100}$, $\lambda_s=7.5$, $\lambda_c=0.26$ & $\mathbf{K=250}$, $\lambda_s=7.5$, $\lambda_c=0.26$ & $\mathbf{K=500}$, $\lambda_s=7.5$, $\lambda_c=0.26$ \\[1ex]
			\rotatebox{90}{(e) Regularity parameter $\lambda_s$}&
			\includegraphics[width=\hhh]{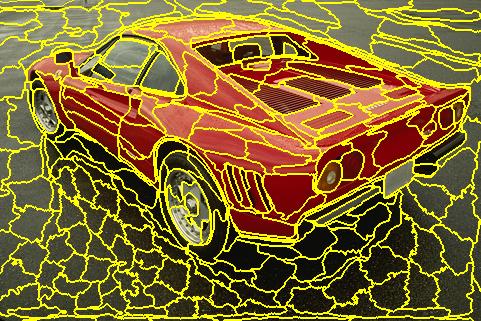}&
			\includegraphics[width=\hhh]{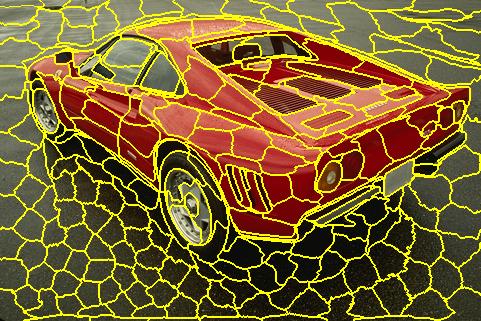}&
			\includegraphics[width=\hhh]{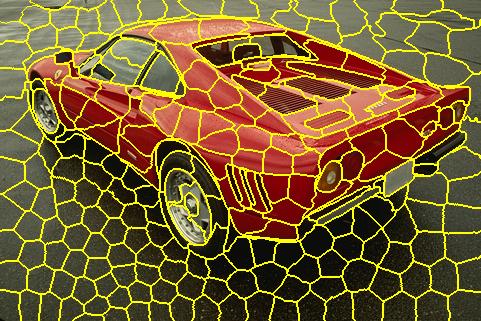}\\
			&{$K=250$}, $\mathbf{\lambda_s=2.5}$, $\lambda_c=0.26$ & {$K=250$}, $\mathbf{\lambda_s=5}$, $\lambda_c=0.26$ & {$K=250$}, $\mathbf{\lambda_s=10}$, $\lambda_c=0.26$ \\[1ex]
			\rotatebox{90}{\hspace{0.1cm} (f) Color parameter $\lambda_c$}&
			\includegraphics[width=\hhh]{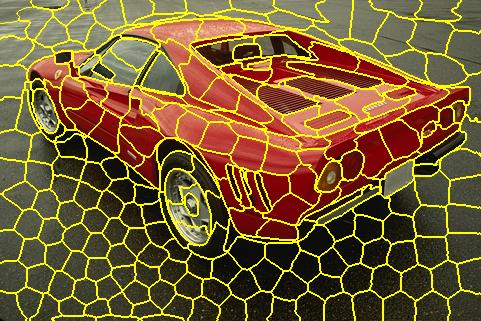}&
			\includegraphics[width=\hhh]{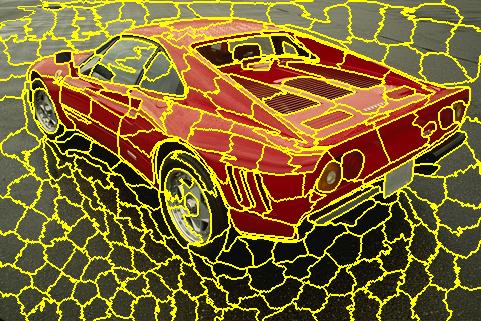}&
			\includegraphics[width=\hhh]{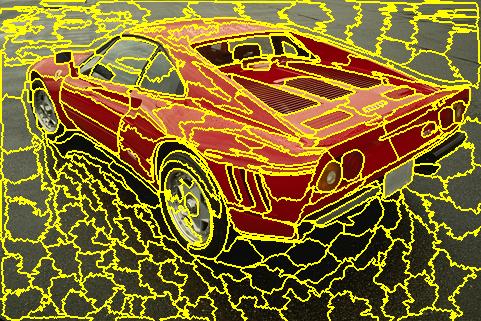}\\
			&{$K=250$}, $\lambda_s=7.5$, $\mathbf{\lambda_c=0.1}$ & {$K=250$}, {$\lambda_s=7.5$}, $\mathbf{\lambda_c=1}$ & {$K=250$}, {$\lambda_s=7.5$}, $\mathbf{\lambda_c=2.5}$ \\[1ex]
	\end{tabular}}%
	\caption{\textbf{Qualitative results of SPAM segmentation on BSD images for different settings.}
		(d) SPAM can generate superpixels at different scales with strong adherence to the number of requested superpixels.
		(e) Spatial parameter $\lambda_s$ enforces spatial regularity,
		while (f) color parameter $\lambda_c$ enforces adherence to low-level color features (input Lab).
		Decreasing $\lambda_s$ and increasing $\lambda_c$ both lead to lower regularity.
		Nevertheless, the generated superpixels remain different since in the first case, low-level colors and deep superpixel features prevail in the differentiable clustering, while in the second case only color low-level features prevail.
		Note that we use $\lambda_s=7.5$ and $\lambda_c=0.26$ as default settings. \vspace{4cm}
	}
	\label{fig:res_quali_SPAM_K}
\end{figure*}

\smallskip
\noindent  \textbf{Impact of parameters at inference time.}
In Fig.~\ref{fig:res_quali_SPAM_K}, we show SPAM superpixel segmentation results for different settings: 
number of requested superpixels $K$ (Fig.~\ref{fig:res_quali_SPAM_K}(d)); 
spatial scale factor $\lambda_s$ applied to spatial features $[X,Y]$ (Fig.~\ref{fig:res_quali_SPAM_K}(e)); 
color scale factor $\lambda_c$ applied to color features $[L,a,b]$ (Fig.~\ref{fig:res_quali_SPAM_K}(f)).
With the settings of $\lambda_s$ and $\lambda_c$, 
we can respectively enforce spatial regularity and adherence to 
low-level features (Lab colors).

\smallskip
\noindent  \textbf{Impact of prompt in SAM.}

SPAM inference takes approximately 200ms for an image of the BSD,
independently of the number of superpixels and for any prior object segmentation.
Note that this prototype implementation could be further improved with more refined optimizations.

\begin{figure*}[t!]
	\centering
	{\footnotesize
		\begin{tabular}{@{\hspace{0mm}}c@{\hspace{1mm}}c@{\hspace{1mm}}c@{\hspace{1mm}}c@{\hspace{1mm}}c@{\hspace{0mm}}}
			\includegraphics[width=0.19\textwidth]{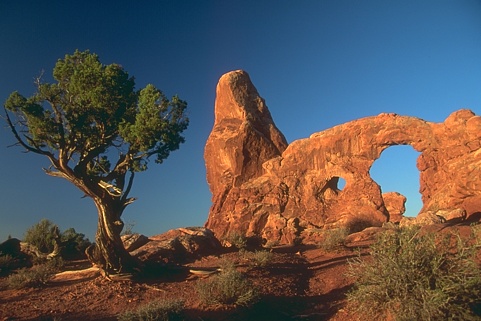}&
			\includegraphics[width=0.19\textwidth]{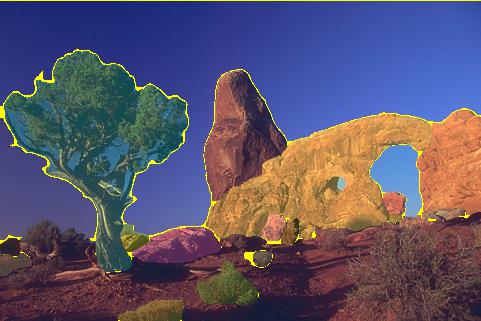}&
			\includegraphics[width=0.19\textwidth]{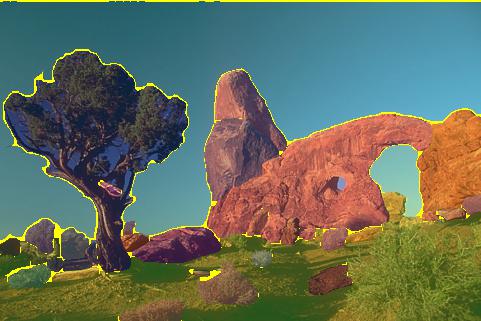}&
			\includegraphics[width=0.19\textwidth]{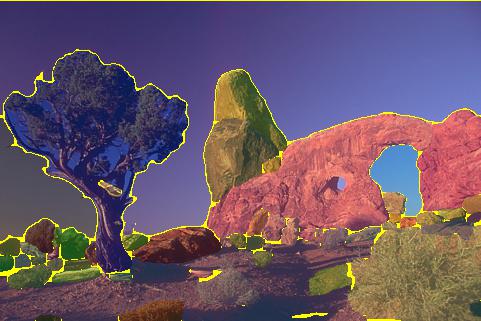}&
			\includegraphics[width=0.19\textwidth]{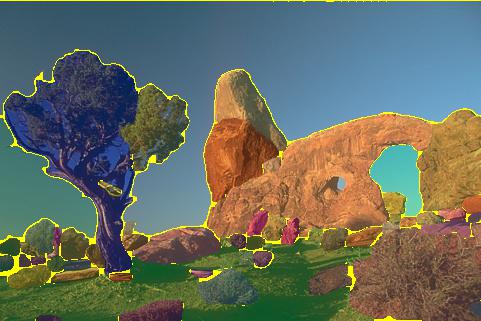}\\
			Image & 8x8 (21 objects) &  16x16 (34 objects) &  32x32 (54 objects) & 64x64 (57 objects)
			\\[1ex]
			\includegraphics[width=0.19\textwidth]{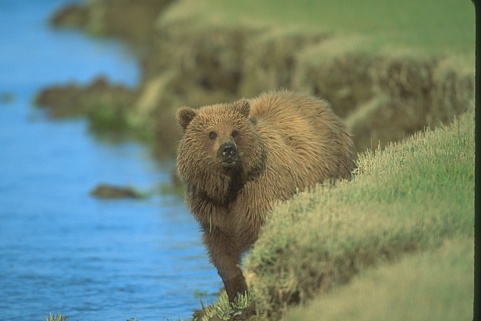}&
			\includegraphics[width=0.19\textwidth]{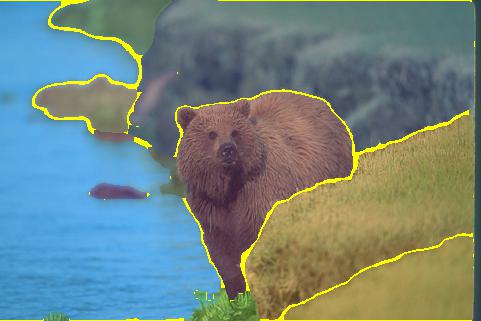}&
			\includegraphics[width=0.19\textwidth]{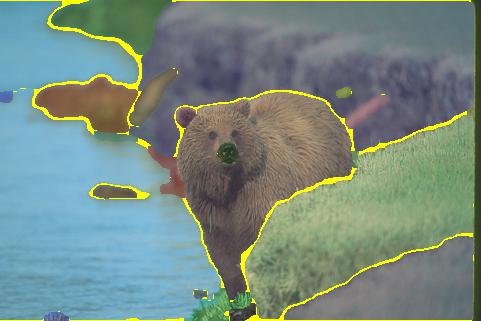}&
			\includegraphics[width=0.19\textwidth]{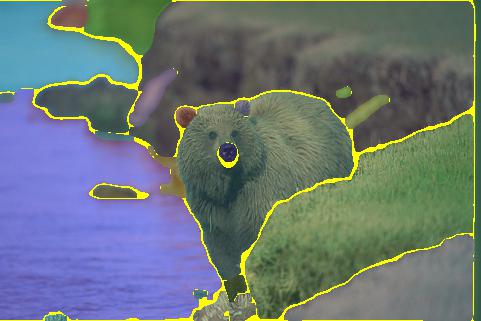}&
			\includegraphics[width=0.19\textwidth]{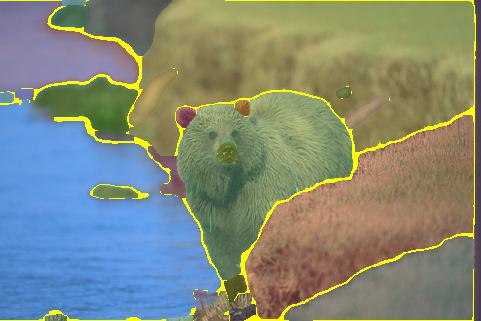}\\
			Image &  8x8 (15 objects) &  16x16 (24 objects) &  32x32 (27 objects) & 64x64 (31 objects)\\[1ex]
	\end{tabular}}
	\caption{\textbf{Examples of prior object segmentations from SAM.} We show the results of our object proposals aggregation for different input grid sizes to the SAM method (unlabeled pixels from the object aggregation are in yellow).} 
	\label{fig:ablation_sam}
\end{figure*}

We show in Fig.~\ref{fig:ablation_sam} examples of the 
influence of the number of grid points as input prompt for the SAM prior object segmentation.
This number affects the number of objects but requires more computational time.
On a BSD image, the object segmentation with SAM {requires on average an additional 0.75s, 1.15s, 2.53s, 8.4s} respectively for 8x8, 16x16, 32x32 and 64x64 input grid points. 
We report in Fig.~\ref{fig:ablation_sam_results}, the quantitative impact of this setting on the segmentation.
The object segmentation performance increases, at the expense of a slight decrease of  regularity and precision in contour detection, which logically suffers from more object-level segmentation.
Unless otherwise mentioned, we use a 32x32 grid of input points as the default.

While the use of SAM with input grids can be time consuming, note that SPAM can be used with any other prior object segmentation.
Table \ref{tab:fastsam} reports the performance of the SPAM variant using FastSAM \cite{zhao2023fast} as the object detector on the BSD.
This version achieves an inference time of approximately 50ms, representing an improvement by a factor 50 compared to version using SAM. This makes it particularly suitable for applications where inference speed is a critical requirement.
The performance in terms of segmentation accuracy (ASA) is slightly lower compared to the version using SAM.
Nevertheless, it still outperforms the accuracy of other state-of-the-art methods, \emph{i.e.} with an ASA for $K=400$ of 0.9685 while the best compared method CDS \cite{xu2024learning} gets an ASA of 0.9676 (cf. Table \ref{tab:evaluation}).

\begin{table}
	\centering
	\caption{\textbf{Semantic segmentation refinement.} We report mean IoU (mIoU $\uparrow$) and pixel-wise accuracy (ACC $\uparrow$) for semantic segmentations with DeepLabV3 on the VOC2012 validation set.  Segmentations were refined with $K$=$500$ superpixels.} 
	\renewcommand{\arraystretch}{1.05}
	{\footnotesize
		\setlength{\tabcolsep}{5pt}
		\begin{tabular}{@{\hspace{1mm}}lcccc@{\hspace{1mm}}}
			
			\textbf{Method (backbone)} & mIoU & ACC 
			\\ \hline
			\arrayrulecolor{gray}\hline
			\arrayrulecolor{black}
			DeepLabV3 (RN50) & 78.48 & 92.30 \\
			DeepLabV3 (RN50)  $\cap$ SLIC~\cite{achanta2012} & 77.48 & 92.13  \\
			DeepLabV3 (RN50)  $\cap$ AINet~\cite{wang2021ainet} & 78.41 & 92.35  \\
			DeepLabV3 (RN50)  w/ NN & 78.00 & 92.11 \\
			DeepLabV3 (RN50)  w/ SPAM  & \textbf{78.91} & \textbf{92.47} \\
			\arrayrulecolor{gray}\hline
			\arrayrulecolor{black}
			DeepLabV3 (RN101) & 79.48 & 92.74 \\
			DeepLabV3 (RN101)  $\cap$ SLIC~\cite{achanta2012} & 78.44 & 92.56  \\
			DeepLabV3 (RN101)  $\cap$ AINet~\cite{wang2021ainet} & 79.40 & 92.78  \\
			DeepLabV3 (RN101)  w/ NN & 78.72 & 92.48 \\
			DeepLabV3 (RN101)  w/ SPAM  & \textbf{79.90} & \textbf{92.90} \\ \hline
	\end{tabular}}
	\label{tab:refinement_supp}
\end{table}

\begin{table*}[t!]
	\centering
	\caption{\textbf{Performance of state-of-the-art methods across the BSD, NYUV2 and SBD datasets.} We report ASA($\uparrow$), GR($\uparrow$), F($\uparrow$), EV($\uparrow$), $\Delta$K($\downarrow$)
		for $K=400$ superpixels. Best and second results in each category are respectively in bold and underlined font.
	}
	\renewcommand{\arraystretch}{1.025}
	\setlength{\tabcolsep}{2.15pt} 
	\footnotesize 
	\begin{tabular}{c|l|ccccc|ccccc|ccccc|} 
		\cline{2-17} 
		&&
		\multicolumn{5}{c|}{\textbf{BSD}} & 
		\multicolumn{5}{c|}{\textbf{NYUv2}} & 
		\multicolumn{5}{c|}{\textbf{SBD}} 
		\\
		&\textbf{Method}   & \textbf{ASA} & \textbf{GR} & \textbf{F} & {EV} & {$\Delta$K} &  \textbf{ASA} & \textbf{GR} & \textbf{F} & {EV} & {$\Delta$K} &           \textbf{ASA} & \textbf{GR} & \textbf{F} & {EV} & {$\Delta$K} \\ 
		\hline
		\multicolumn{1}{|c|}{\multirow{8}{*}{\rotatebox{90}{Traditional methods}}}
		&SLIC \cite{achanta2012} 
		& 0.9547 & \underline{0.377} & \underline{0.529} & 0.836 & 0.078 
		& 0.9257 & \underline{0.401} & 0.387 & 0.924 & 0.040
		& 0.9437 & \underline{0.422} & 0.335 & 0.895 & 0.056 
		\\
		\multicolumn{1}{|c|}{} &LSC \cite{chen2017}
		& \underline{0.9585} & 0.272 & 0.497 & 0.879 & 0.044 
		& \underline{0.9295} & 0.298 & 0.367 & 0.949 & 0.035 
		& \underline{0.9460} & 0.321 & 0.314 & 0.920 & 0.038
		\\
		\multicolumn{1}{|c|}{} &SNIC \cite{achanta2017superpixels} 
		& 0.9509 & 0.269 & 0.525 & 0.864 & 0.057  
		& 0.9231 & 0.300 & \underline{0.387} & 0.941 & \underline{0.006} 
		& 0.9392 & 0.304 & {\color{gray}\textbf{0.337}} & 0.916 & 0.043
		\\
		\multicolumn{1}{|c|}{} &GMMSP \cite{Ban18} 
		& {\color{gray}\textbf{0.9606}} & 0.258 & 0.507 & 0.874 & 0.044 
		& {\color{gray}\textbf{0.9317}} & 0.285 & 0.368 & 0.937 & 0.084 
		& {\color{gray}\textbf{0.9482}} & 0.322 & 0.319 & 0.918 & 0.141
		\\
		\multicolumn{1}{|c|}{} &BASS \cite{Uziel:ICCV:2019:BASS}
		& 0.9496 & 0.209 & 0.498 & {\color{gray}\textbf{0.901}} & 0.220 
		& 0.9209 & 0.274 & 0.360 & {\color{gray}\textbf{0.964}} & 0.095
		& 0.9394 & 0.242 & 0.324 & {\color{gray}\textbf{0.937}} & 0.339
		
		\\
		\multicolumn{1}{|c|}{} &SCAC \cite{yuan2021superpixels}
		& 0.9578 & 0.137 & 0.478 & \underline{0.892} & \underline{0.014} 
		& 0.9227 & 0.136 & 0.323 & \underline{0.952} & {\color{gray}\textbf{0.006}}
		& 0.9448 & 0.156 & 0.293 & \underline{0.932} & {\color{gray}\textbf{0.008}}
		
		\\
		\multicolumn{1}{|c|}{} &FSLIC \cite{wu2020fuzzy}
		& 0.9563 & {\color{gray}\textbf{0.461}} & {\color{gray}\textbf{0.543}} & 0.834 & 0.058 
		& 0.9246 & {\color{gray}\textbf{0.449}} & {\color{gray}\textbf{0.390}} & 0.919 & 0.040 
		& 0.9442 & {\color{gray}\textbf{0.519}} & \underline{0.335} & 0.888 & 0.045
		
		\\
		\multicolumn{1}{|c|}{} &VSSS \cite{zhou2023vine}
		& 0.9463 & 0.170 & 0.496 & 0.869 & {\color{gray}\textbf{0.009}}
		& 0.9152 & 0.201 & 0.364 & 0.950 & {0.008}    
		& 0.9415 & 0.188 & 0.331 & 0.924 & \underline{0.016}
		
		\\ 
		\arrayrulecolor{gray}\hline
		\arrayrulecolor{black}
		\multicolumn{1}{|c|}{\multirow{7}{*}{\rotatebox{90}{DL-based methods}}}
		& SSN \cite{jampani2018superpixel}    
		& 0.9654 & 0.239 & 0.517 & \textbf{0.842} & 0.253 
		& 0.9373 & 0.216 & 0.348 &  \textbf{0.930} & 0.336 
		& 0.9539 & 0.273 & 0.302 & \textbf{0.896} & 0.132
		
		\\
		\multicolumn{1}{|c|}{} &SEAL \cite{tu2018learning}   
		& 0.9659 & 0.168 & 0.509 & 0.811 & \textbf{0.000}
		& 0.9399 & 0.143 & 0.352 & 0.907 &   \textbf{0.000}
		& 0.9551 & 0.135 & 0.293 & 0.877 & \textbf{0.000}
		
		\\
		\multicolumn{1}{|c|}{} &SFCN \cite{yang2020superpixel}   
		& 0.9648 & \underline{0.324} & 0.600 & 0.791 & 0.085 
		& 0.9395 & \underline{0.305} & \underline{0.388} & 0.896 & 0.127 
		& 0.9547 & \underline{0.326} & 0.343 & 0.867 & 0.107
		
		\\
		\multicolumn{1}{|c|}{} &LNS-Net \cite{zhu2021learning} 
		& 0.9503 & 0.271 & 0.469 & 0.820 & 0.453 
		& 0.9084 & 0.184 & 0.278 & 0.884 & 0.555 
		& 0.9386 & 0.238 & 0.275 & \underline{0.879} & 0.436
		
		\\
		\multicolumn{1}{|c|}{} &AINet \cite{wang2021ainet}   
		& 0.9651 & {0.314} & \underline{0.619} & 0.787 & 0.104 
		& 0.9370 & 0.297 & 0.386 & 0.894 & 0.156
		& 0.9540 & 0.318 & 0.360 & 0.863 & 0.141
		
		\\
		\multicolumn{1}{|c|}{} &CDS \cite{xu2024learning}
		& \underline{0.9676} & 0.313 & 0.606 & 0.797 & 0.091 
		& \underline{0.9412} & 0.288 & 0.383 & 0.897 & 0.133 
		& \underline{0.9566} & 0.313 & \underline{0.345} & 0.869 & 0.109 
		
		\\
		\multicolumn{1}{|c|}{} &\textbf{SPAM} & \textbf{0.9708} & \textbf{0.461} & \textbf{0.652} & \underline{0.820} & \underline{0.036}
		& \textbf{0.9506} & \textbf{0.430} & \textbf{0.468} & \underline{0.915} & \underline{0.040}
		& \textbf{0.9568} & \textbf{0.471} & \textbf{0.393} & 0.876 & \underline{0.018}   
		\\
		\hline  
	\end{tabular}
	\label{tab:evaluation}
\end{table*}

\begin{figure*}[ht]
	\centering
	\newcommand{\www}{0.24\textwidth}
	{\footnotesize \begin{tabular}{@{\hspace{0mm}}c@{\hspace{1mm}}c@{\hspace{1mm}}c@{\hspace{1mm}}c@{\hspace{0mm}}}
			\includegraphics[width=\www]{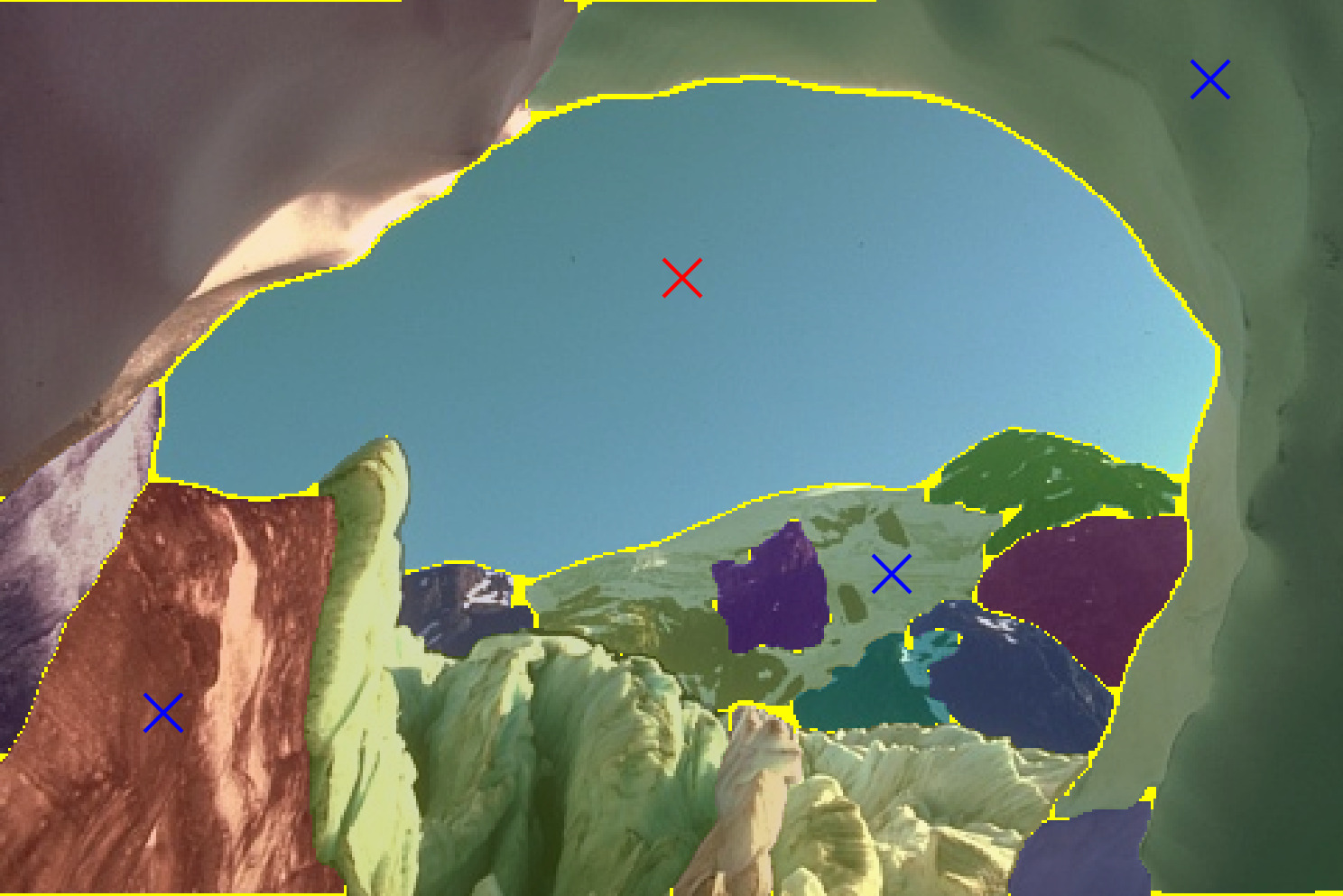} &
			\includegraphics[width=\www]{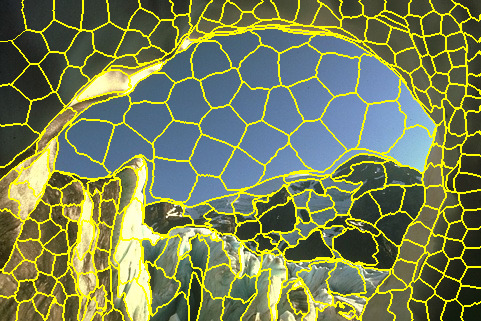} &
			\includegraphics[width=\www]{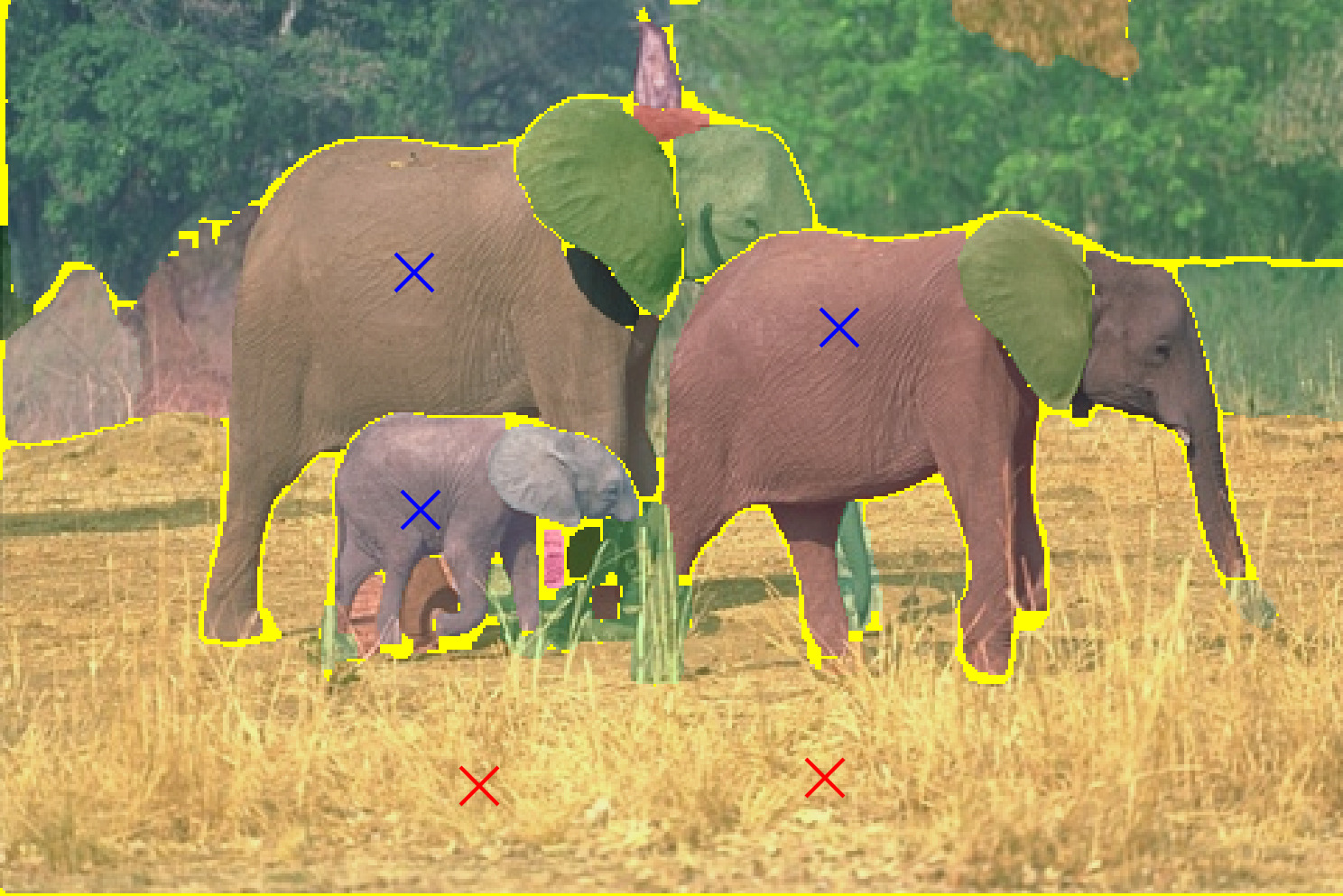} &
			\includegraphics[width=\www]{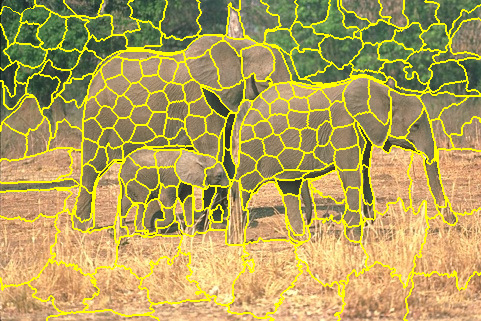} \\[-1ex]
	\end{tabular}}%
	\caption{\textbf{SPAM user-driven attention adaptive segmentation mode.}
		The results are generated for $K=250$ superpixels.
		The user can manually select different factors for each object.
		Crosses identify the user clicks (blue=x2 and red=x0.5 scale increase factor).
	}
	\label{fig:res_quali_SPAM_mode_user}
\end{figure*}
\begin{figure*}[ht]
	\centering
	\newcommand{\jjj}{0.16\textwidth}
	\newcommand{\ddd}{0.1065\textwidth}
	{\footnotesize
		\setlength{\tabcolsep}{1pt} 
		\begin{tabular}{c@{\hspace{0mm}}cccccccccccc}
			&&&&\includegraphics[width=\jjj]{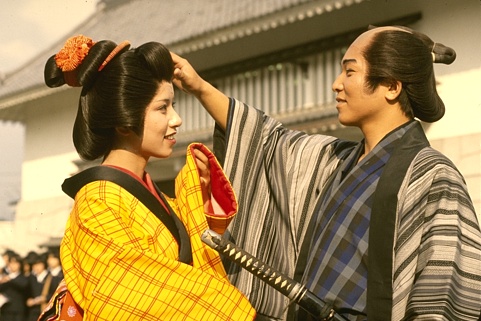}&&
			\includegraphics[width=\jjj]{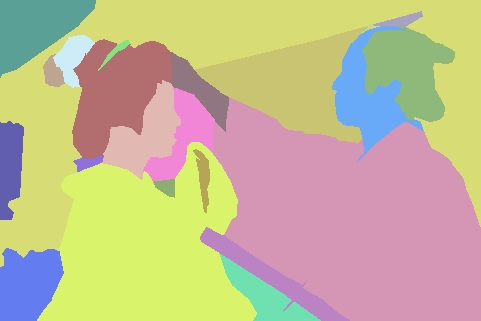}&&
			\includegraphics[width=\jjj]{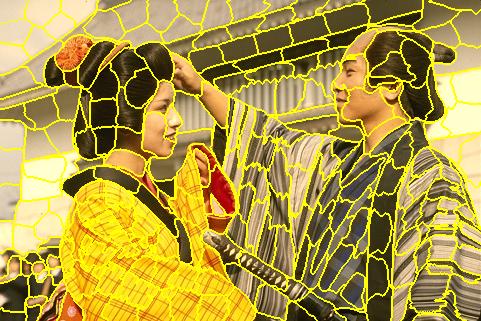}\\
			&&&&RGB image&&Groundtruth&&SPAM superpixels\\[0.5ex]
			&\rotatebox{90}{Inputs $F_c$, $F_s$}&
			\includegraphics[width=\jjj,height=\ddd]{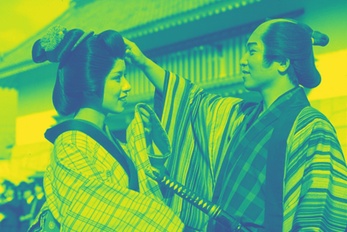}&
			\includegraphics[height=\ddd]{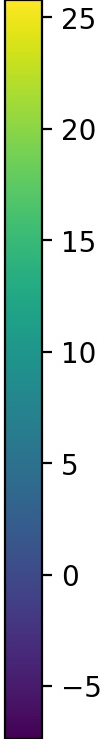}&
			\includegraphics[width=\jjj,height=\ddd]{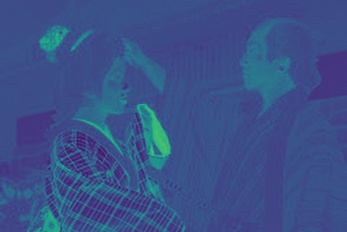}&
			\includegraphics[height=\ddd]{img/dsf_scale_fc.png}&
			\includegraphics[width=\jjj,height=\ddd]{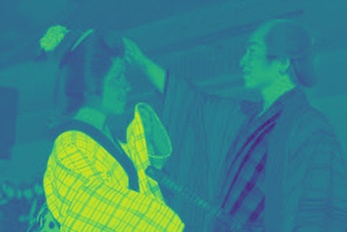}&
			\includegraphics[height=\ddd]{img/dsf_scale_fc.png}&
			\includegraphics[width=\jjj,height=\ddd]{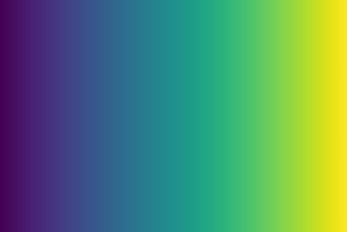}&
			\includegraphics[height=\ddd]{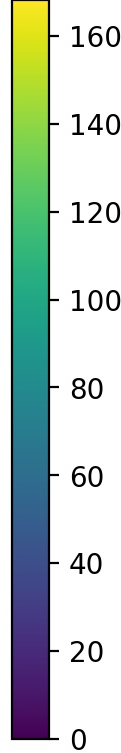}&
			\includegraphics[width=\jjj,height=\ddd]{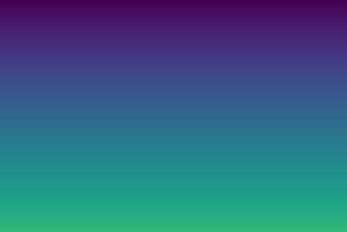}&
			\includegraphics[height=\ddd]{img/dsf_scale_fs.png}\\
			&& L && a && b && X && Y \\
			&\multirow{3}{*}{\rotatebox{90}{\hspace{0.5cm} Deep superpixel features $F_e$ \hspace{-0.4cm}}}& 
			\includegraphics[width=\jjj,height=\ddd]{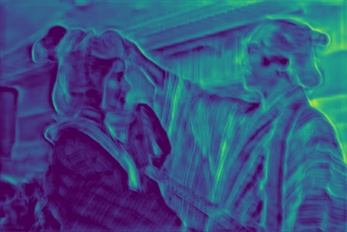}&
			\includegraphics[height=\ddd]{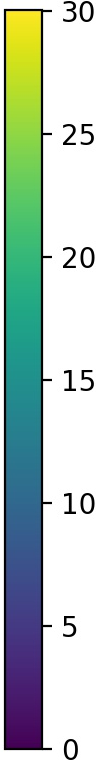}&
			\includegraphics[width=\jjj,height=\ddd]{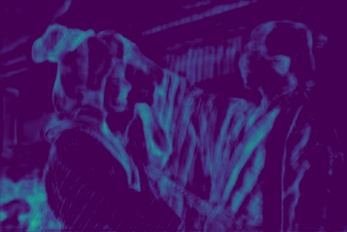}&
			\includegraphics[height=\ddd]{img/dsf_scale_fe.png}&
			\includegraphics[width=\jjj,height=\ddd]{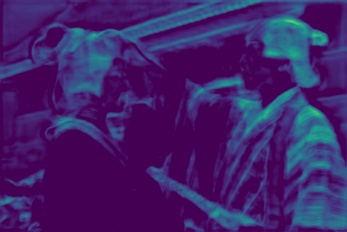}&
			\includegraphics[height=\ddd]{img/dsf_scale_fe.png}&
			\includegraphics[width=\jjj,height=\ddd]{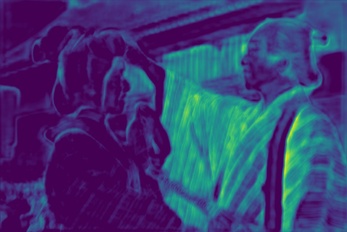}&
			\includegraphics[height=\ddd]{img/dsf_scale_fe.png}&
			\includegraphics[width=\jjj,height=\ddd]{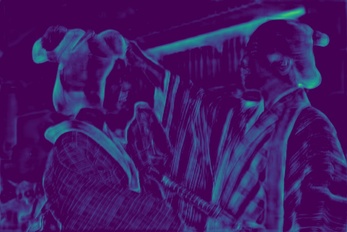}&
			\includegraphics[height=\ddd]{img/dsf_scale_fe.png}\\
			&&\includegraphics[width=\jjj,height=\ddd]{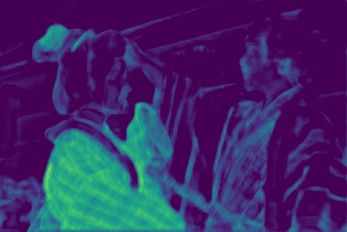}&
			\includegraphics[height=\ddd]{img/dsf_scale_fe.png}&
			\includegraphics[width=\jjj,height=\ddd]{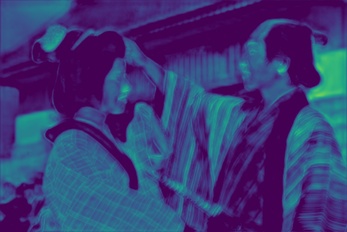}&
			\includegraphics[height=\ddd]{img/dsf_scale_fe.png}&
			\includegraphics[width=\jjj,height=\ddd]{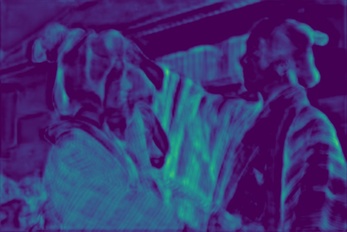}&
			\includegraphics[height=\ddd]{img/dsf_scale_fe.png}&
			\includegraphics[width=\jjj,height=\ddd]{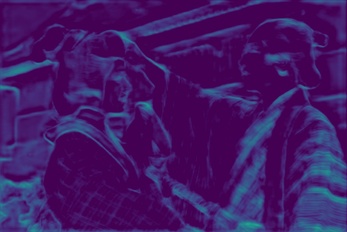}&
			\includegraphics[height=\ddd]{img/dsf_scale_fe.png}&
			\includegraphics[width=\jjj,height=\ddd]{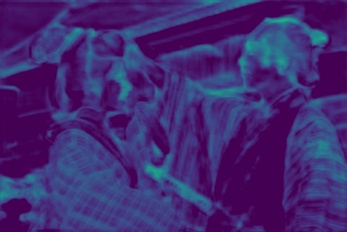}&
			\includegraphics[height=\ddd]{img/dsf_scale_fe.png}\\
			&&\includegraphics[width=\jjj,height=\ddd]{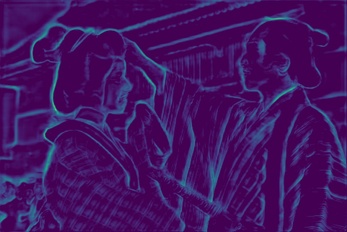}&
			\includegraphics[height=\ddd]{img/dsf_scale_fe.png}&
			\includegraphics[width=\jjj,height=\ddd]{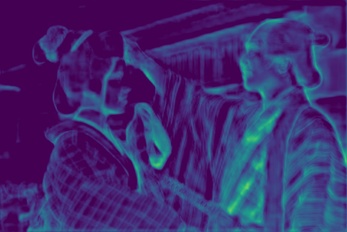}&
			\includegraphics[height=\ddd]{img/dsf_scale_fe.png}&
			\includegraphics[width=\jjj,height=\ddd]{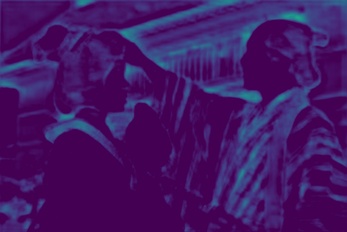}&
			\includegraphics[height=\ddd]{img/dsf_scale_fe.png}&
			\includegraphics[width=\jjj,height=\ddd]{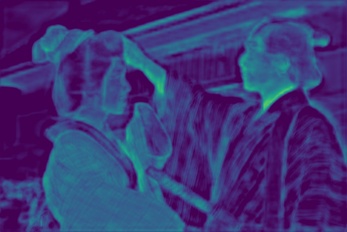}&
			\includegraphics[height=\ddd]{img/dsf_scale_fe.png}&
			\includegraphics[width=\jjj,height=\ddd]{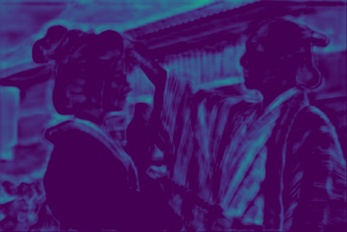}&
			\includegraphics[height=\ddd]{img/dsf_scale_fe.png}\\
	\end{tabular}}%
	\caption{\textbf{Visualization of deep superpixel features from the CNN encoder of SPAM.}
		Color $F_c=\lambda_c\times[L,a,b]$ and spatial $F_s=\lambda_s\times[X,Y]$ inputs are concatenated to the learnable 
		deep superpixel features $F_e$, which are activated in different image regions.
		Note that spatial coordinates $[X,Y]$ are normalized as in \cite{jampani2018superpixel}
		to compensate for different aspect ratios and requested superpixel scale.
	}
	\label{fig:deep_features}
\end{figure*}

\begin{figure*}[h]
	\centering
	\newcommand{\heee}{0.16\textwidth}
	{\scriptsize
		\begin{tabular}{@{\hspace{0mm}}c@{\hspace{2mm}}c@{\hspace{2mm}}c@{\hspace{2mm}}c@{\hspace{0mm}}}
			\includegraphics[width=0.24\textwidth,height=\heee]{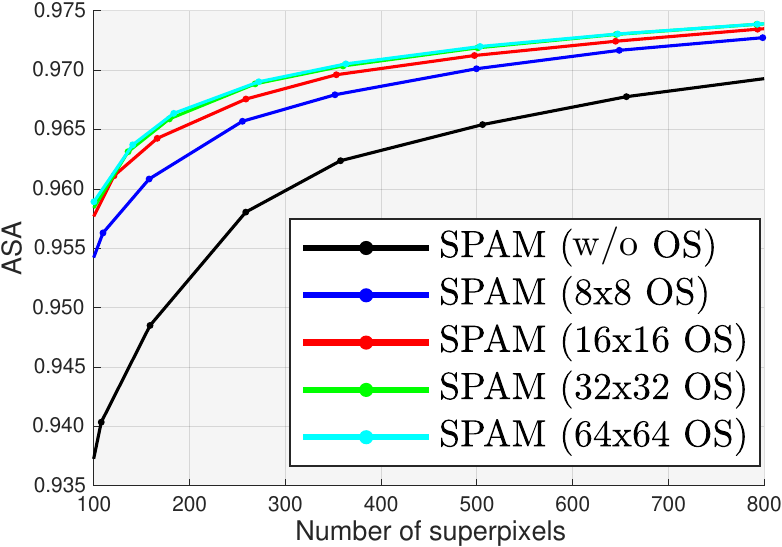}&
			\includegraphics[width=0.24\textwidth,height=\heee]{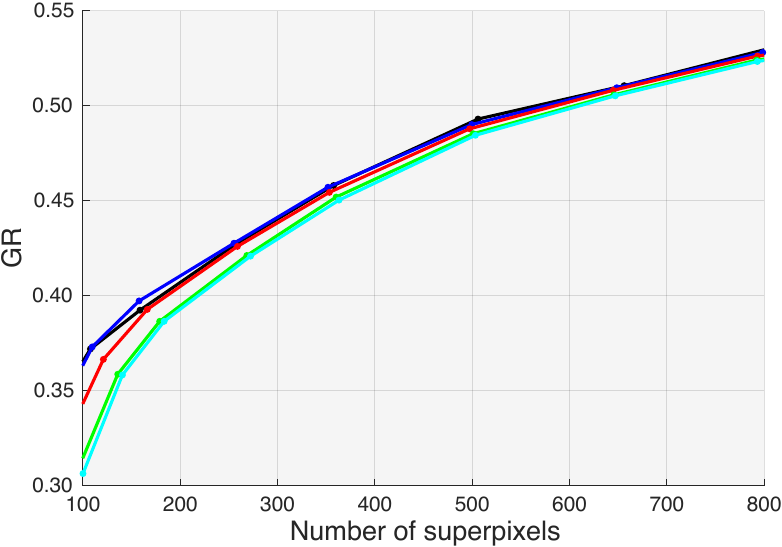}&
			\includegraphics[width=0.20\textwidth,height=\heee]{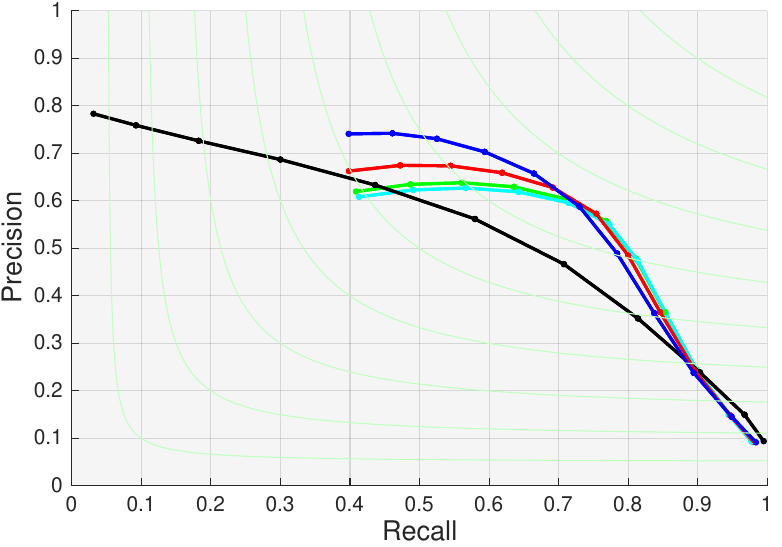}&
			\includegraphics[width=0.24\textwidth,height=\heee]{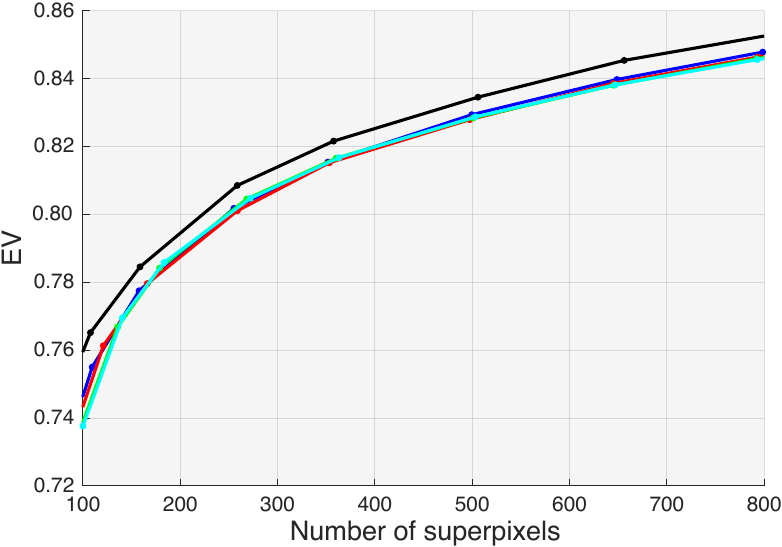}\\ 
	\end{tabular}}%
	\caption{\textbf{Impact of the number of grid points in SAM object segmentation (OS).} 
		SPAM segmentation accuracy increases, reaching a plateau for 32x32 while regularity and precision in contour detection slightly decrease due to more irregular objects detected.}
	\label{fig:ablation_sam_results}
\end{figure*}

\begin{table*}[p]
	\caption{Performance of SPAM on the BSD dataset for $K=400$ superpixels. We report results on superpixel metrics: ASA($\uparrow$), GR($\uparrow$), F($\uparrow$), and
		inference times.
	}
	\centering
	\begin{tabular}{|l|c|c|c|c|c|}
		\hline
		&\multicolumn{3}{c|}{Superpixel metrics}&\multicolumn{2}{c|}{Inference times} \\ 
		\cline{2-6}
		Method & ASA & GR & F & Prior seg. & Clustering  \\ \hline
		SPAM w/ FastSAM \cite{zhao2023fast} &0.9685  & {0.462} & 0.643& \textbf{$\approx$50ms}& 190ms \\ \hline
		SPAM w/ SAM \cite{kirillov23sam} & {0.9708} & {0.461} & {0.652} & $\approx$2.5s& 190ms \\ \hline
	\end{tabular}
	\label{tab:fastsam}
\end{table*}

\section{Additional results}\label{sec:qualitative}
\subsection{ Quantitative results}

\textbf{Comparison to state-of-the-art methods.} 
Table \ref{tab:evaluation} corresponds to Figure 6 in the main paper, focusing on the specific case where the number of requested superpixels is set to K = 400.
Note how SPAM gets the best performance, 
with the best results in terms of ASA and F on all datasets, compared to all methods.
Compared to DL-based methods, SPAM greatly improves and ranks first in terms of GR, while achieving on average second best results on EV and $\Delta$K.
No previous method is able to perform well on multiple criteria, especially being both accurate (ASA) and regular (GR).

\smallskip

\noindent\textbf{Semantic segmentation refinement.}
As an additional downstream evaluation, we demonstrate that SPAM can enhance the performance of semantic segmentation methods, particularly in the challenging task of maintaining accuracy around object borders. Although superpixels do not explicitly carry semantic information, they may be efficient at segmenting local image content.

\begin{figure*}[t!]
	\centering
	\newcommand{\www}{0.27\textwidth}
	\newcommand{\hhh}{0.19\textwidth}
	{\small \begin{tabular}{@{\hspace{0mm}}c@{\hspace{1mm}}c@{\hspace{1mm}}c@{\hspace{1mm}}c@{\hspace{0mm}}}
			\includegraphics[width=\www,height=\hhh]{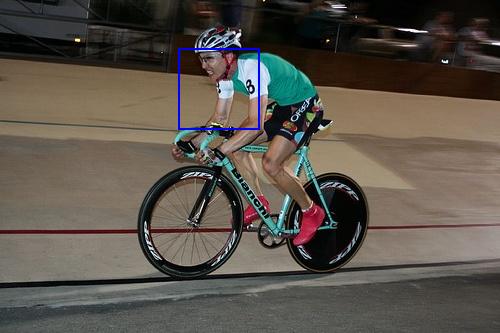} &
			\includegraphics[height=\hhh]{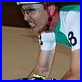} &
			\includegraphics[width=\www,height=\hhh]{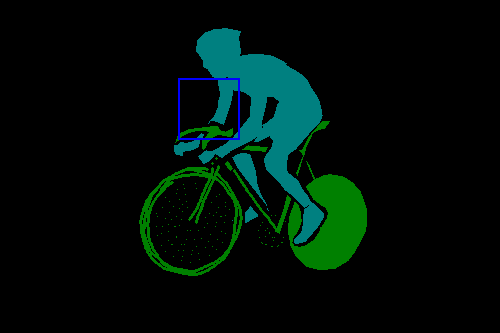} &
			\includegraphics[height=\hhh]{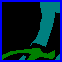} \\
			\multicolumn{2}{c}{(a) Image} & \multicolumn{2}{c}{(b) Groundtruth} \\[0.5ex]
			\includegraphics[width=\www,height=\hhh]{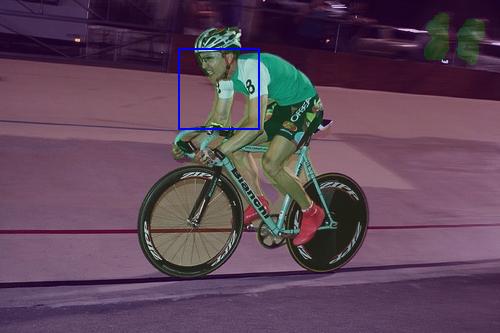} &
			\includegraphics[height=\hhh]{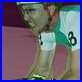} &
			\includegraphics[width=\www,height=\hhh]{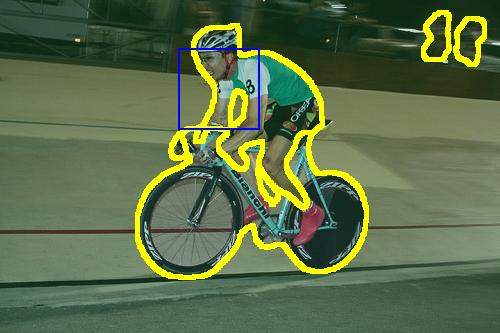} &
			\includegraphics[height=\hhh]{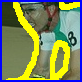} \\
			\multicolumn{2}{c}{(c) DeepLabV3 semantic segmentation} &
			\multicolumn{2}{c}{(d) Dilated object contours of (c)} \\[0.5ex]
			\includegraphics[width=\www,height=\hhh]{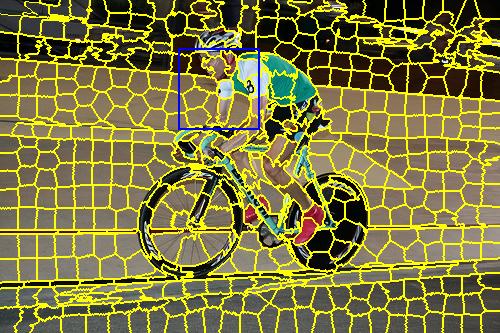} &
			\includegraphics[height=\hhh]{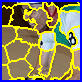} &
			\includegraphics[width=\www,height=\hhh]{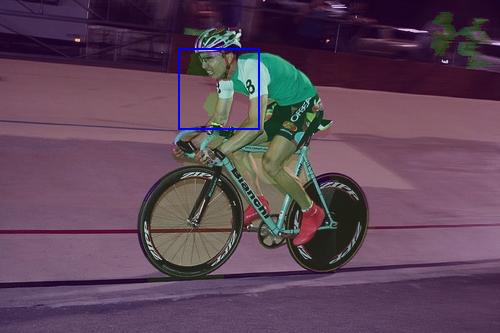} &
			\includegraphics[height=\hhh]{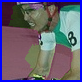} \\
			\multicolumn{2}{c}{(e) SLIC superpixels} &
			\multicolumn{2}{c}{(f) SLIC superpixels $\cap$ (c)} \\[0.5ex]
			\includegraphics[width=\www,height=\hhh]{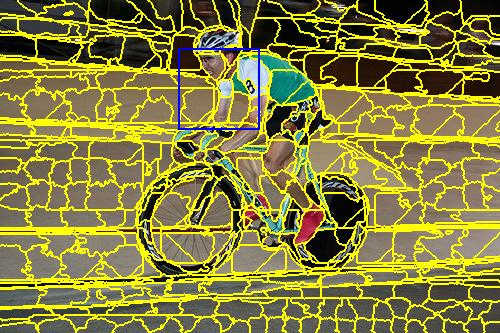} &
			\includegraphics[height=\hhh]{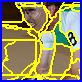} &
			\includegraphics[width=\www,height=\hhh]{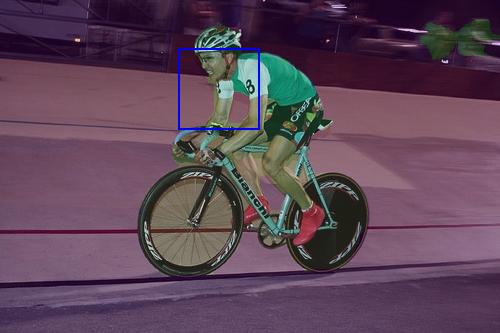} &
			\includegraphics[height=\hhh]{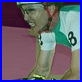} \\
			\multicolumn{2}{c}{(g) AINet superpixels} &
			\multicolumn{2}{c}{(h) AINet superpixels $\cap$ (c)} \\[0.5ex]
			\includegraphics[width=\www,height=\hhh]{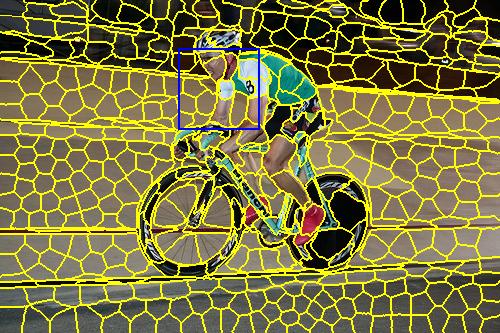} &
			\includegraphics[height=\hhh]{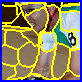} &
			\includegraphics[width=\www,height=\hhh]{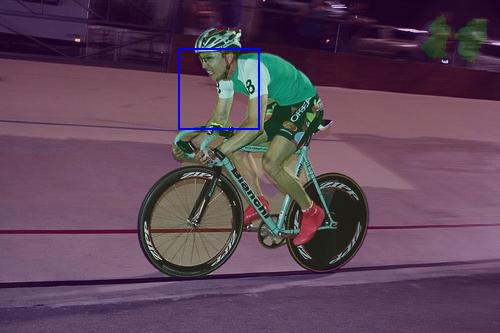} &
			\includegraphics[height=\hhh]{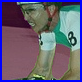} \\
			\multicolumn{2}{c}{(i) SPAM superpixels} &
			\multicolumn{2}{@{\hspace{0mm}}c@{\hspace{0mm}}}{(j) Refined semantic seg. w/ SPAM superpixels} \\
	\end{tabular}}%
	\caption{\textbf{Example of semantic segmentation refinement using superpixels on a PASCAL VOC2012 image.}
		A DeepLabV3~\cite{chen2017rethinking} architecture with a ResNet50 backbone is used to estimate a semantic segmentation (c), that may be inaccurate on the object borders. 
		SLIC~\cite{achanta2012} (e) or AINet~\cite{wang2021ainet} (g) superpixels cannot leverage this prior semantic map to generate an accurate segmentation. Therefore, their semantic overlaps (f)-(h), {i.e.} taking the majority labels in each superpixel, struggle to improve the segmentation.  
		To refine the semantic segmentation, we can dilate all object borders by a 5x5 kernel (d) 
		and consider these pixels as uncertain or unlabeled.
		This map is provided as input prior for SPAM that respects the object labels 
		and associate pixels in uncertainty regions to the closest similar superpixels (i).
		SPAM superpixels being associated to semantic labels, we directly obtain a refined semantic segmentation (j).
	}
	\label{fig:refinement}
\end{figure*}

As shown in Fig.~\ref{fig:refinement}, even when using a large-scale semantic segmentation model, the accuracy can be improved using superpixels to capture small objects or uncertain borders. SPAM is especially suited in this context, as it can take a semantic segmentation with unlabeled borders as prior and assign these pixels to superpixels within semantic regions. 

In Tab.~\ref{tab:refinement_supp}, we compare performance on PASCAL VOC2012~\cite{pascalvoc} validation set in terms of mean Intersection over Union (mIoU) and pixel-wise accuracy (ACC), using SPAM to refine DeepLabV3\cite{chen2017rethinking} segmentations with different backbones. We achieve this by dilating object borders of DeepLabV3 segmentations with a 5x5 kernel, marking these pixels as unlabeled, and then using this map as a prior for SPAM. We thus improve segmentation accuracy by locally refining object borders, whereas standard nearest-neighbor (NN) clustering on color features reduces performance. Other superpixel methods \cite{wang2021ainet,achanta2012} cannot leverage a prior map as input and instead simply project onto the segmentation and assign majority labels.

\subsection{Qualitative results}
\begin{figure*}[t!]
	\centering
	\newcommand{\www}{0.19\textwidth}
	{\footnotesize \begin{tabular}{@{\hspace{0mm}}c@{\hspace{1mm}}c@{\hspace{1mm}}c@{\hspace{1mm}}c@{\hspace{1mm}}c@{\hspace{0mm}}}
			\includegraphics[width=\www]{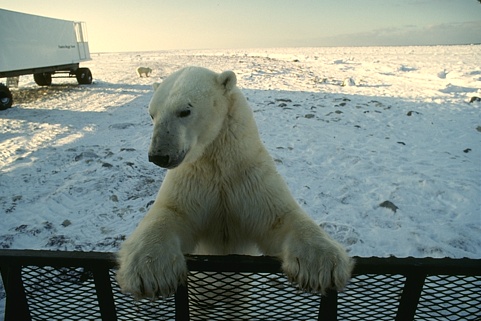}&
			\includegraphics[width=\www]{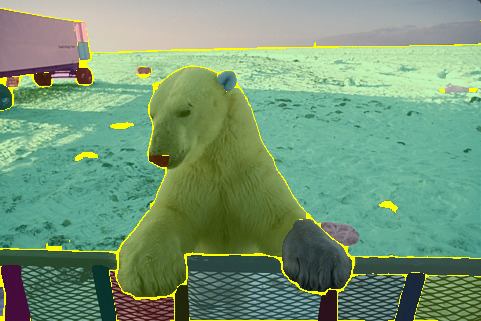} & 
			\includegraphics[width=\www]{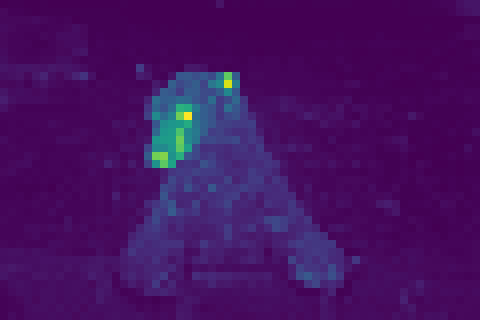} &
			\includegraphics[width=\www]{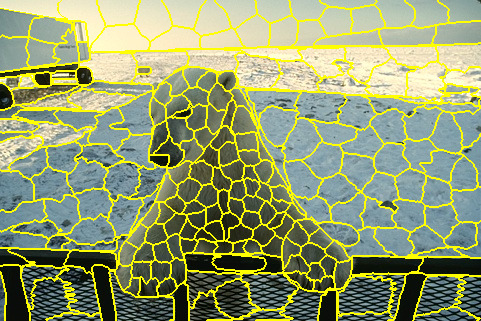}&
			\includegraphics[width=\www]{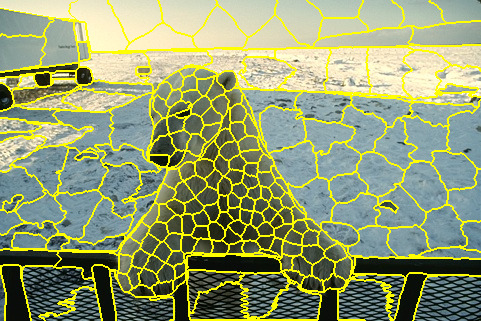}  \\
			\includegraphics[width=\www]{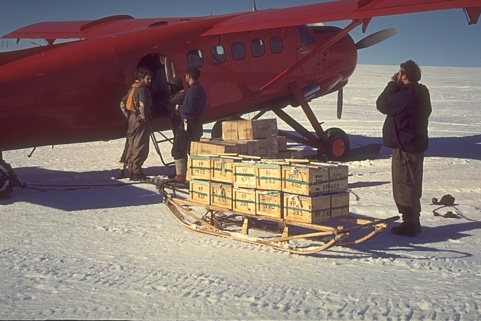}&
			\includegraphics[width=\www]{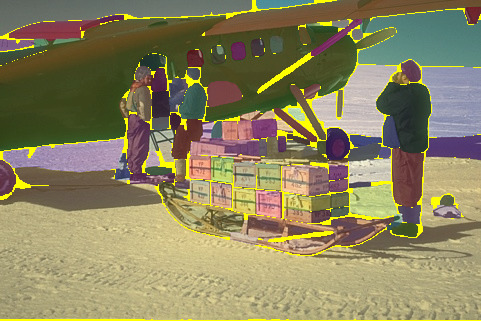} & 
			\includegraphics[width=\www]{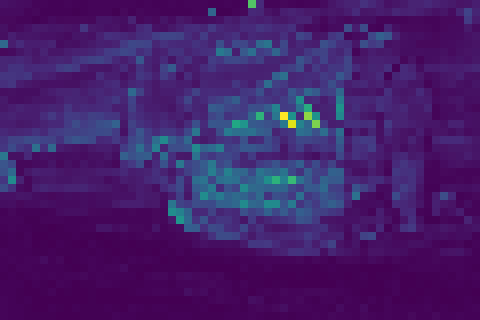} &
			\includegraphics[width=\www]{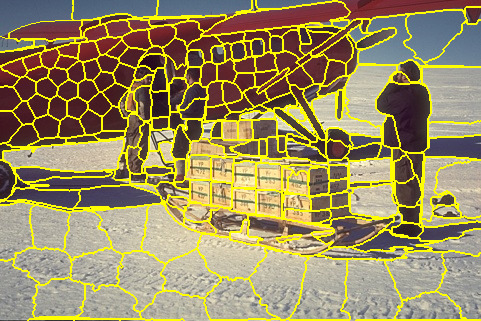}&
			\includegraphics[width=\www]{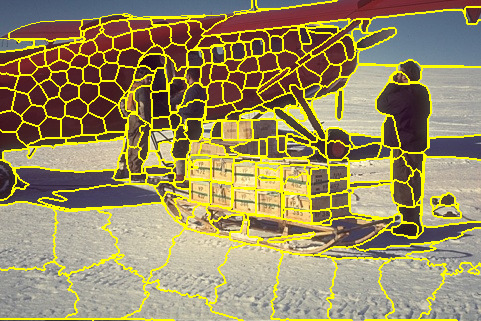}  \\
			(a) Image & (b) Object map & (c) Saliency map & (d) SPAM (VA $r=2$) & (e) SPAM (VA $r=3$)\\[-1ex]
	\end{tabular}}%
	\caption{\textbf{SPAM visual attention (VA) adaptive segmentation mode.}  
		The results are generated for $K=250$ superpixels and several scale factors $r$, which increase the number of superpixels in objects of interest detected by the saliency map.
		Note that the number of superpixels is decreased in other regions to maintain the requested number of superpixels $K$.
	}
	\label{fig:res_quali_SPAM_mode_va}
\end{figure*}

\noindent\textbf{SPAM superpixel segmentation.}
We show segmentation results for the two SPAM adaptive modes that generate superpixels of varying sizes within objects.
Fig.~\ref{fig:res_quali_SPAM_mode_va} illustrates results for the visual attention (VA) mode.
In this mode, we use DINO~\cite{caron2021emerging}
to compute a visual saliency map and identify regions of interest.
This map is intersected with the SAM objects to identify objects of interest and increase the number of superpixels within them.

In Fig.~\ref{fig:res_quali_SPAM_mode_user}, we show segmentations obtained from user interaction.
A user can also manually select different factors for each object to increase or decrease the superpixel scale, providing valuable results for annotation purposes.

\begin{figure*}[t!]
	\centering
	\newcommand{\hhh}{0.22\textwidth}
	{\footnotesize \begin{tabular}{@{\hspace{0mm}}c@{\hspace{1mm}}c@{\hspace{1mm}}c@{\hspace{1mm}}c@{\hspace{0mm}}}
			\includegraphics[width=\hhh]{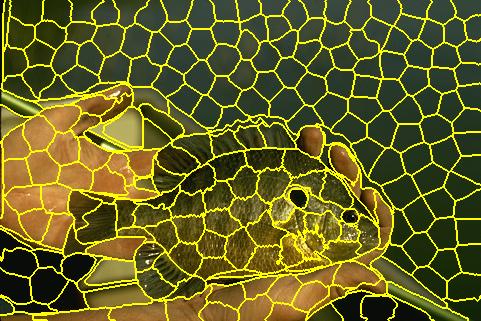}&
			\includegraphics[width=\hhh]{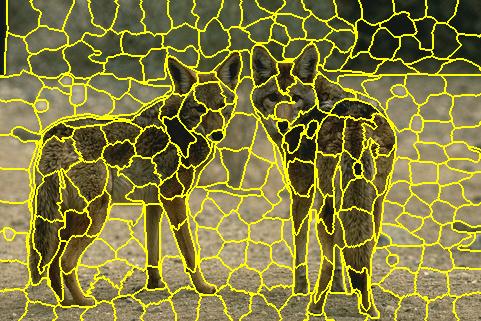}&
			\includegraphics[width=\hhh]{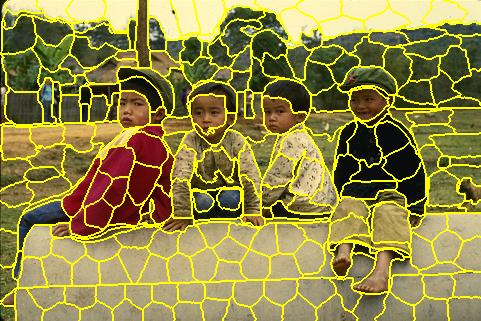}&
			\includegraphics[width=\hhh]{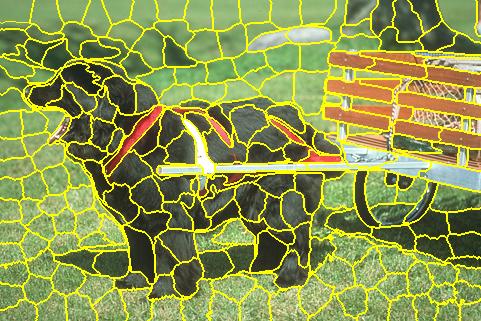}\\
			\includegraphics[width=\hhh]{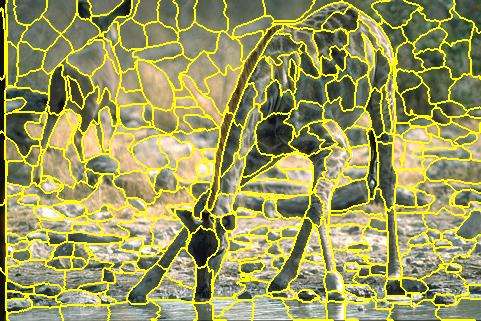}&
			\includegraphics[width=\hhh]{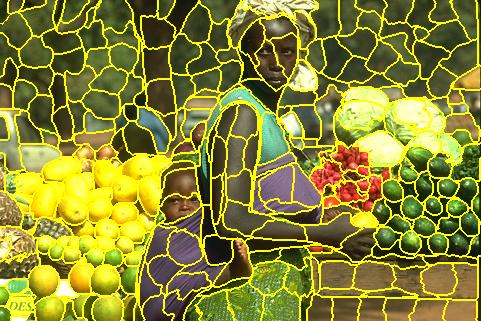}&
			\includegraphics[width=\hhh]{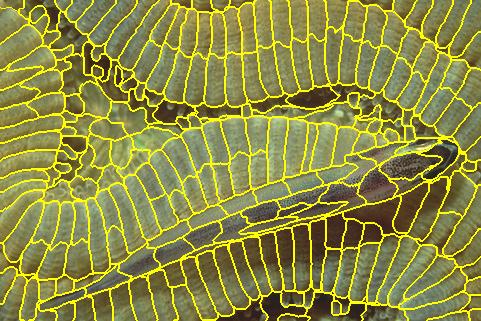}&
			\includegraphics[width=\hhh]{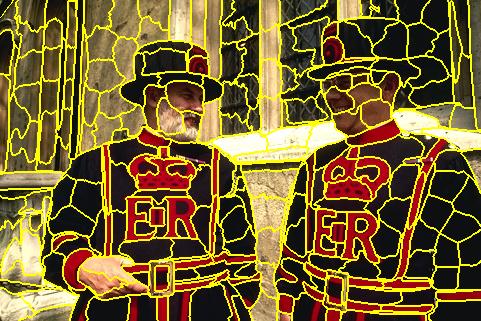}\\
			\includegraphics[width=\hhh]{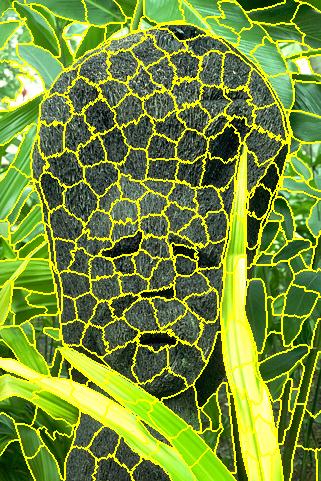}&
			\includegraphics[width=\hhh]{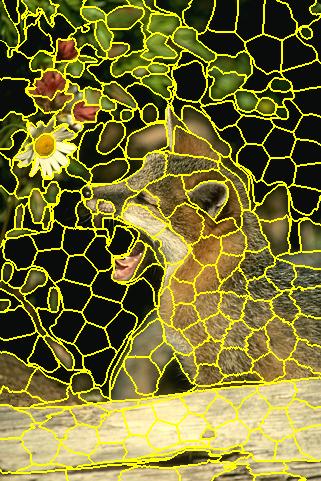}&
			\includegraphics[width=\hhh]{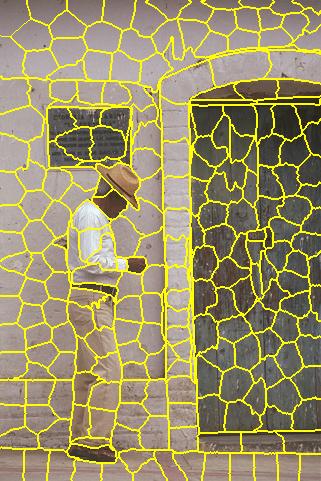}&
			\includegraphics[width=\hhh]{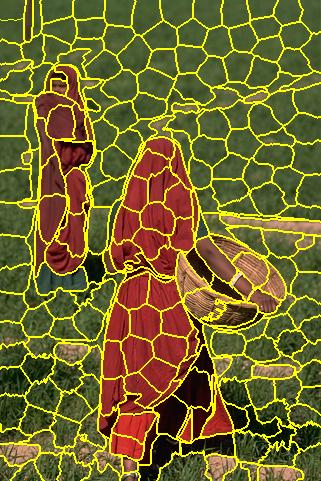}\\[-1ex]
	\end{tabular}}%
	\caption{\textbf{Qualitative results of SPAM segmentation on BSD test images.}
		The results are generated for $K=250$ superpixels.
		See how SPAM generates highly accurate yet regular superpixels.
	}
	\label{fig:res_quali_SPAM}
\end{figure*}

Finally, in Fig.~\ref{fig:res_quali_SPAM} we present several examples of SPAM superpixel segmentation with its default settings. 

\begin{figure*}[t!]
	\centering
	\newcommand{\hhh}{0.16\textwidth}
	\newcommand{\hee}{0.1065\textwidth}
	{\footnotesize \begin{tabular}{@{\hspace{0mm}}c@{\hspace{1mm}}c@{\hspace{1mm}}c@{\hspace{1mm}}c@{\hspace{1mm}}c@{\hspace{1mm}}c@{\hspace{1mm}}@{\hspace{0mm}}}
			\includegraphics[width=\hhh,height=\hee]{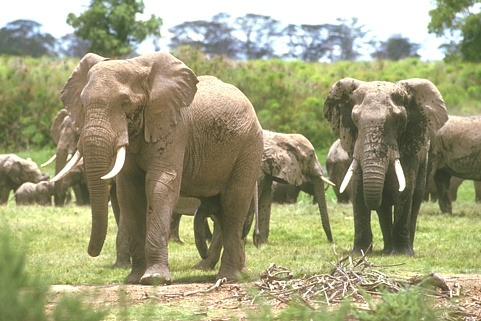}&
			\includegraphics[width=\hhh,height=\hee]{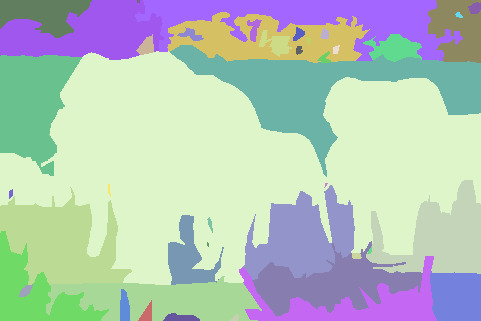}&
			\includegraphics[width=\hhh,height=\hee]{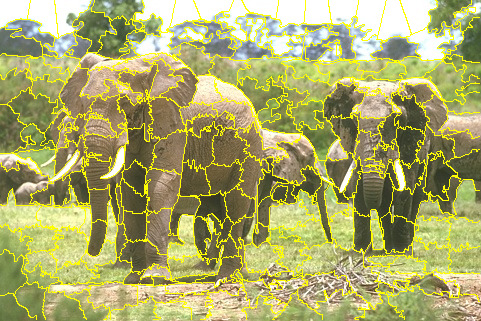}&
			\includegraphics[width=\hhh,height=\hee]{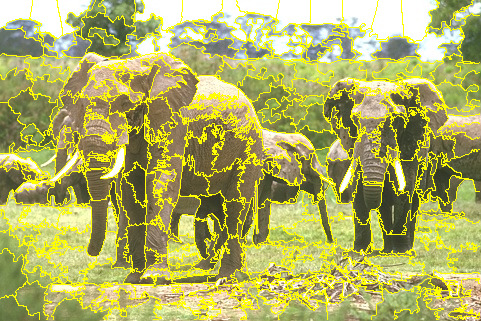}&
			\includegraphics[width=\hhh,height=\hee]{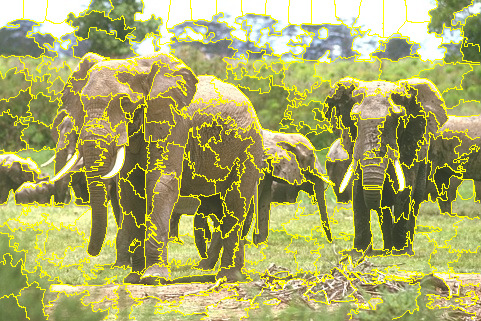}&
			\includegraphics[width=\hhh,height=\hee]{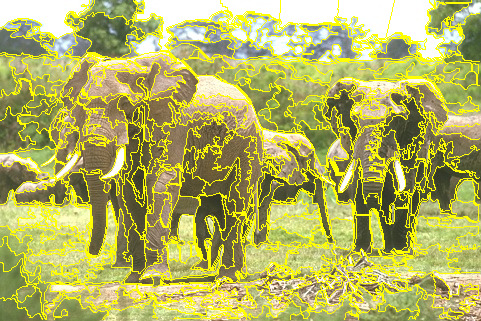}\\
			Image & Groundtruth & SLIC \cite{achanta2012} & LSC \cite{li2015} & SNIC \cite{achanta2017superpixels}&SCAC \cite{yuan2021superpixels}\\[0.5ex]
			\includegraphics[width=\hhh,height=\hee]{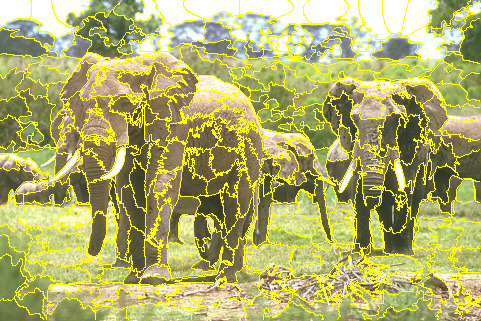}&
			\includegraphics[width=\hhh,height=\hee]{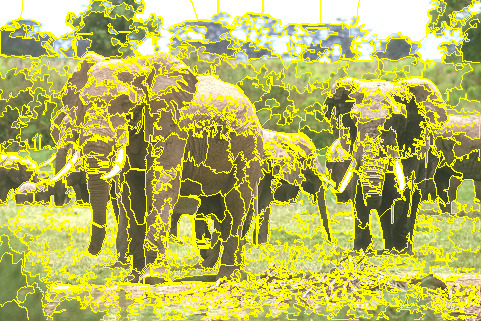}&
			\includegraphics[width=\hhh,height=\hee]{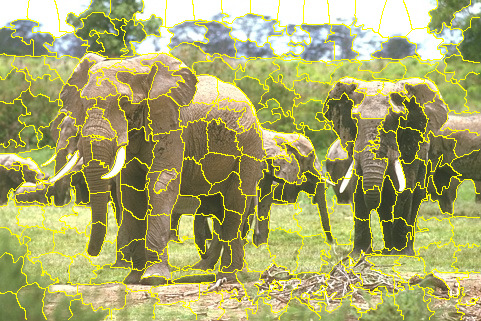}&
			\includegraphics[width=\hhh,height=\hee]{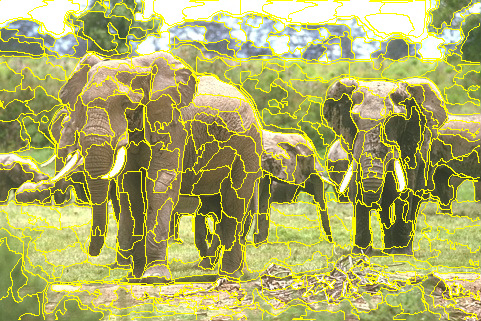}&
			\includegraphics[width=\hhh,height=\hee]{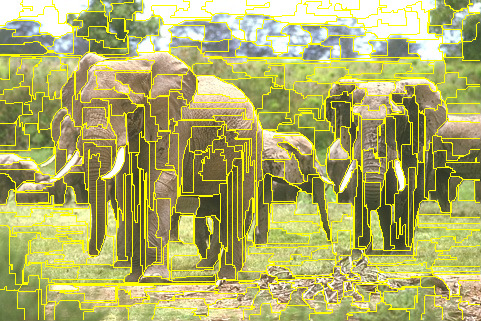}&
			\includegraphics[width=\hhh,height=\hee]{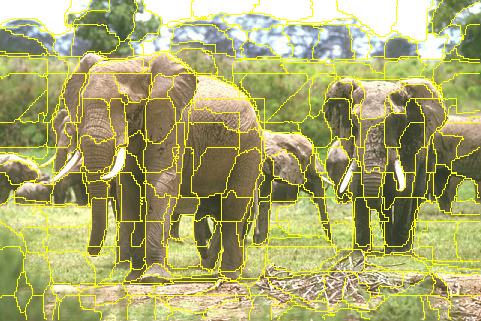}\\
			GMMSP \cite{Ban18} & BASS \cite{Uziel:ICCV:2019:BASS} & FSLIC \cite{wu2020fuzzy} & SSN \cite{jampani2018superpixel}&
			SEAL \cite{tu2018learning} & SFCN \cite{yang2020superpixel}\\[0.5ex]
			\includegraphics[width=\hhh,height=\hee]{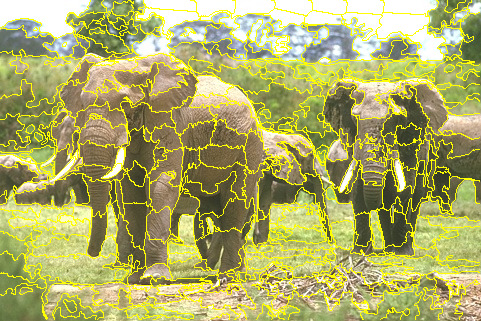}&
			\includegraphics[width=\hhh,height=\hee]{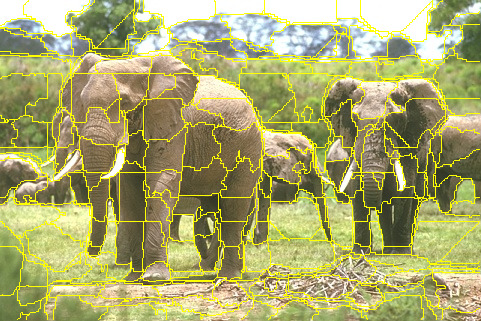}&
			\includegraphics[width=\hhh,height=\hee]{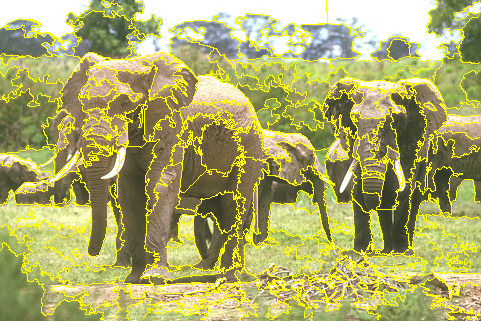}&
			\includegraphics[width=\hhh,height=\hee]{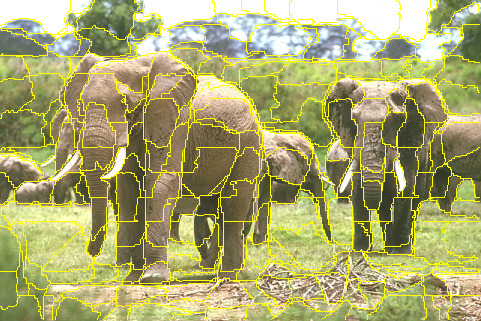}&
			\includegraphics[width=\hhh,height=\hee]{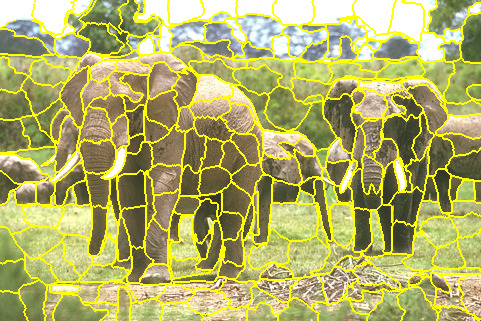}&
			\includegraphics[width=\hhh,height=\hee]{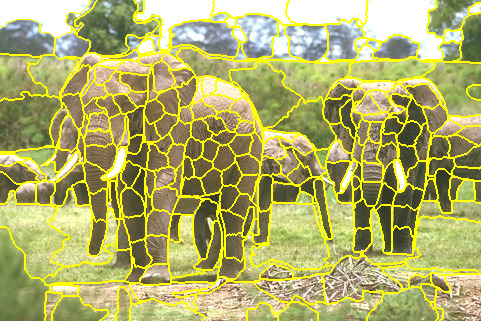}\\
			LNS-Net \cite{zhu2021learning} & AINet \cite{wang2021ainet} & VSSS \cite{zhou2023vine} & CDS \cite{xu2024learning} & SPAM & SPAM (VA $r$=2) \\[0.5ex]
			\includegraphics[width=\hhh,height=\hee]{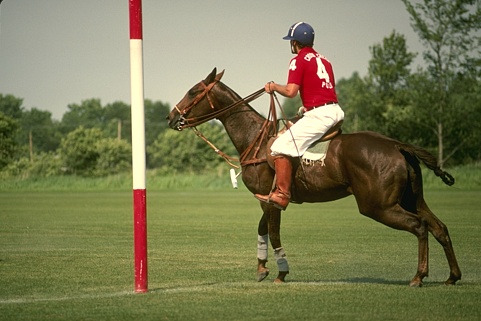}&
			\includegraphics[width=\hhh,height=\hee]{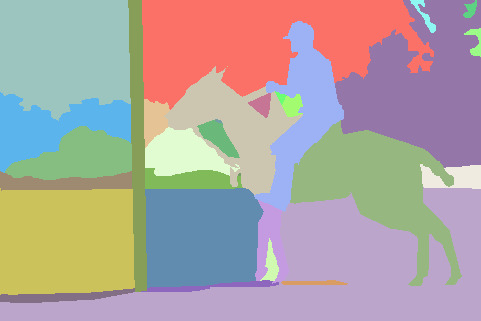}&
			\includegraphics[width=\hhh,height=\hee]{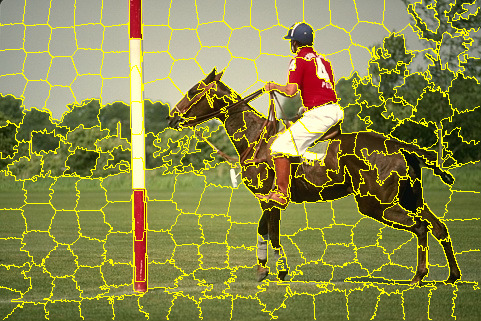}&
			\includegraphics[width=\hhh,height=\hee]{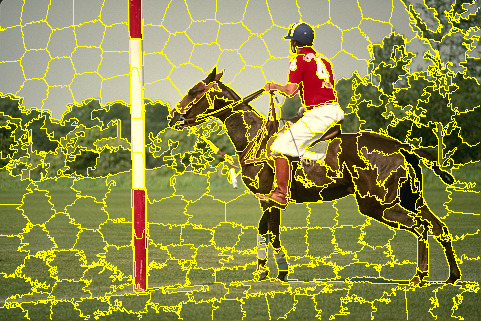}&
			\includegraphics[width=\hhh,height=\hee]{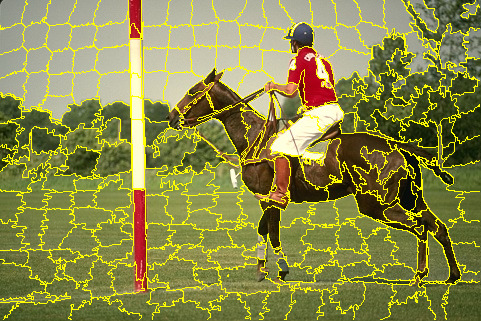}&
			\includegraphics[width=\hhh,height=\hee]{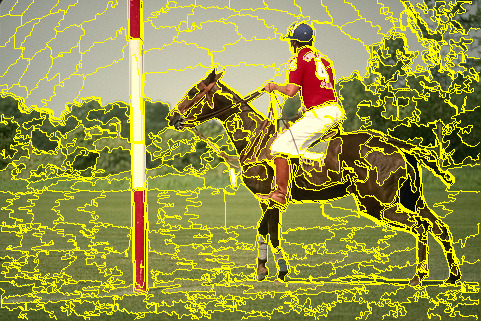}\\
			Image & Groundtruth & SLIC \cite{achanta2012} & LSC \cite{li2015} & SNIC \cite{achanta2017superpixels}&SCAC \cite{yuan2021superpixels}\\[0.5ex]
			\includegraphics[width=\hhh,height=\hee]{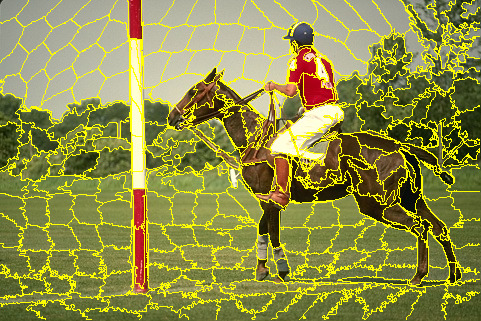}&
			\includegraphics[width=\hhh,height=\hee]{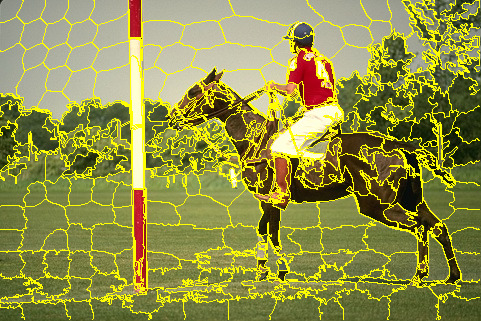}&
			\includegraphics[width=\hhh,height=\hee]{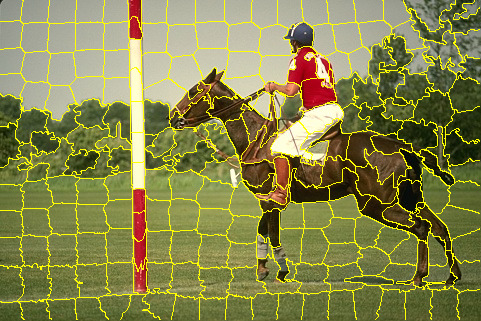}&
			\includegraphics[width=\hhh,height=\hee]{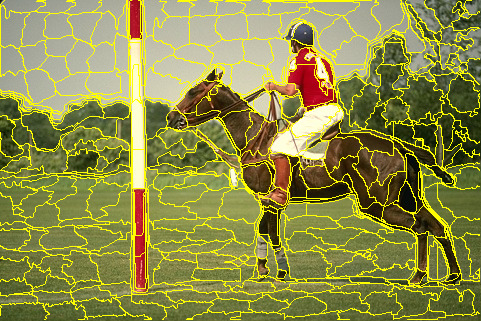}&
			\includegraphics[width=\hhh,height=\hee]{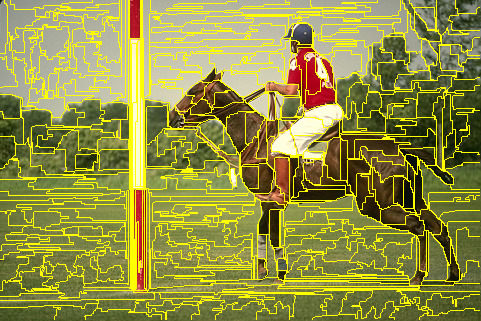}&
			\includegraphics[width=\hhh,height=\hee]{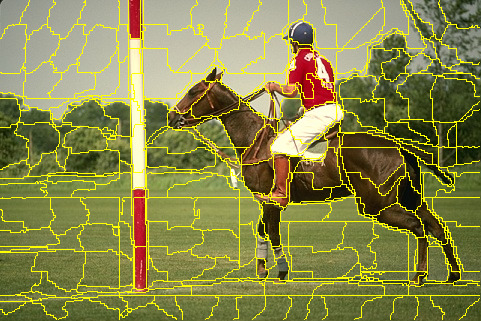}\\
			GMMSP \cite{Ban18} & BASS \cite{Uziel:ICCV:2019:BASS} & FSLIC \cite{wu2020fuzzy} & SSN \cite{jampani2018superpixel}&
			SEAL \cite{tu2018learning} & SFCN \cite{yang2020superpixel}\\[0.5ex]
			\includegraphics[width=\hhh,height=\hee]{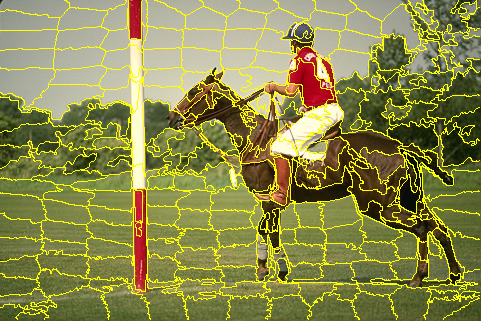}&
			\includegraphics[width=\hhh,height=\hee]{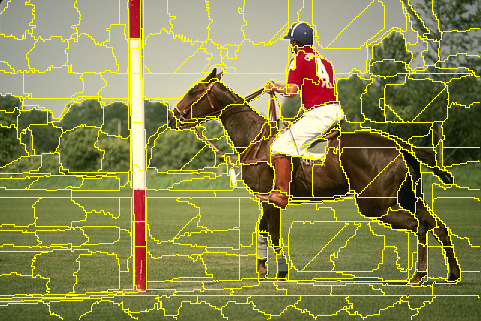}&
			\includegraphics[width=\hhh,height=\hee]{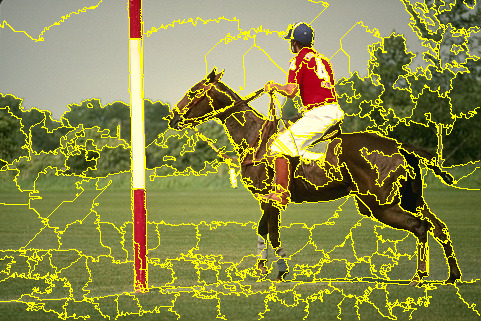}&
			\includegraphics[width=\hhh,height=\hee]{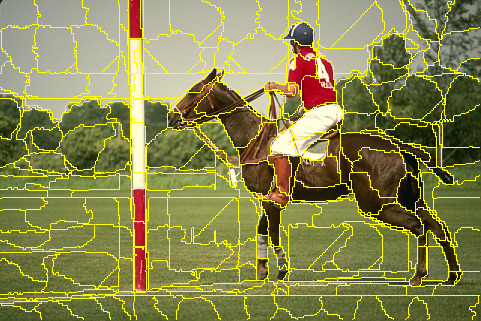}&
			\includegraphics[width=\hhh,height=\hee]{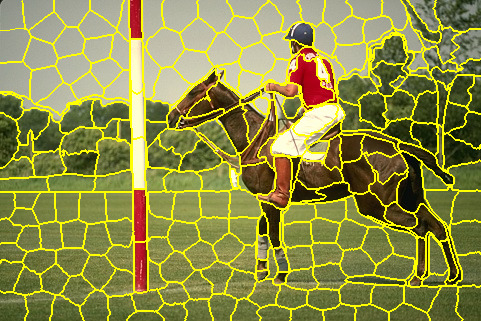}&
			\includegraphics[width=\hhh,height=\hee]{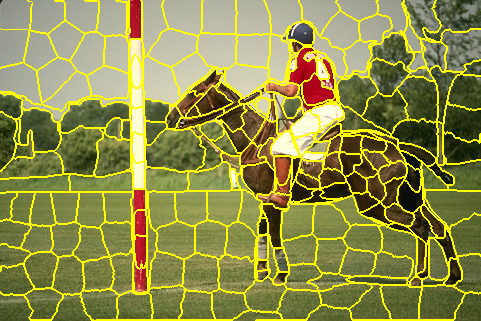}\\
			LNS-Net \cite{zhu2021learning} & AINet \cite{wang2021ainet} & VSSS \cite{zhou2023vine} & CDS \cite{xu2024learning} & SPAM & SPAM (VA $r$=2) \\[-0.25ex] 
			\includegraphics[width=\hhh,height=\hee]{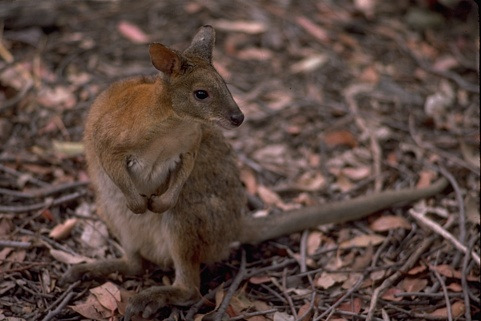}&
			\includegraphics[width=\hhh,height=\hee]{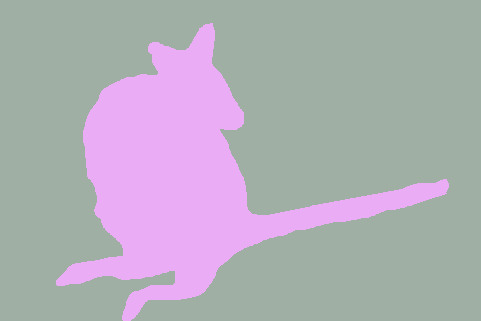}&
			\includegraphics[width=\hhh,height=\hee]{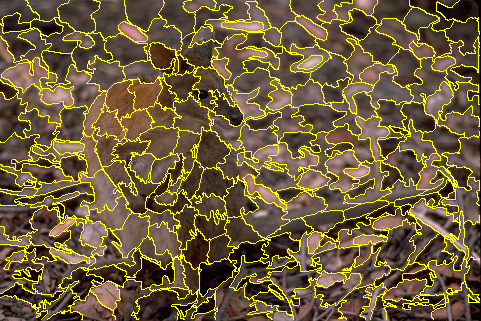}&
			\includegraphics[width=\hhh,height=\hee]{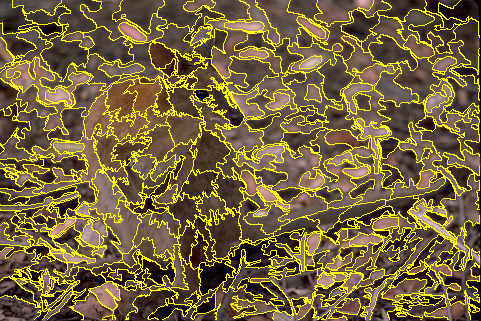}&
			\includegraphics[width=\hhh,height=\hee]{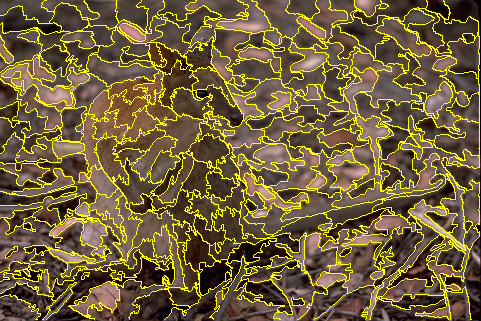}&
			\includegraphics[width=\hhh,height=\hee]{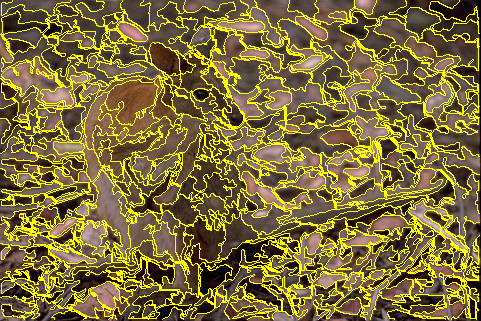}\\
			Image & Groundtruth & SLIC \cite{achanta2012} & LSC \cite{li2015} & SNIC \cite{achanta2017superpixels}&SCAC \cite{yuan2021superpixels}\\[0.5ex]
			\includegraphics[width=\hhh,height=\hee]{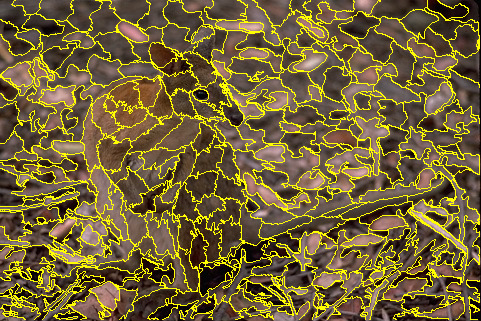}&
			\includegraphics[width=\hhh,height=\hee]{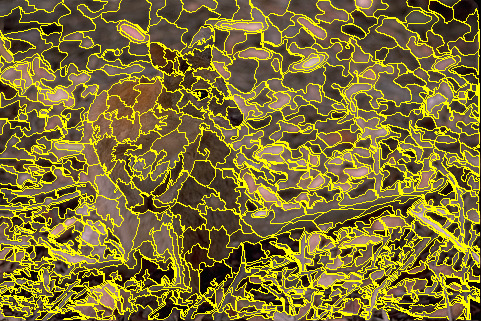}&
			\includegraphics[width=\hhh,height=\hee]{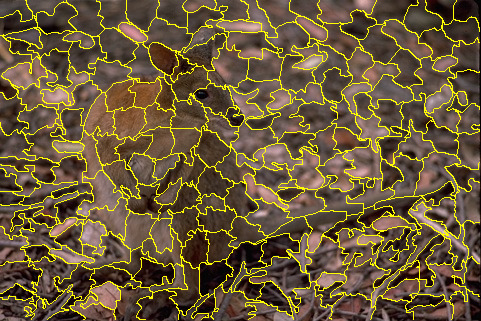}&
			\includegraphics[width=\hhh,height=\hee]{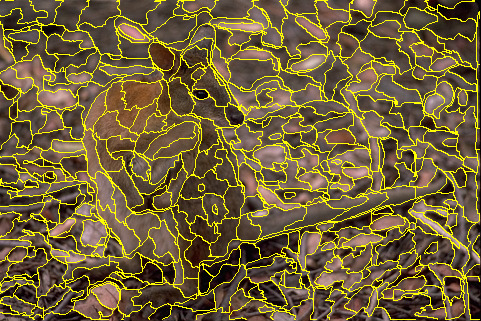}&
			\includegraphics[width=\hhh,height=\hee]{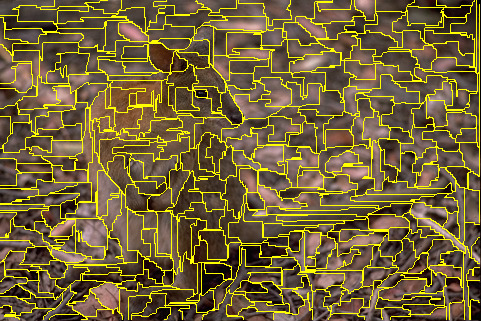}&
			\includegraphics[width=\hhh,height=\hee]{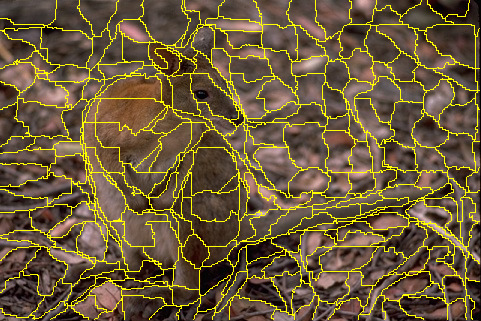}\\
			GMMSP \cite{Ban18} & BASS \cite{Uziel:ICCV:2019:BASS} & FSLIC \cite{wu2020fuzzy} & SSN \cite{jampani2018superpixel}&
			SEAL \cite{tu2018learning} & SFCN \cite{yang2020superpixel}\\[0.5ex]
			\includegraphics[width=\hhh,height=\hee]{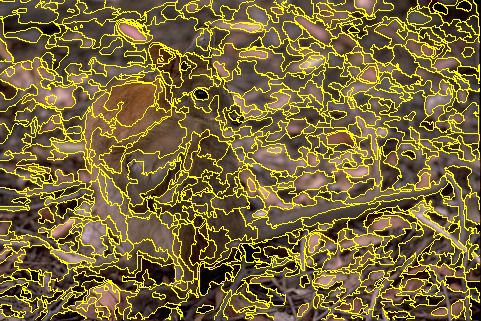}&
			\includegraphics[width=\hhh,height=\hee]{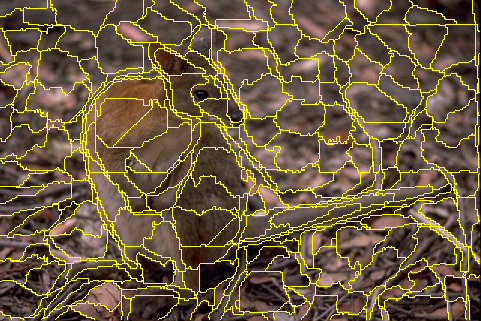}&
			\includegraphics[width=\hhh,height=\hee]{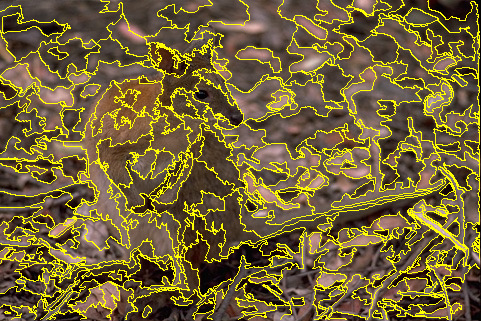}&
			\includegraphics[width=\hhh,height=\hee]{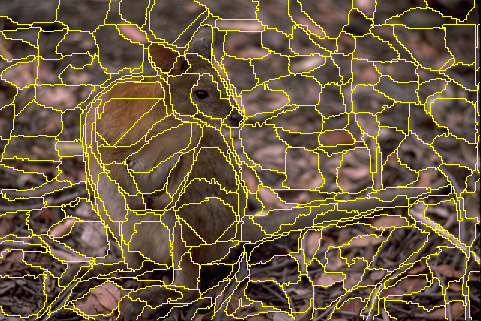}&
			\includegraphics[width=\hhh,height=\hee]{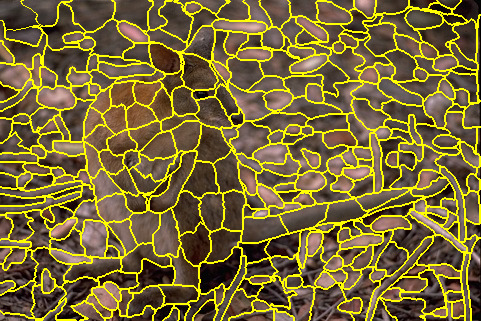}&
			\includegraphics[width=\hhh,height=\hee]{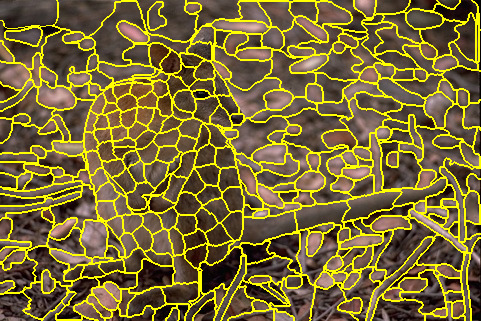}\\
			LNS-Net \cite{zhu2021learning} & AINet \cite{wang2021ainet} & VSSS \cite{zhou2023vine} & CDS \cite{xu2024learning} & SPAM & SPAM (VA $r$=2) \\
	\end{tabular}}%
	\caption{\textbf{Qualitative evaluation of superpixel methods on the BSD test set.}
		The number of requested superpixels is set to $K=250$. 
		SPAM generates superpixels that are both more accurate and considerably more regular than other state-of-the-art 
		DL-based methods.}
	\label{fig:res_quali_BSD}
\end{figure*}

\begin{figure*}[t!]
	\centering
	\newcommand{\hhh}{0.16\textwidth}
	\newcommand{\hee}{0.1065\textwidth}
	{\footnotesize \begin{tabular}{@{\hspace{0mm}}c@{\hspace{1mm}}c@{\hspace{1mm}}c@{\hspace{1mm}}c@{\hspace{1mm}}c@{\hspace{1mm}}c@{\hspace{1mm}}@{\hspace{0mm}}}
			\includegraphics[width=\hhh,height=\hee]{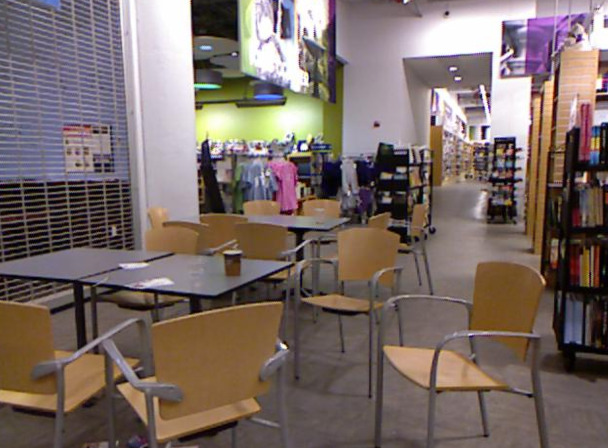}&
			\includegraphics[width=\hhh,height=\hee]{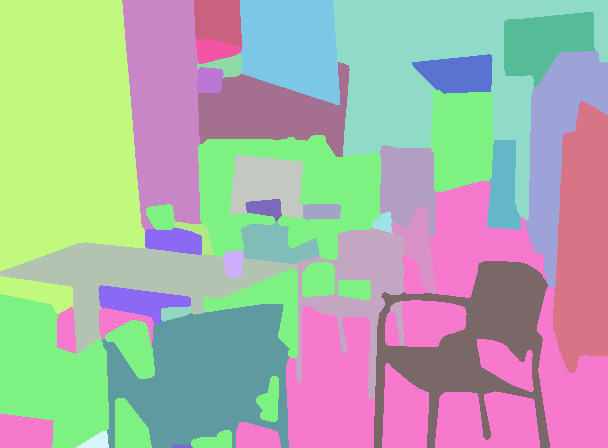}&
			\includegraphics[width=\hhh,height=\hee]{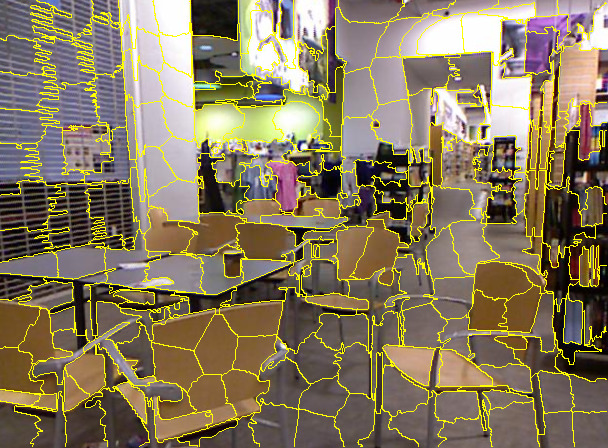}&
			\includegraphics[width=\hhh,height=\hee]{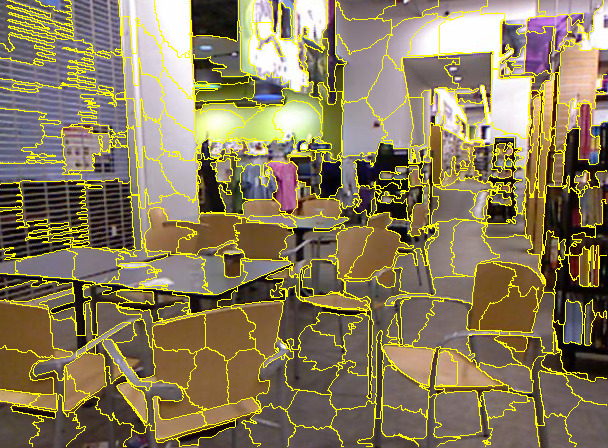}&
			\includegraphics[width=\hhh,height=\hee]{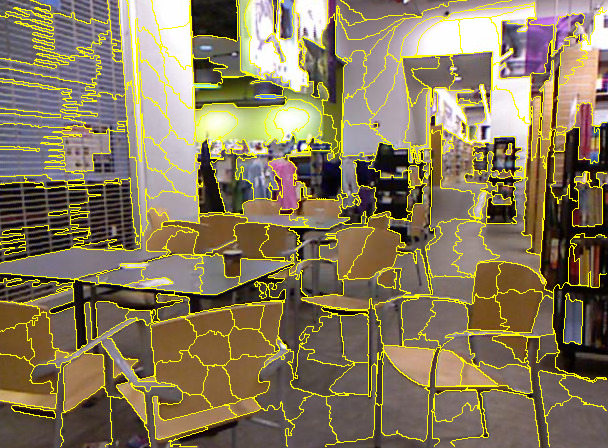}&
			\includegraphics[width=\hhh,height=\hee]{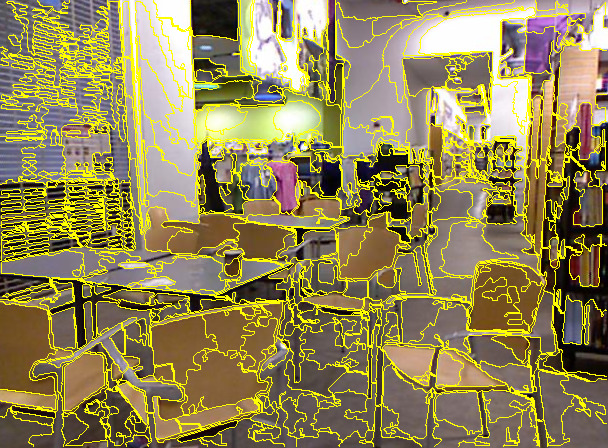}\\
			Image & Groundtruth & SLIC \cite{achanta2012} & LSC \cite{li2015} & SNIC \cite{achanta2017superpixels}&SCAC \cite{yuan2021superpixels}\\[0.5ex]
			\includegraphics[width=\hhh,height=\hee]{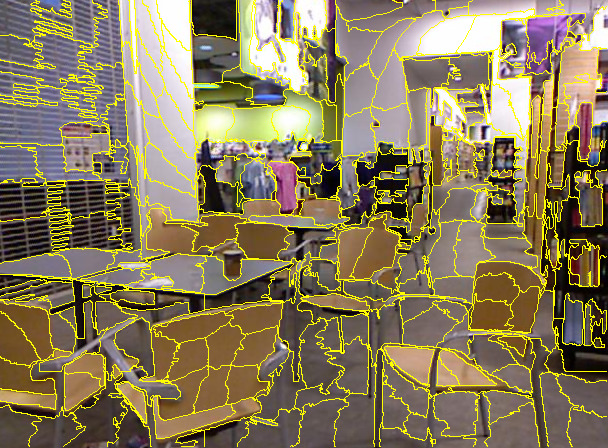}&
			\includegraphics[width=\hhh,height=\hee]{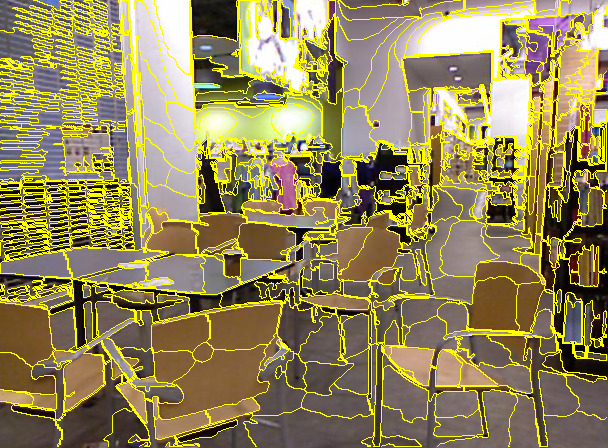}&
			\includegraphics[width=\hhh,height=\hee]{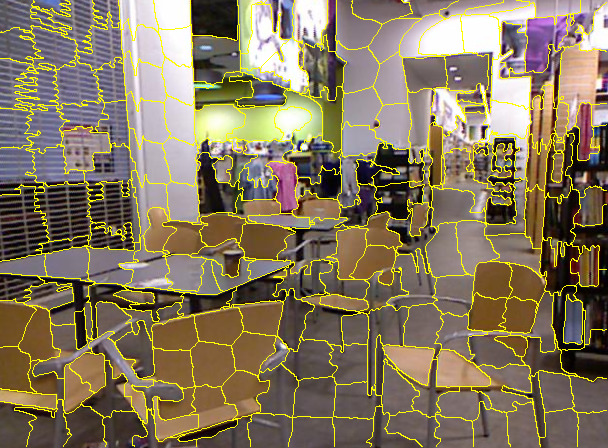}&
			\includegraphics[width=\hhh,height=\hee]{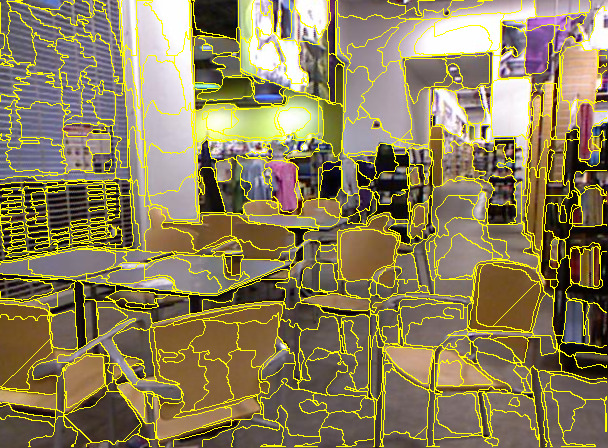}&
			\includegraphics[width=\hhh,height=\hee]{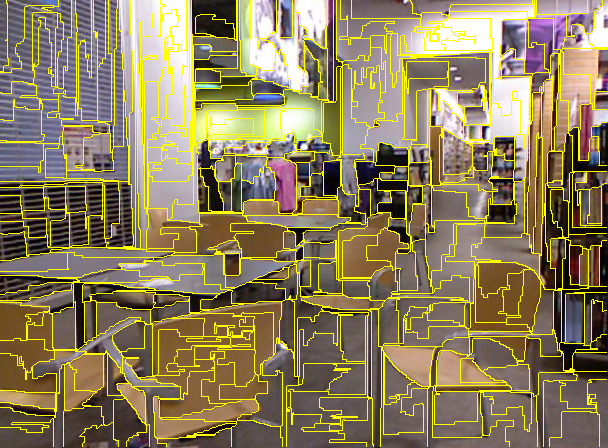}&
			\includegraphics[width=\hhh,height=\hee]{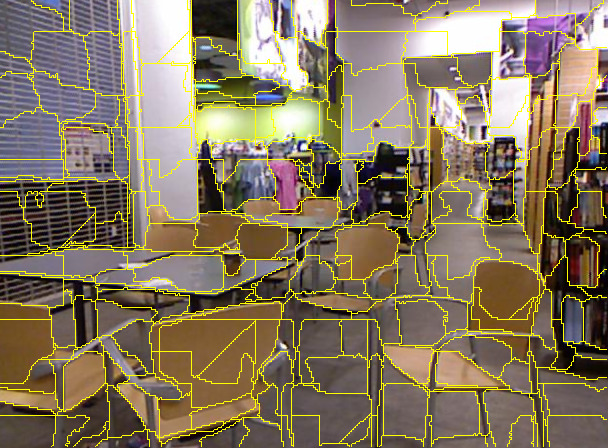}\\
			GMMSP \cite{Ban18} & BASS \cite{Uziel:ICCV:2019:BASS} & FSLIC \cite{wu2020fuzzy} & SSN \cite{jampani2018superpixel}&
			SEAL \cite{tu2018learning} & SFCN \cite{yang2020superpixel}\\[0.5ex]
			\includegraphics[width=\hhh,height=\hee]{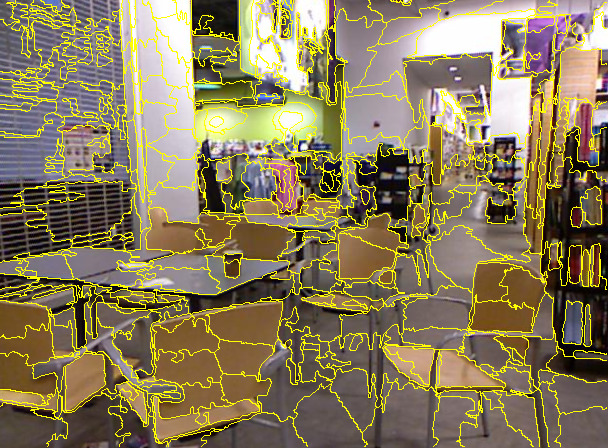}&
			\includegraphics[width=\hhh,height=\hee]{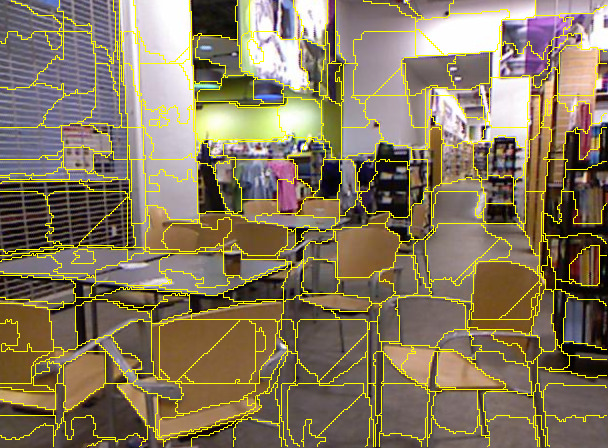}&
			\includegraphics[width=\hhh,height=\hee]{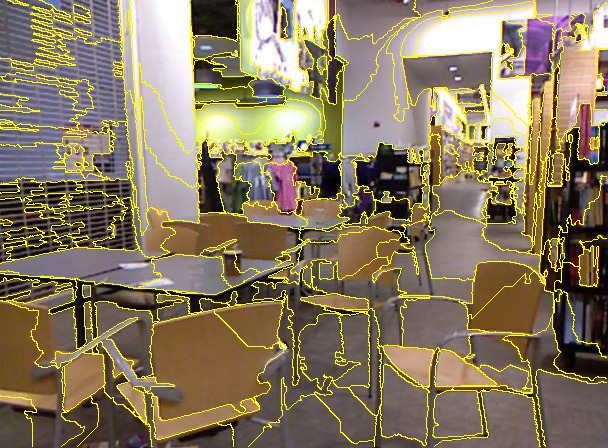}&
			\includegraphics[width=\hhh,height=\hee]{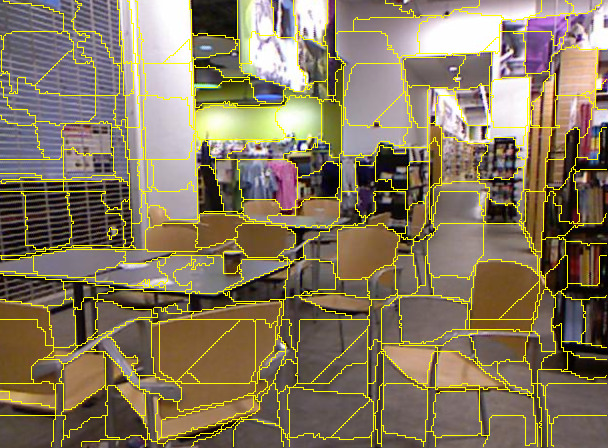}&
			\includegraphics[width=\hhh,height=\hee]{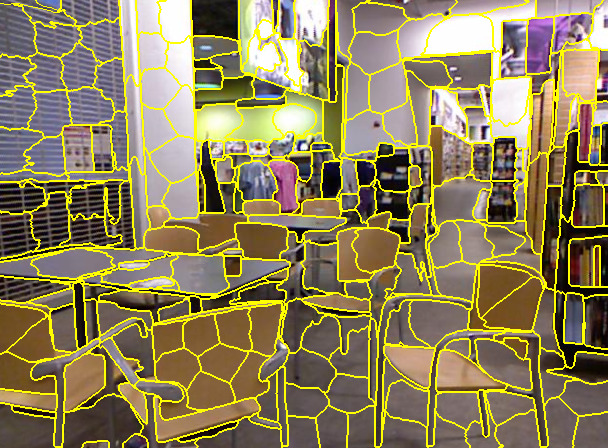}&
			\includegraphics[width=\hhh,height=\hee]{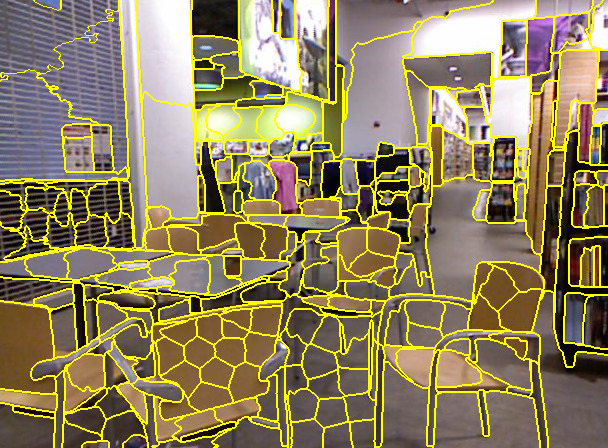}\\
			LNS-Net \cite{zhu2021learning} & AINet \cite{wang2021ainet} & VSSS \cite{zhou2023vine} & CDS \cite{xu2024learning} & SPAM & SPAM (VA $r$=2) \\[0.5ex]
			\includegraphics[width=\hhh,height=\hee]{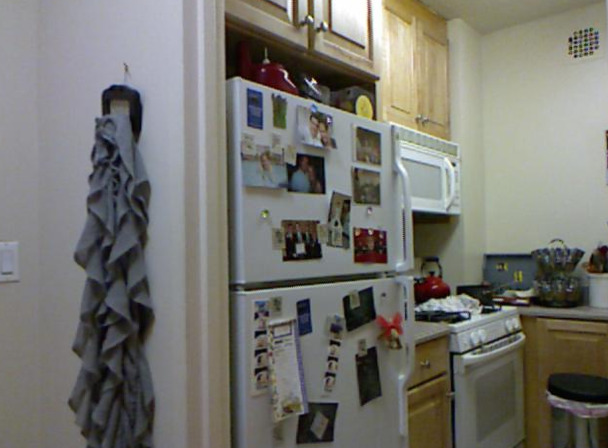}&
			\includegraphics[width=\hhh,height=\hee]{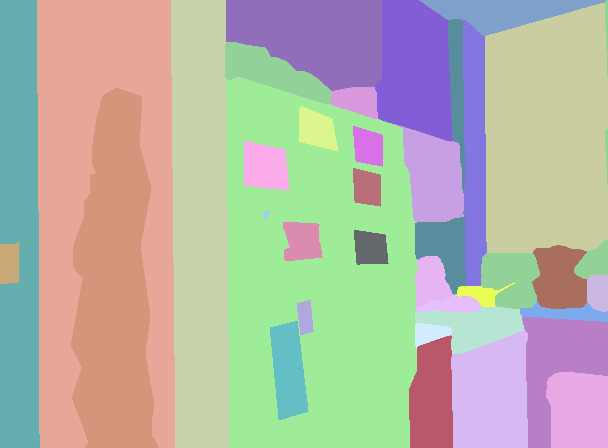}&
			\includegraphics[width=\hhh,height=\hee]{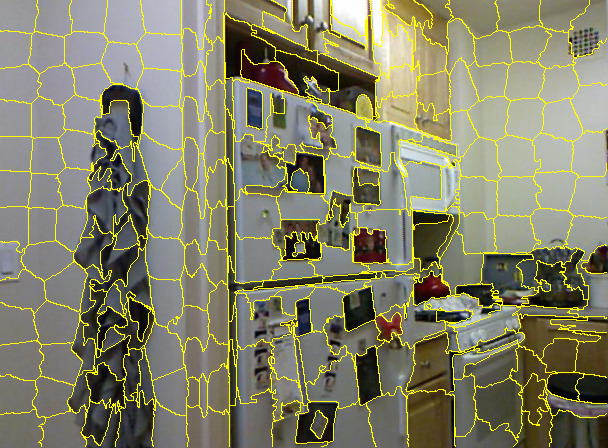}&
			\includegraphics[width=\hhh,height=\hee]{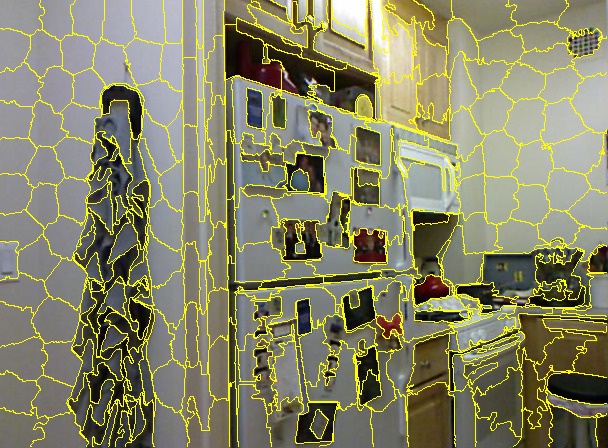}&
			\includegraphics[width=\hhh,height=\hee]{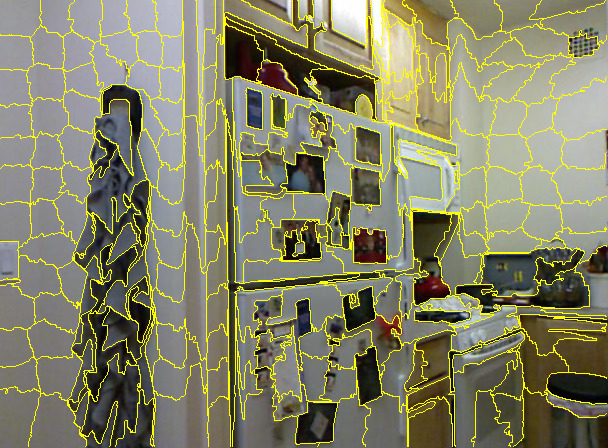}&
			\includegraphics[width=\hhh,height=\hee]{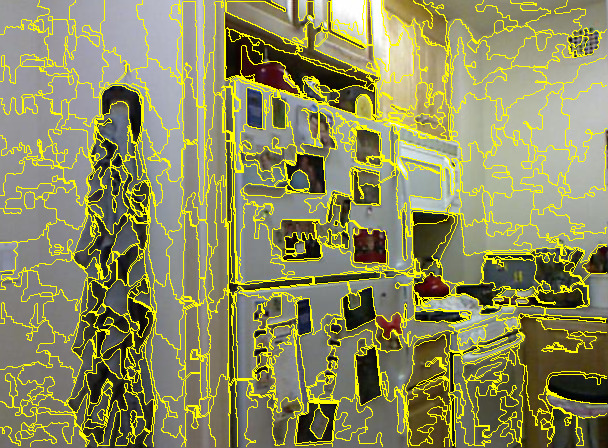}\\
			Image & Groundtruth & SLIC \cite{achanta2012} & LSC \cite{li2015} & SNIC \cite{achanta2017superpixels}&SCAC \cite{yuan2021superpixels}\\[0.5ex]
			\includegraphics[width=\hhh,height=\hee]{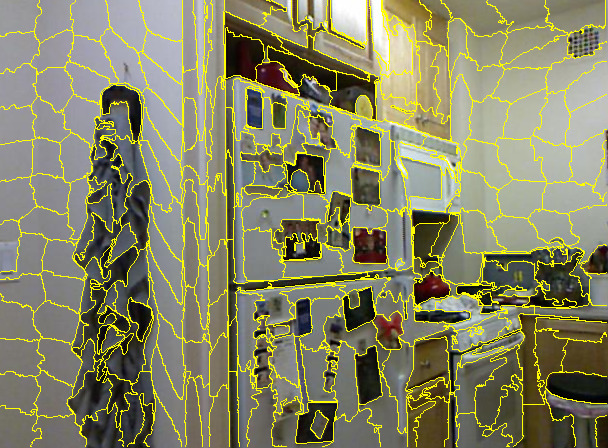}&
			\includegraphics[width=\hhh,height=\hee]{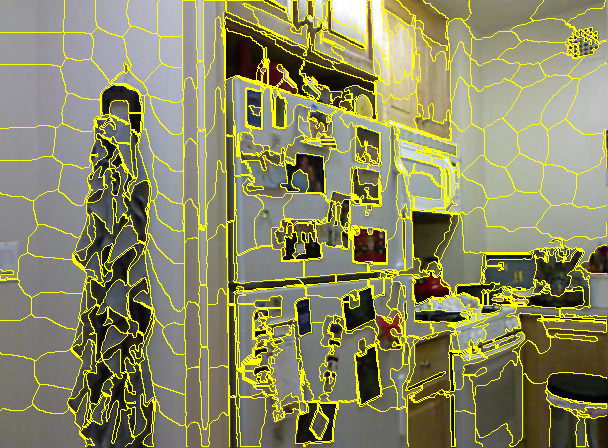}&
			\includegraphics[width=\hhh,height=\hee]{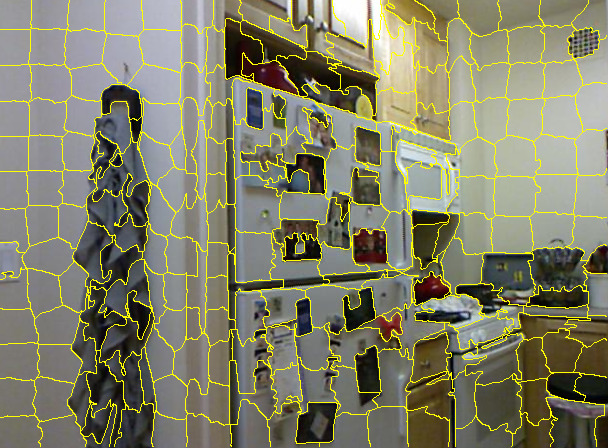}&
			\includegraphics[width=\hhh,height=\hee]{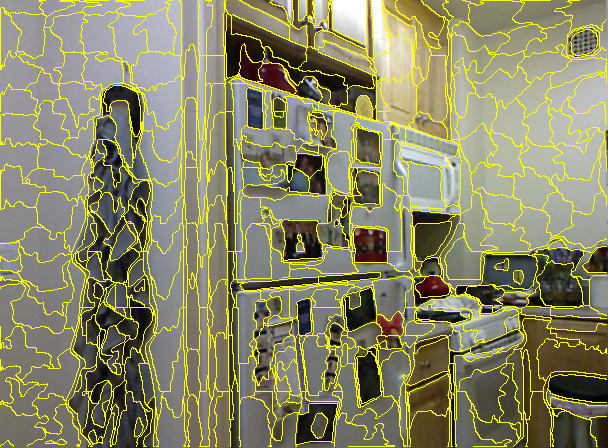}&
			\includegraphics[width=\hhh,height=\hee]{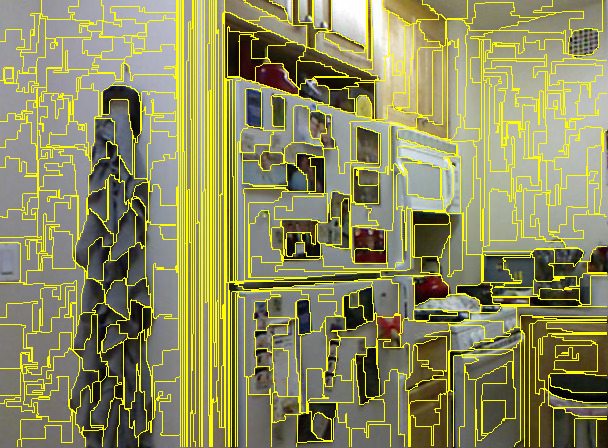}&
			\includegraphics[width=\hhh,height=\hee]{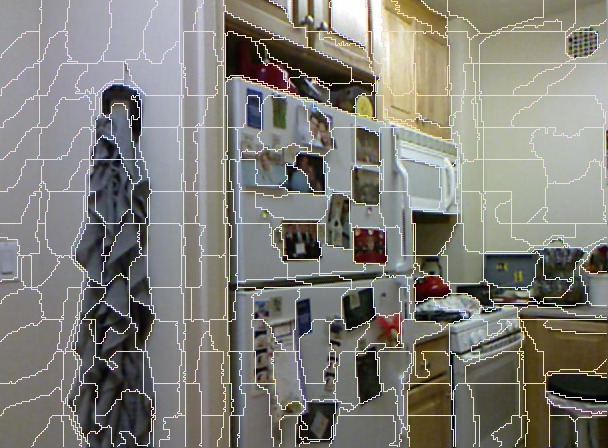}\\
			GMMSP \cite{Ban18} & BASS \cite{Uziel:ICCV:2019:BASS} & FSLIC \cite{wu2020fuzzy} & SSN \cite{jampani2018superpixel}&
			SEAL \cite{tu2018learning} & SFCN \cite{yang2020superpixel}\\[0.5ex]
			\includegraphics[width=\hhh,height=\hee]{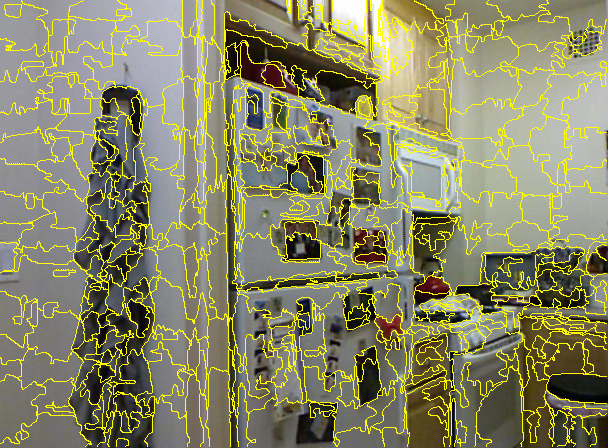}&
			\includegraphics[width=\hhh,height=\hee]{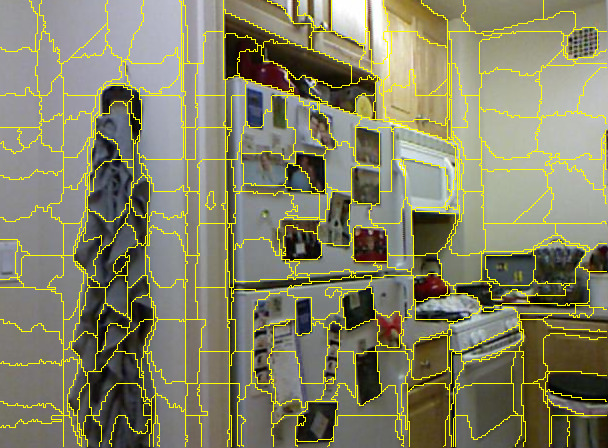}&
			\includegraphics[width=\hhh,height=\hee]{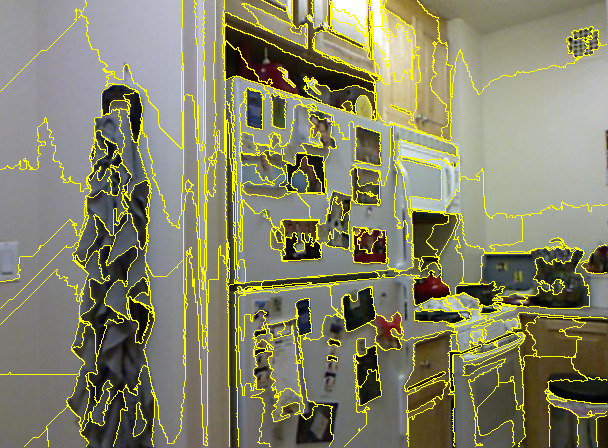}&
			\includegraphics[width=\hhh,height=\hee]{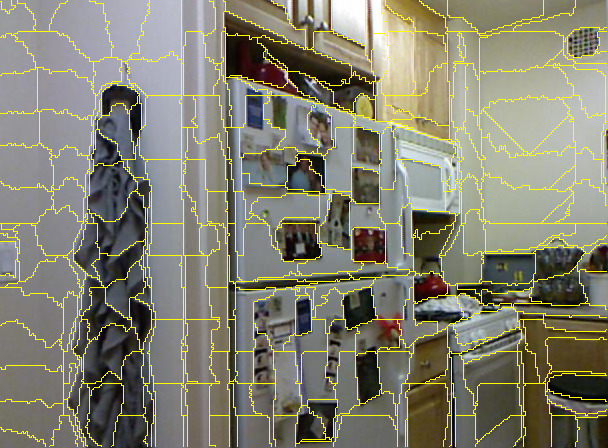}&
			\includegraphics[width=\hhh,height=\hee]{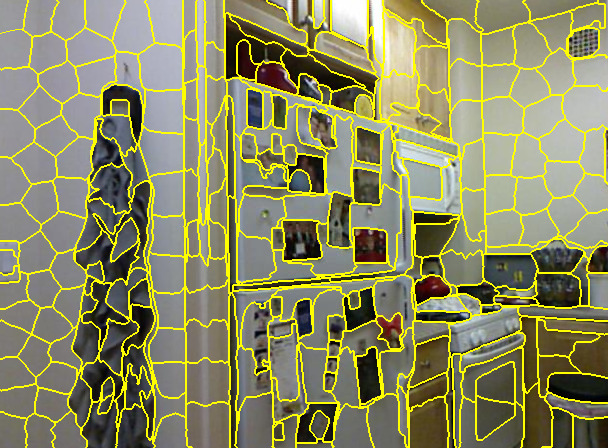}&
			\includegraphics[width=\hhh,height=\hee]{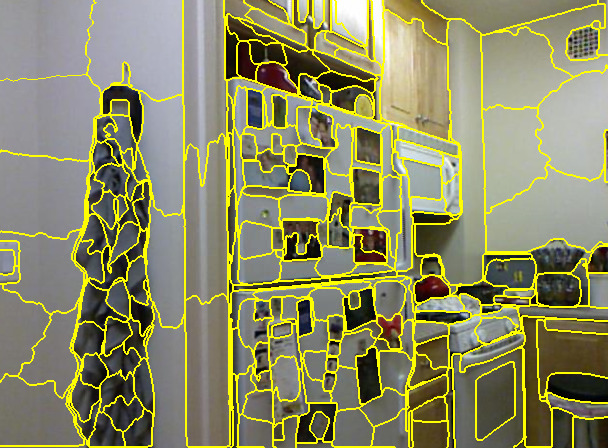}\\
			LNS-Net \cite{zhu2021learning} & AINet \cite{wang2021ainet} & VSSS \cite{zhou2023vine} & CDS \cite{xu2024learning} & SPAM & SPAM (VA $r$=2) \\[-0.25ex] 
			\includegraphics[width=\hhh,height=\hee]{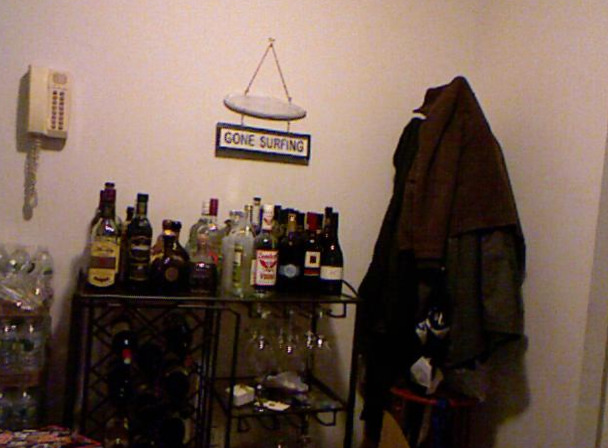}&
			\includegraphics[width=\hhh,height=\hee]{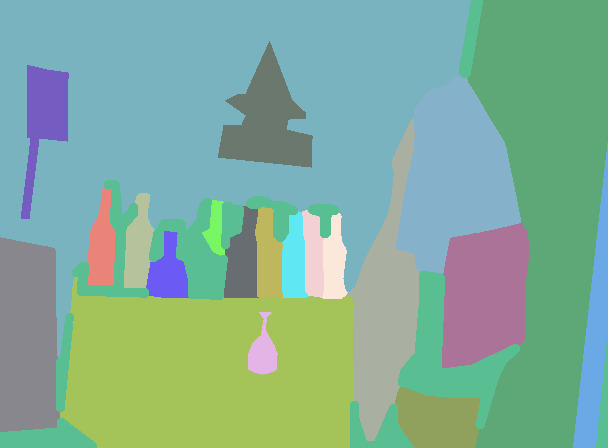}&
			\includegraphics[width=\hhh,height=\hee]{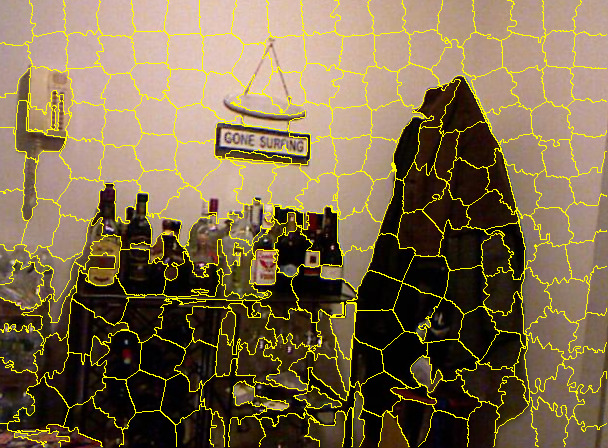}&
			\includegraphics[width=\hhh,height=\hee]{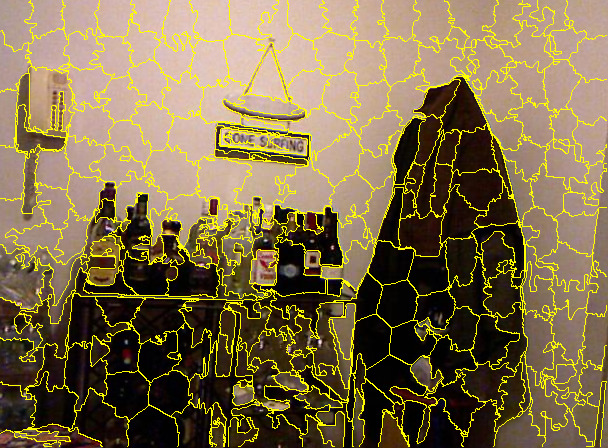}&
			\includegraphics[width=\hhh,height=\hee]{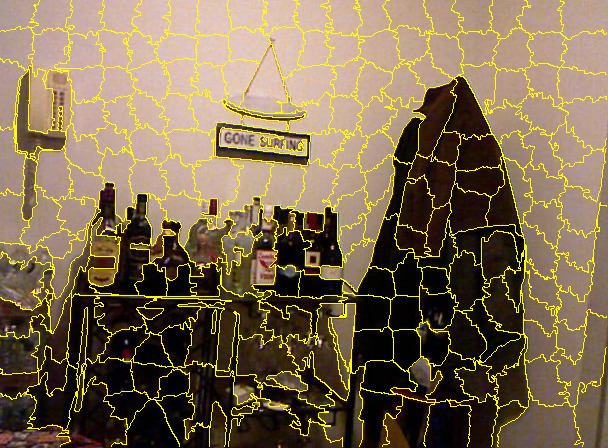}&
			\includegraphics[width=\hhh,height=\hee]{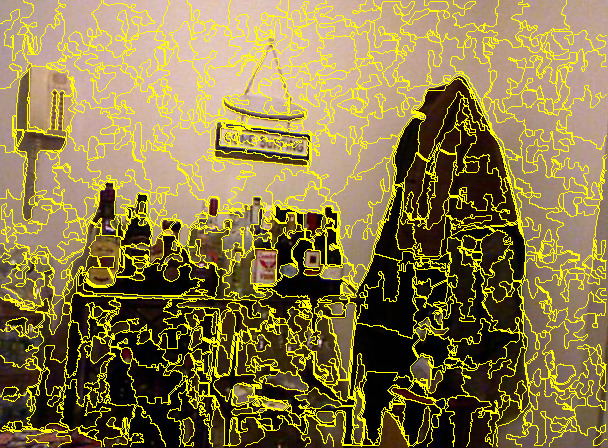}\\
			Image & Groundtruth & SLIC \cite{achanta2012} & LSC \cite{li2015} & SNIC \cite{achanta2017superpixels}&SCAC \cite{yuan2021superpixels}\\[0.5ex]
			\includegraphics[width=\hhh,height=\hee]{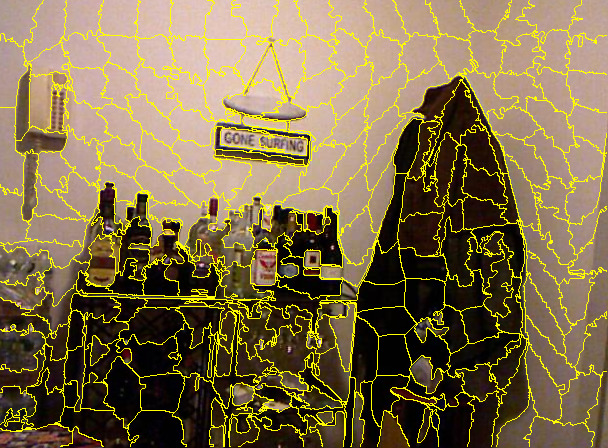}&
			\includegraphics[width=\hhh,height=\hee]{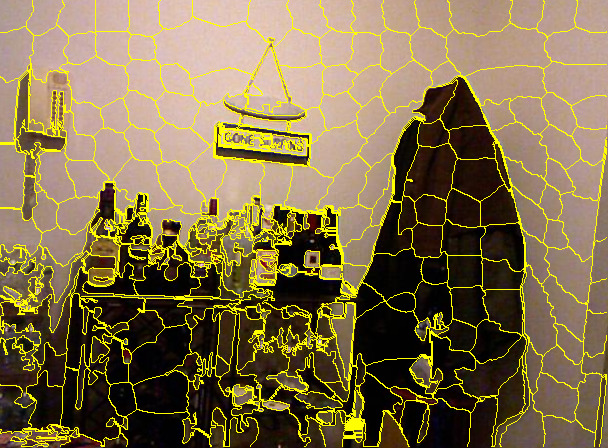}&
			\includegraphics[width=\hhh,height=\hee]{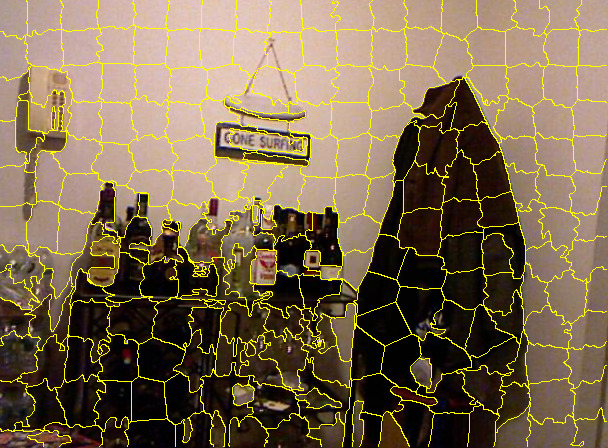}&
			\includegraphics[width=\hhh,height=\hee]{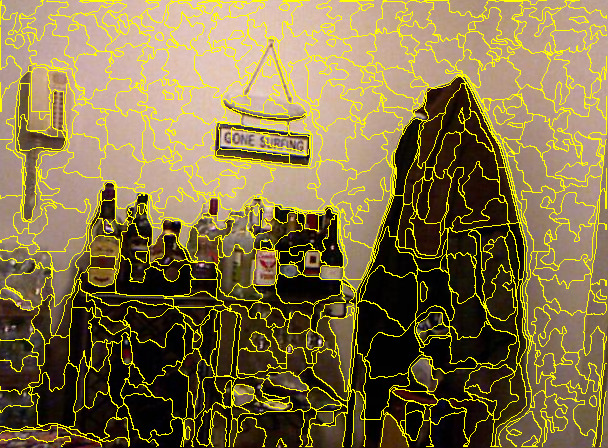}&
			\includegraphics[width=\hhh,height=\hee]{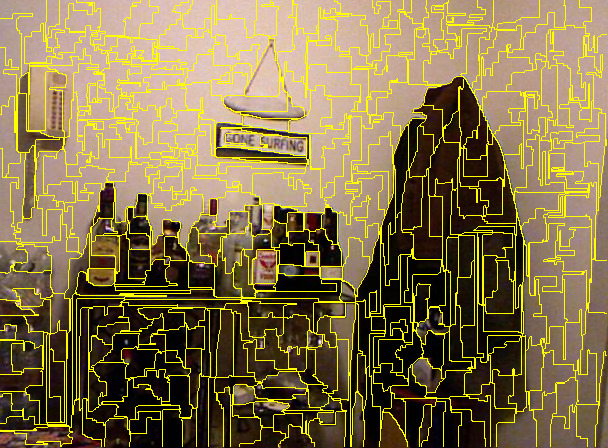}&
			\includegraphics[width=\hhh,height=\hee]{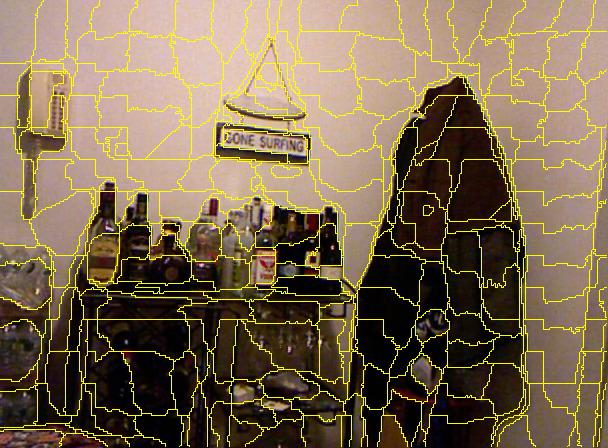}\\
			GMMSP \cite{Ban18} & BASS \cite{Uziel:ICCV:2019:BASS} & FSLIC \cite{wu2020fuzzy} & SSN \cite{jampani2018superpixel}&
			SEAL \cite{tu2018learning} & SFCN \cite{yang2020superpixel}\\[0.5ex]
			\includegraphics[width=\hhh,height=\hee]{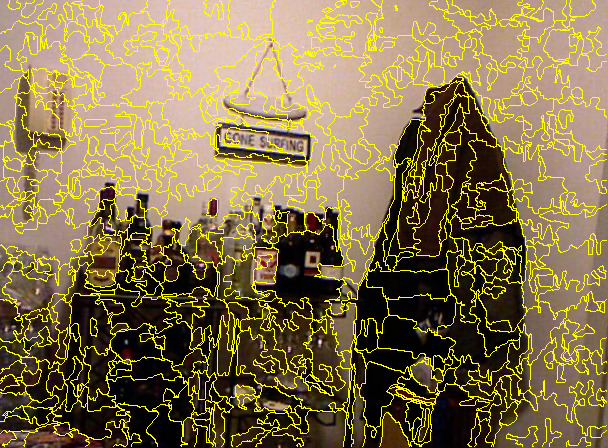}&
			\includegraphics[width=\hhh,height=\hee]{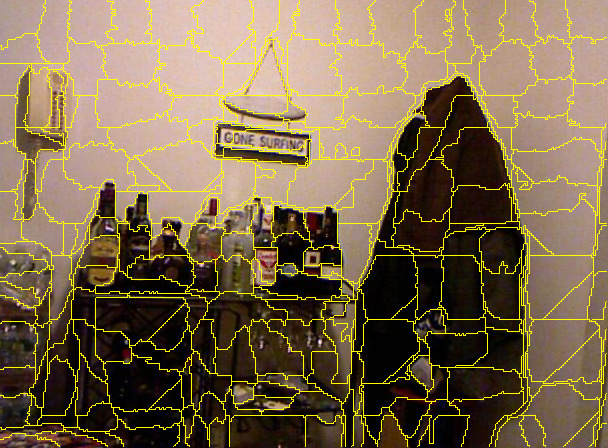}&
			\includegraphics[width=\hhh,height=\hee]{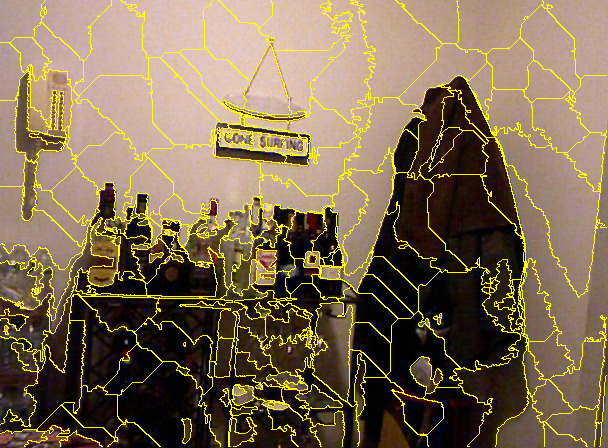}&
			\includegraphics[width=\hhh,height=\hee]{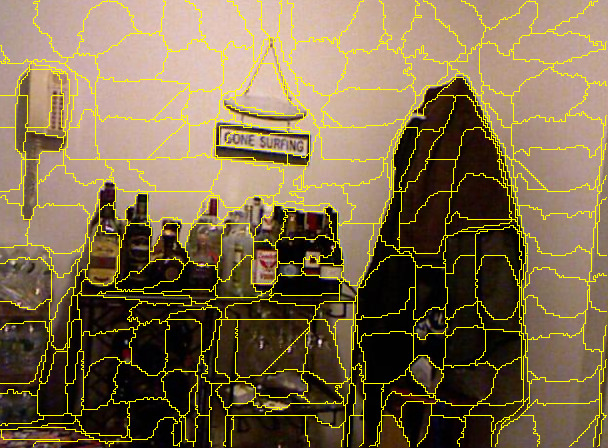}&
			\includegraphics[width=\hhh,height=\hee]{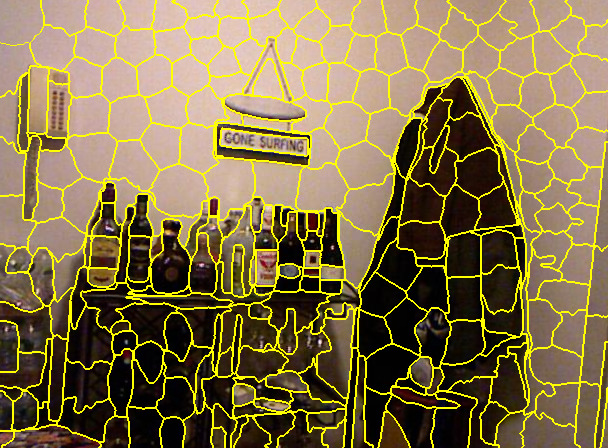}&
			\includegraphics[width=\hhh,height=\hee]{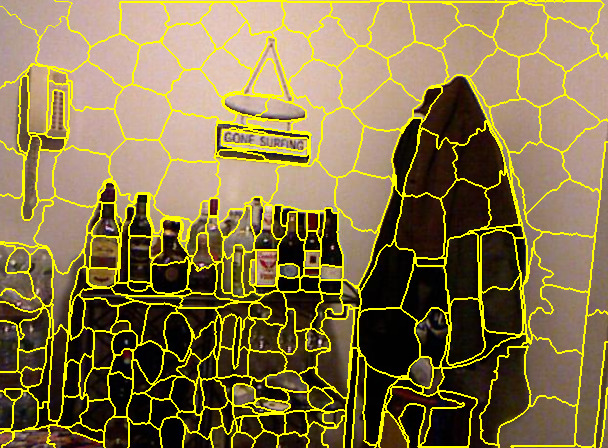}\\
			LNS-Net \cite{zhu2021learning} & AINet \cite{wang2021ainet} & VSSS \cite{zhou2023vine} & CDS \cite{xu2024learning} & SPAM & SPAM (VA $r$=2) \\
	\end{tabular}}%
	\caption{\textbf{Qualitative evaluation of superpixel methods on the NYUv2 test set.}
		The number of requested superpixels is set to $K=250$. 
		SPAM generates superpixels that are both more accurate and considerably more regular than other state-of-the-art 
		DL-based methods.}
	\label{fig:res_quali_NYUV2}
\end{figure*}

\begin{figure*}[t!]
	\centering
	\newcommand{\hhh}{0.16\textwidth}
	\newcommand{\hee}{0.1065\textwidth}
	{\footnotesize \begin{tabular}{@{\hspace{0mm}}c@{\hspace{1mm}}c@{\hspace{1mm}}c@{\hspace{1mm}}c@{\hspace{1mm}}c@{\hspace{1mm}}c@{\hspace{1mm}}@{\hspace{0mm}}}
			\includegraphics[width=\hhh,height=\hee]{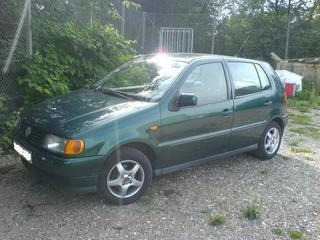}&
			\includegraphics[width=\hhh,height=\hee]{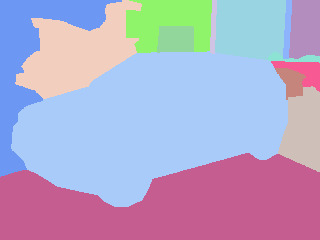}&
			\includegraphics[width=\hhh,height=\hee]{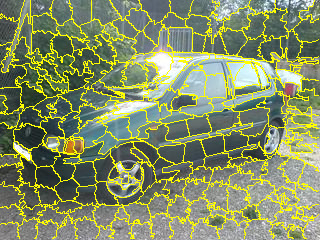}&
			\includegraphics[width=\hhh,height=\hee]{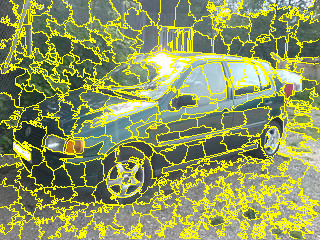}&
			\includegraphics[width=\hhh,height=\hee]{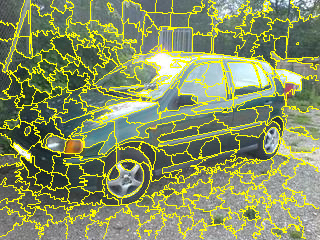}&
			\includegraphics[width=\hhh,height=\hee]{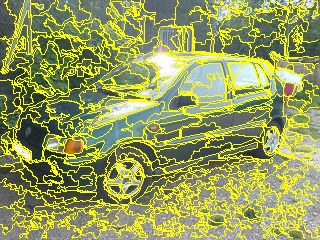}\\
			Image & Groundtruth & SLIC \cite{achanta2012} & LSC \cite{li2015} & SNIC \cite{achanta2017superpixels}&SCAC \cite{yuan2021superpixels}\\[0.5ex]
			\includegraphics[width=\hhh,height=\hee]{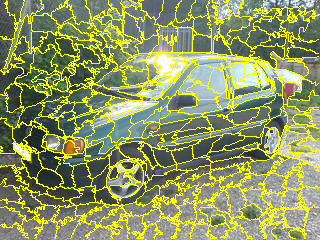}&
			\includegraphics[width=\hhh,height=\hee]{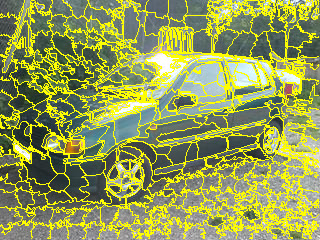}&
			\includegraphics[width=\hhh,height=\hee]{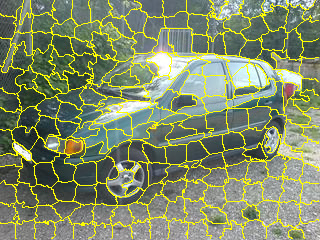}&
			\includegraphics[width=\hhh,height=\hee]{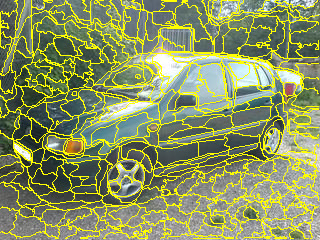}&
			\includegraphics[width=\hhh,height=\hee]{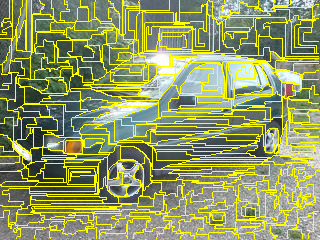}&
			\includegraphics[width=\hhh,height=\hee]{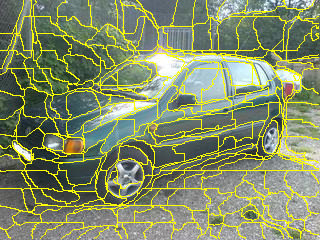}\\
			GMMSP \cite{Ban18} & BASS \cite{Uziel:ICCV:2019:BASS} & FSLIC \cite{wu2020fuzzy} & SSN \cite{jampani2018superpixel}&
			SEAL \cite{tu2018learning} & SFCN \cite{yang2020superpixel}\\[0.5ex]
			\includegraphics[width=\hhh,height=\hee]{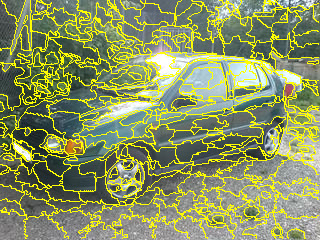}&
			\includegraphics[width=\hhh,height=\hee]{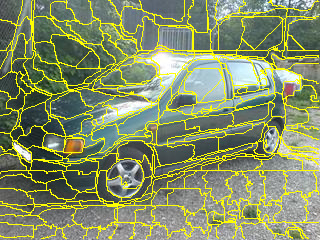}&
			\includegraphics[width=\hhh,height=\hee]{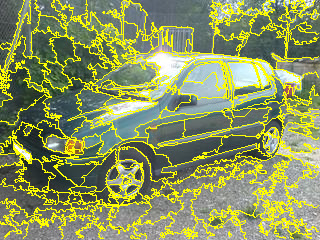}&
			\includegraphics[width=\hhh,height=\hee]{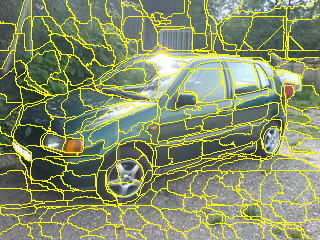}&
			\includegraphics[width=\hhh,height=\hee]{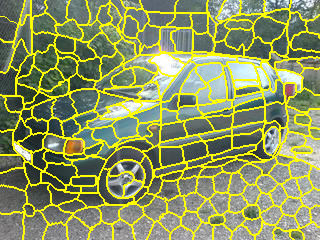}&
			\includegraphics[width=\hhh,height=\hee]{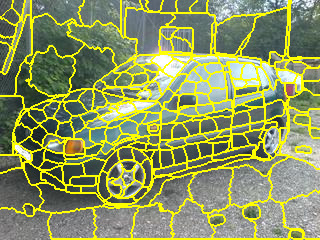}\\
			LNS-Net \cite{zhu2021learning} & AINet \cite{wang2021ainet} & VSSS \cite{zhou2023vine} & CDS \cite{xu2024learning} & SPAM & SPAM (VA $r$=2) \\[0.5ex]
			\includegraphics[width=\hhh,height=\hee]{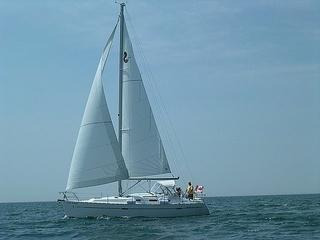}&
			\includegraphics[width=\hhh,height=\hee]{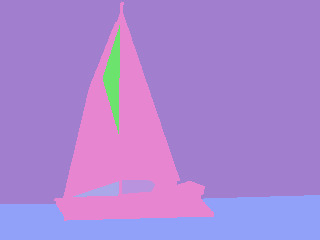}&
			\includegraphics[width=\hhh,height=\hee]{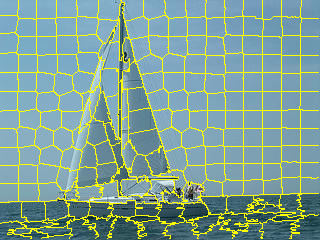}&
			\includegraphics[width=\hhh,height=\hee]{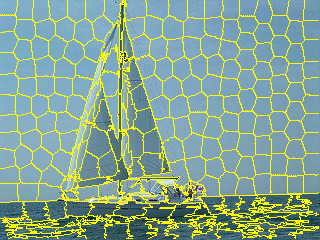}&
			\includegraphics[width=\hhh,height=\hee]{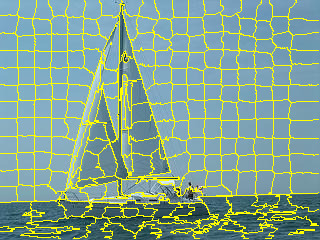}&
			\includegraphics[width=\hhh,height=\hee]{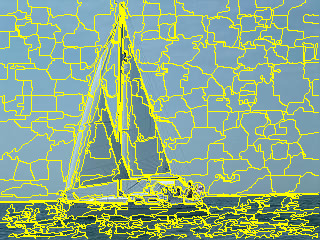}\\
			Image & Groundtruth & SLIC \cite{achanta2012} & LSC \cite{li2015} & SNIC \cite{achanta2017superpixels}&SCAC \cite{yuan2021superpixels}\\[0.5ex]
			\includegraphics[width=\hhh,height=\hee]{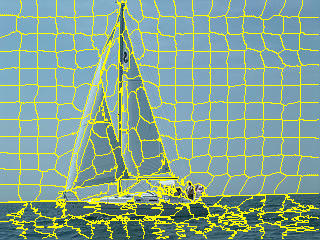}&
			\includegraphics[width=\hhh,height=\hee]{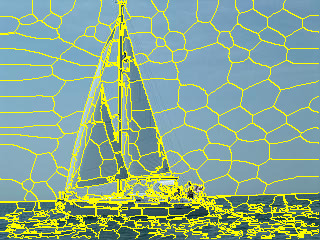}&
			\includegraphics[width=\hhh,height=\hee]{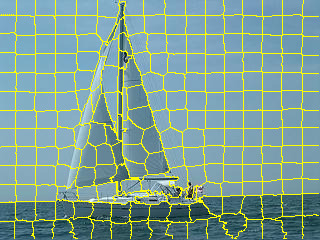}&
			\includegraphics[width=\hhh,height=\hee]{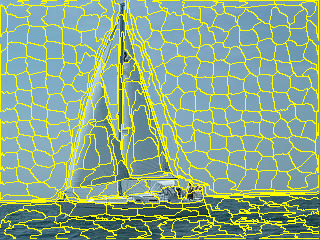}&
			\includegraphics[width=\hhh,height=\hee]{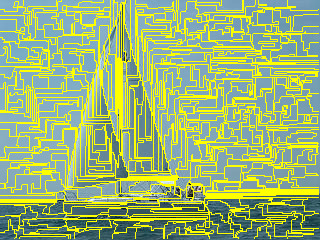}&
			\includegraphics[width=\hhh,height=\hee]{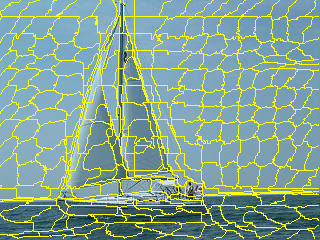}\\
			GMMSP \cite{Ban18} & BASS \cite{Uziel:ICCV:2019:BASS} & FSLIC \cite{wu2020fuzzy} & SSN \cite{jampani2018superpixel}&
			SEAL \cite{tu2018learning} & SFCN \cite{yang2020superpixel}\\[0.5ex]
			\includegraphics[width=\hhh,height=\hee]{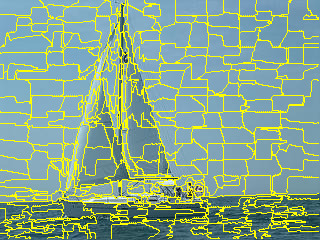}&
			\includegraphics[width=\hhh,height=\hee]{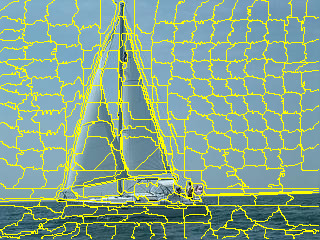}&
			\includegraphics[width=\hhh,height=\hee]{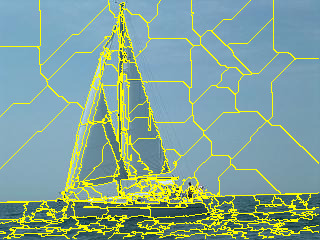}&
			\includegraphics[width=\hhh,height=\hee]{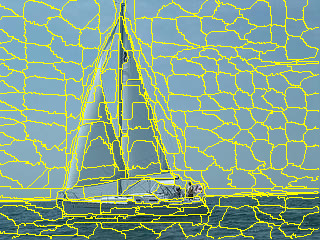}&
			\includegraphics[width=\hhh,height=\hee]{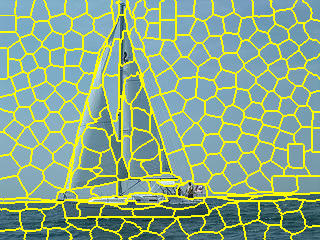}&
			\includegraphics[width=\hhh,height=\hee]{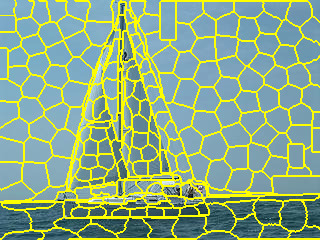}\\
			LNS-Net \cite{zhu2021learning} & AINet \cite{wang2021ainet} & VSSS \cite{zhou2023vine} & CDS \cite{xu2024learning} & SPAM & SPAM (VA $r$=2) \\[-0.25ex] 
			\includegraphics[width=\hhh,height=\hee]{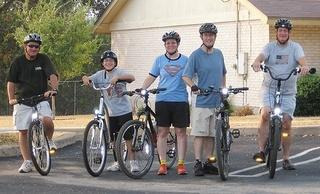}&
			\includegraphics[width=\hhh,height=\hee]{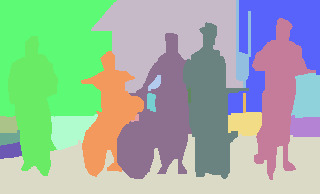}&
			\includegraphics[width=\hhh,height=\hee]{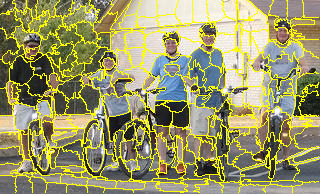}&
			\includegraphics[width=\hhh,height=\hee]{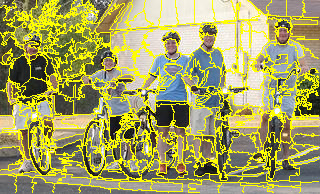}&
			\includegraphics[width=\hhh,height=\hee]{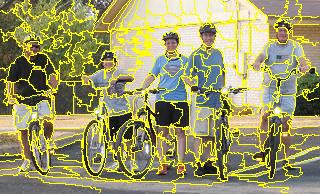}&
			\includegraphics[width=\hhh,height=\hee]{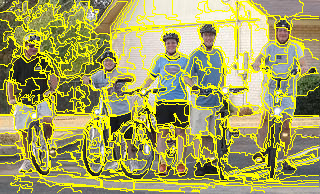}\\
			Image & Groundtruth & SLIC \cite{achanta2012} & LSC \cite{li2015} & SNIC \cite{achanta2017superpixels}&SCAC \cite{yuan2021superpixels}\\[0.5ex]
			\includegraphics[width=\hhh,height=\hee]{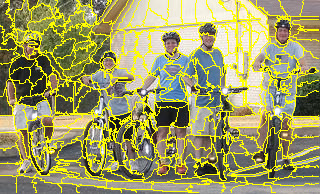}&
			\includegraphics[width=\hhh,height=\hee]{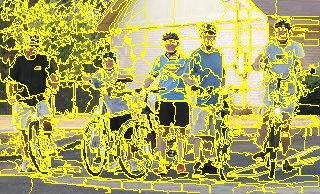}&
			\includegraphics[width=\hhh,height=\hee]{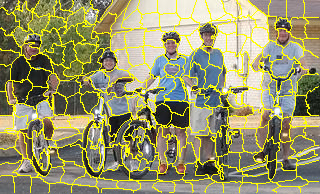}&
			\includegraphics[width=\hhh,height=\hee]{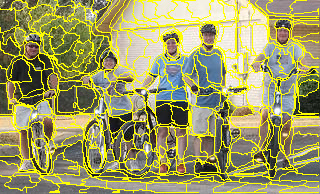}&
			\includegraphics[width=\hhh,height=\hee]{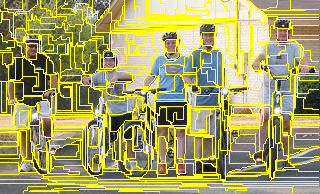}&
			\includegraphics[width=\hhh,height=\hee]{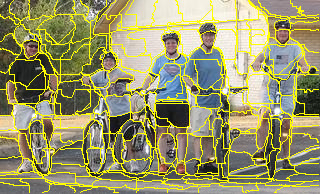}\\
			GMMSP \cite{Ban18} & BASS \cite{Uziel:ICCV:2019:BASS} & FSLIC \cite{wu2020fuzzy} & SSN \cite{jampani2018superpixel}&
			SEAL \cite{tu2018learning} & SFCN \cite{yang2020superpixel}\\[0.5ex]
			\includegraphics[width=\hhh,height=\hee]{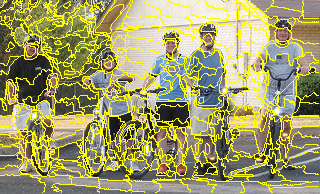}&
			\includegraphics[width=\hhh,height=\hee]{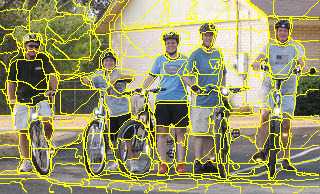}&
			\includegraphics[width=\hhh,height=\hee]{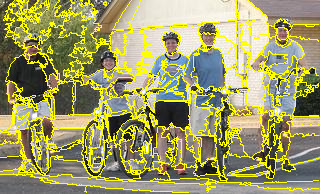}&
			\includegraphics[width=\hhh,height=\hee]{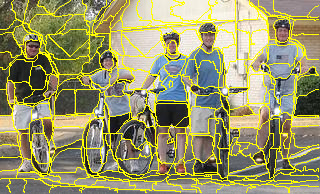}&
			\includegraphics[width=\hhh,height=\hee]{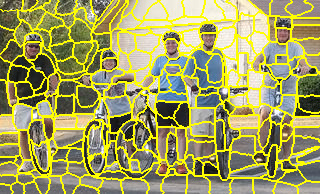}&
			\includegraphics[width=\hhh,height=\hee]{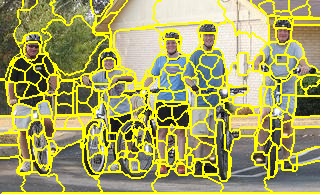}\\
			LNS-Net \cite{zhu2021learning} & AINet \cite{wang2021ainet} & VSSS \cite{zhou2023vine} & CDS \cite{xu2024learning} & SPAM & SPAM (VA $r$=2) \\
	\end{tabular}}
	\caption{\textbf{Qualitative evaluation of superpixel methods on the SBD test set.}
		The number of requested superpixels is set to $K=250$. 
		SPAM generates superpixels that are both more accurate and considerably more regular than other state-of-the-art 
		DL-based methods.}
	\label{fig:res_quali_SBD}
\end{figure*}

\smallskip
\noindent\textbf{Comparison to state-of-the-art methods.}
%
%
In Figs.~\ref{fig:res_quali_BSD}, \ref{fig:res_quali_NYUV2} and \ref{fig:res_quali_SBD}, we respectively show qualitative examples on BSD, NYUv2 and SBD test images.
SPAM generates highly accurate yet regular superpixels, compared to more irregular ones of recent DL-based methods.

\end{document}

%% file: preamble.tex
\usepackage[dvipsnames]{xcolor}


%% file: main.bbl
\begin{thebibliography}{60}
\providecommand{\natexlab}[1]{#1}
\providecommand{\url}[1]{\texttt{#1}}
\expandafter\ifx\csname urlstyle\endcsname\relax
  \providecommand{\doi}[1]{doi: #1}\else
  \providecommand{\doi}{doi: \begingroup \urlstyle{rm}\Url}\fi

\bibitem[Achanta and S{\"u}sstrunk(2017)]{achanta2017superpixels}
Radhakrishna Achanta and Sabine S{\"u}sstrunk.
\newblock Superpixels and polygons using simple non-iterative clustering.
\newblock In \emph{CVPR}, 2017.

\bibitem[Achanta et~al.(2012)Achanta, Shaji, Smith, Lucchi, Fua, and
  S{\"u}sstrunk]{achanta2012}
Radhakrishna Achanta, Appu Shaji, Kevin Smith, Aur{\'e}lien Lucchi, Pascal Fua,
  and Sabine S{\"u}sstrunk.
\newblock {SLIC} superpixels compared to state-of-the-art superpixel methods.
\newblock \emph{IEEE TPAMI}, 34:\penalty0 2274--2282, 2012.

\bibitem[Ban et~al.(2018)Ban, Liu, and Cao]{Ban18}
Zhihua Ban, Jianguo Liu, and Li Cao.
\newblock Superpixel segmentation using gaussian mixture model.
\newblock \emph{IEEE TIP}, 27\penalty0 (8):\penalty0 4105--4117, 2018.

\bibitem[Berg et~al.(2019)Berg, Kutra, Kroeger, Straehle, Kausler, Haubold,
  Schiegg, Ales, Beier, Rudy, et~al.]{berg2019ilastik}
Stuart Berg, Dominik Kutra, Thorben Kroeger, Christoph~N Straehle, Bernhard~X
  Kausler, Carsten Haubold, Martin Schiegg, Janez Ales, Thorsten Beier, Markus
  Rudy, et~al.
\newblock {Ilastik: Interactive machine learning for (bio) image analysis}.
\newblock \emph{Nature methods}, 16\penalty0 (12):\penalty0 1226--1232, 2019.

\bibitem[Caron et~al.(2021)Caron, Touvron, Misra, J{\'e}gou, Mairal,
  Bojanowski, and Joulin]{caron2021emerging}
Mathilde Caron, Hugo Touvron, Ishan Misra, Herv{\'e} J{\'e}gou, Julien Mairal,
  Piotr Bojanowski, and Armand Joulin.
\newblock Emerging properties in self-supervised vision transformers.
\newblock In \emph{CVPR}, 2021.

\bibitem[Chang et~al.(2013)Chang, Wei, and Fisher]{chang2013video}
Jason Chang, Donglai Wei, and John~W. Fisher.
\newblock A video representation using temporal superpixels.
\newblock In \emph{CVPR}, 2013.

\bibitem[Chen et~al.(2017)Chen, Li, and Huang]{chen2017}
Jiansheng Chen, Zhengqin Li, and Bo Huang.
\newblock Linear spectral clustering superpixel.
\newblock \emph{IEEE TIP}, 26:\penalty0 3317--3330, 2017.

\bibitem[Chen(2017)]{chen2017rethinking}
Liang-Chieh Chen.
\newblock Rethinking atrous convolution for semantic image segmentation.
\newblock \emph{arXiv:1706.05587}, 2017.

\bibitem[Chen et~al.(2023)Chen, Wang, Guo, and Zhang]{chen2023structnerf}
Zheng Chen, Chen Wang, Yuan-Chen Guo, and Song-Hai Zhang.
\newblock Structnerf: Neural radiance fields for indoor scenes with structural
  hints.
\newblock \emph{IEEE TPAMI}, 45\penalty0 (12):\penalty0 15694--15705, 2023.

\bibitem[Everingham et~al.(2015)Everingham, Eslami, Van~Gool, Williams, Winn,
  and Zisserman]{pascalvoc}
Mark Everingham, SM~Ali Eslami, Luc Van~Gool, Christopher~KI Williams, John
  Winn, and Andrew Zisserman.
\newblock {The PASCAL Visual Object Classes Challenge: A Retrospective}.
\newblock \emph{IJCV}, 111:\penalty0 98--136, 2015.

\bibitem[Felzenszwalb and Huttenlocher(2004)]{felzenszwalb2004}
Pedro~F. Felzenszwalb and Daniel~P. Huttenlocher.
\newblock Efficient graph-based image segmentation.
\newblock \emph{IJCV}, 59\penalty0 (2):\penalty0 167--181, 2004.

\bibitem[Giraud and Cl{\'e}ment(2024)]{giraud2024_tip}
R{\'e}mi Giraud and Micha{\"e}l Cl{\'e}ment.
\newblock Superpixel segmentation: A long-lasting ill-posed problem.
\newblock \emph{arXiv:2411.06478}, 2024.

\bibitem[Giraud et~al.(2017{\natexlab{a}})Giraud, Ta, Bugeau, Coup{\'e}, and
  Papadakis]{giraud2017_spm}
R{\'e}mi Giraud, Vinh-Thong Ta, Aurélie Bugeau, Pierrick Coup{\'e}, and
  Nicolas Papadakis.
\newblock {SuperPatchMatch}: An algorithm for robust correspondences using
  superpixel patches.
\newblock \emph{IEEE TIP}, 26\penalty0 (8):\penalty0 4068--4078,
  2017{\natexlab{a}}.

\bibitem[Giraud et~al.(2017{\natexlab{b}})Giraud, Ta, and
  Papadakis]{giraud2017_jei}
R{\'e}mi Giraud, Vinh-Thong Ta, and Nicolas Papadakis.
\newblock Evaluation framework of superpixel methods with a global regularity
  measure.
\newblock \emph{JEI}, 26\penalty0 (6), 2017{\natexlab{b}}.

\bibitem[Gould et~al.(2008)Gould, Rodgers, Cohen, Elidan, and
  Koller]{gould2008}
Stephen Gould, Jim Rodgers, David Cohen, Gal Elidan, and Daphne Koller.
\newblock Multi-class segmentation with relative location prior.
\newblock \emph{IJCV}, 80\penalty0 (3):\penalty0 300--316, 2008.

\bibitem[Gould et~al.(2009)Gould, Fulton, and Koller]{gould2009decomposing}
Stephen Gould, Richard Fulton, and Daphne Koller.
\newblock Decomposing a scene into geometric and semantically consistent
  regions.
\newblock In \emph{ICCV}, 2009.

\bibitem[Jampani et~al.(2018)Jampani, Sun, Liu, Yang, and
  Kautz]{jampani2018superpixel}
Varun Jampani, Deqing Sun, Ming-Yu Liu, Ming-Hsuan Yang, and Jan Kautz.
\newblock Superpixel sampling networks.
\newblock In \emph{ECCV}, 2018.

\bibitem[Kang et~al.(2020)Kang, Zhu, and Ming]{kang2020dynamic}
Xuejing Kang, Lei Zhu, and Anlong Ming.
\newblock Dynamic random walk for superpixel segmentation.
\newblock \emph{IEEE TIP}, 29:\penalty0 3871--3884, 2020.

\bibitem[Ke et~al.(2024)Ke, Mo, and Stella]{ke2023learning}
Tsung-Wei Ke, Sangwoo Mo, and X~Yu Stella.
\newblock Learning hierarchical image segmentation for recognition and by
  recognition.
\newblock In \emph{ICLR}, 2024.

\bibitem[Kim et~al.(2023)Kim, Oh, Hwang, Kwak, and Ok]{kim2023adaptive}
Hoyoung Kim, Minhyeon Oh, Sehyun Hwang, Suha Kwak, and Jungseul Ok.
\newblock Adaptive superpixel for active learning in semantic segmentation.
\newblock In \emph{ICCV}, 2023.

\bibitem[Kirillov et~al.(2023)Kirillov, Mintun, Ravi, Mao, Rolland, Gustafson,
  Xiao, Whitehead, Berg, Lo, Doll{\'{a}}r, and Girshick]{kirillov23sam}
Alexander Kirillov, Eric Mintun, Nikhila Ravi, Hanzi Mao, Chlo{\'{e}} Rolland,
  Laura Gustafson, Tete Xiao, Spencer Whitehead, Alexander~C. Berg, Wan{-}Yen
  Lo, Piotr Doll{\'{a}}r, and Ross~B. Girshick.
\newblock Segment anything.
\newblock In \emph{ICCV}, 2023.

\bibitem[Lee et~al.(2022)Lee, Park, Cho, and Lee]{lee2022spsn}
Minhyeok Lee, Chaewon Park, Suhwan Cho, and Sangyoun Lee.
\newblock Spsn: Superpixel prototype sampling network for rgb-d salient object
  detection.
\newblock In \emph{ECCV}, 2022.

\bibitem[Li et~al.(2024)Li, Zhao, Wang, Wang, and Wang]{li2024superpixel}
Jiayi Li, Xile Zhao, Jianli Wang, Chao Wang, and Min Wang.
\newblock Superpixel-informed implicit neural representation for
  multi-dimensional data.
\newblock In \emph{ECCV}, 2024.

\bibitem[Li et~al.(2016)Li, Zhang, Yu, and Zhang]{li2016pmsc}
Lincheng Li, Shunli Zhang, Xin Yu, and Li Zhang.
\newblock {PMSC: PatchMatch-based superpixel cut for accurate stereo matching}.
\newblock \emph{IEEE TCSVT}, 28\penalty0 (3):\penalty0 679--692, 2016.

\bibitem[Li and Chen(2015)]{li2015}
Zhengqin Li and Jiansheng Chen.
\newblock Superpixel segmentation using linear spectral clustering.
\newblock In \emph{CVPR}, 2015.

\bibitem[Liu et~al.(2011)Liu, Tuzel, Ramalingam, and Chellappa]{liu2011}
Ming-Yu Liu, Oncel Tuzel, Srikumar Ramalingam, and Rama Chellappa.
\newblock Entropy rate superpixel segmentation.
\newblock In \emph{CVPR}, 2011.

\bibitem[Liu et~al.(2019)Liu, Lyu, King, and Xu]{liu2019selflow}
Pengpeng Liu, Michael Lyu, Irwin King, and Jia Xu.
\newblock Selflow: Self-supervised learning of optical flow.
\newblock In \emph{CVPR}, 2019.

\bibitem[Liu et~al.(2016)Liu, Yu, Yu, and He]{liu2016manifold}
Y.-J. Liu, C.-C. Yu, M.-J. Yu, and Y. He.
\newblock Manifold {SLIC}: A fast method to compute content-sensitive
  superpixels.
\newblock In \emph{CVPR}, 2016.

\bibitem[Long et~al.(2018)Long, Feng, Zhu, Zhang, and Gou]{long2018efficient}
Jianwu Long, Xin Feng, Xiaofei Zhu, Jianxun Zhang, and Guanglei Gou.
\newblock Efficient superpixel-guided interactive image segmentation based on
  graph theory.
\newblock \emph{Symmetry}, 10\penalty0 (5):\penalty0 169, 2018.

\bibitem[Machairas et~al.(2015)Machairas, Faessel, C{\'a}rdenas-Pe{\~n}a,
  Chabardes, Walter, and Decenci{\`e}re]{machairas2015}
Va{\"\i}a Machairas, Matthieu Faessel, David C{\'a}rdenas-Pe{\~n}a,
  Th{\'e}odore Chabardes, Thomas Walter, and Etienne Decenci{\`e}re.
\newblock Waterpixels.
\newblock \emph{IEEE TIP}, 24\penalty0 (11):\penalty0 3707--3716, 2015.

\bibitem[Martin et~al.(2001)Martin, Fowlkes, Tal, and Malik]{martin2001}
David Martin, Charless Fowlkes, Doron Tal, and Jitendra Malik.
\newblock A database of human segmented natural images and its application to
  evaluating segmentation algorithms and measuring ecological statistics.
\newblock In \emph{ICCV}, 2001.

\bibitem[Martin et~al.(2004)Martin, Fowlkes, and Malik]{martin2004}
David~R Martin, Charless~C Fowlkes, and Jitendra Malik.
\newblock Learning to detect natural image boundaries using local brightness,
  color, and texture cues.
\newblock \emph{IEEE TPAMI}, 26\penalty0 (5):\penalty0 530--549, 2004.

\bibitem[Massoudifar et~al.(2014)Massoudifar, Rangarajan, and
  Gader]{massoudifar2014superpixel}
Pegah Massoudifar, Anand Rangarajan, and Paul Gader.
\newblock Superpixel estimation for hyperspectral imagery.
\newblock In \emph{CVPRW}, 2014.

\bibitem[Mei et~al.(2025)Mei, Chen, Yuille, and Xie]{mei2025spformer}
Jieru Mei, Liang-Chieh Chen, Alan Yuille, and Cihang Xie.
\newblock {SPF}ormer: Enhancing vision transformer with superpixel
  representation.
\newblock \emph{TMLR}, 2025.

\bibitem[Moore et~al.(2008)Moore, Prince, Warrell, Mohammed, and
  Jones]{moore2008}
A.~P. Moore, S.~J.~D. Prince, J. Warrell, U. Mohammed, and G. Jones.
\newblock Superpixel lattices.
\newblock In \emph{CVPR}, 2008.

\bibitem[Pan et~al.(2022)Pan, Zhou, Zhang, and Zhang]{pan2022fast}
Xiao Pan, Yuanfeng Zhou, Yunfeng Zhang, and Caiming Zhang.
\newblock Fast generation of superpixels with lattice topology.
\newblock \emph{IEEE TIP}, 31:\penalty0 4828--4841, 2022.

\bibitem[Peng et~al.(2022)Peng, Aviles-Rivero, and Sch{\"o}nlieb]{peng2022hers}
Hankui Peng, Angelica~I Aviles-Rivero, and Carola-Bibiane Sch{\"o}nlieb.
\newblock {HERS} superpixels: Deep affinity learning for hierarchical entropy
  rate segmentation.
\newblock In \emph{WACV}, 2022.

\bibitem[Ren and Malik(2003)]{ren2003}
Xiaofeng Ren and Jitendra Malik.
\newblock Learning a classification model for segmentation.
\newblock In \emph{ICCV}, 2003.

\bibitem[Sarkar and Sahay(2021)]{sarkar2021non}
Sourish Sarkar and Rajiv~Ranjan Sahay.
\newblock A non-local superpatch-based algorithm exploiting low rank prior for
  restoration of hyperspectral images.
\newblock \emph{IEEE TIP}, 30:\penalty0 6335--6348, 2021.

\bibitem[Schick et~al.(2014)Schick, Fischer, and
  Stiefelhagen]{schick2014evaluation}
Alexander Schick, Mika Fischer, and Rainer Stiefelhagen.
\newblock An evaluation of the compactness of superpixels.
\newblock \emph{PRL}, 43:\penalty0 71--80, 2014.

\bibitem[Silberman et~al.(2012)Silberman, Hoiem, Kohli, and
  Fergus]{silberman2012indoor}
Nathan Silberman, Derek Hoiem, Pushmeet Kohli, and Rob Fergus.
\newblock {Indoor segmentation and support inference from RGBD images}.
\newblock In \emph{ECCV}, 2012.

\bibitem[Stutz et~al.(2018)Stutz, Hermans, and Leibe]{stutz2018superpixels}
David Stutz, Alexander Hermans, and Bastian Leibe.
\newblock Superpixels: An evaluation of the state-of-the-art.
\newblock \emph{CVIU}, 166:\penalty0 1--27, 2018.

\bibitem[Tian et~al.(2017)Tian, Liu, Zhang, Xue, and Fei]{tian2017supervoxel}
Zhiqiang Tian, Lizhi Liu, Zhenfeng Zhang, Jianru Xue, and Baowei Fei.
\newblock A supervoxel-based segmentation method for prostate mr images.
\newblock \emph{Medical physics}, 44\penalty0 (2):\penalty0 558--569, 2017.

\bibitem[Tu et~al.(2018)Tu, Liu, Jampani, Sun, Chien, Yang, and
  Kautz]{tu2018learning}
Wei-Chih Tu, Ming-Yu Liu, Varun Jampani, Deqing Sun, Shao-Yi Chien, Ming-Hsuan
  Yang, and Jan Kautz.
\newblock Learning superpixels with segmentation-aware affinity loss.
\newblock In \emph{CVPR}, 2018.

\bibitem[Uziel et~al.(2019)Uziel, Ronen, and Freifeld]{Uziel:ICCV:2019:BASS}
Roy Uziel, Meitar Ronen, and Oren Freifeld.
\newblock Bayesian adaptive superpixel segmentation.
\newblock In \emph{ICCV}, 2019.

\bibitem[{Van den Bergh} et~al.(2012){Van den Bergh}, Boix, Roig, {de
  Capitani}, and {Van Gool}]{vandenbergh2012}
Michael {Van den Bergh}, Xavier Boix, Gemma Roig, Benjamin {de Capitani}, and
  Luc {Van Gool}.
\newblock {SEEDS}: Superpixels extracted via energy-driven sampling.
\newblock In \emph{ECCV}, 2012.

\bibitem[Wang et~al.(2021{\natexlab{a}})Wang, Bertasius, Oh, Gupta, Hoai, and
  Torresani]{wang2021supervoxel}
Yang Wang, Gedas Bertasius, Tae-Hyun Oh, Abhinav Gupta, Minh Hoai, and Lorenzo
  Torresani.
\newblock Supervoxel attention graphs for long-range video modeling.
\newblock In \emph{WACV}, 2021{\natexlab{a}}.

\bibitem[Wang et~al.(2021{\natexlab{b}})Wang, Wei, Qian, Zhu, and
  Yang]{wang2021ainet}
Yaxiong Wang, Yunchao Wei, Xueming Qian, Li Zhu, and Yi Yang.
\newblock {AINet: Association implantation for superpixel segmentation}.
\newblock In \emph{ICCV}, 2021{\natexlab{b}}.

\bibitem[Wei et~al.(2018)Wei, Yang, Gong, Ahuja, and Yang]{wei2018}
Xing Wei, Qingxiong Yang, Yihong Gong, Narendra Ahuja, and Ming-Hsuan Yang.
\newblock Superpixel hierarchy.
\newblock \emph{IEEE TIP}, 27\penalty0 (10):\penalty0 4838--4849, 2018.

\bibitem[Wu et~al.(2020)Wu, Zheng, Feng, Zhang, Zhang, Cao, and
  Yan]{wu2020fuzzy}
Chong Wu, Jiangbin Zheng, Zhenan Feng, Houwang Zhang, Le Zhang, Jiawang Cao,
  and Hong Yan.
\newblock Fuzzy {SLIC}: Fuzzy simple linear iterative clustering.
\newblock \emph{IEEE TCSVT}, 31\penalty0 (6):\penalty0 2114--2124, 2020.

\bibitem[Xu et~al.(2024)Xu, Wei, Ruan, and Liao]{xu2024learning}
Sen Xu, Shikui Wei, Tao Ruan, and Lixin Liao.
\newblock Learning invariant inter-pixel correlations for superpixel
  generation.
\newblock In \emph{AAAI}, 2024.

\bibitem[Xu et~al.(2022)Xu, Gao, Zhang, Tan, and Li]{xu2022high}
Yunyang Xu, Xifeng Gao, Caiming Zhang, Jianchao Tan, and Xuemei Li.
\newblock High quality superpixel generation through regional decomposition.
\newblock \emph{IEEE TCSVT}, 33\penalty0 (4):\penalty0 1802--1815, 2022.

\bibitem[Yang et~al.(2020)Yang, Sun, Jin, and Zhou]{yang2020superpixel}
Fengting Yang, Qian Sun, Hailin Jin, and Zihan Zhou.
\newblock Superpixel segmentation with fully convolutional networks.
\newblock In \emph{CVPR}, 2020.

\bibitem[Yao et~al.(2015)Yao, Boben, Fidler, and Urtasun]{yao2015}
Jian Yao, Marko Boben, Sanja Fidler, and Raquel Urtasun.
\newblock Real-time coarse-to-fine topologically preserving segmentation.
\newblock In \emph{CVPR}, 2015.

\bibitem[Yuan et~al.(2021)Yuan, Zhang, Yu, and Zhu]{yuan2021superpixels}
Ye Yuan, Wei Zhang, Hai Yu, and Zhiliang Zhu.
\newblock Superpixels with content-adaptive criteria.
\newblock \emph{IEEE TIP}, 30:\penalty0 7702--7716, 2021.

\bibitem[Zhang et~al.(2016)Zhang, Li, Gao, and Zhang]{zhang2016}
Yongxia Zhang, Xuemei Li, Xifeng Gao, and Caiming Zhang.
\newblock A simple algorithm of superpixel segmentation with boundary
  constraint.
\newblock \emph{IEEE TCSVT}, 2016.

\bibitem[Zhao et~al.(2021)Zhao, Zhou, Bruzzone, Guan, and
  Yang]{zhao2021superpixel}
Haishi Zhao, Fengfeng Zhou, Lorenzo Bruzzone, Renchu Guan, and Chen Yang.
\newblock Superpixel-level global and local similarity graph-based clustering
  for large hyperspectral images.
\newblock \emph{IEEE TGRS}, 60:\penalty0 1--16, 2021.

\bibitem[Zhao et~al.(2023)Zhao, Ding, An, Du, Yu, Li, Tang, and
  Wang]{zhao2023fast}
Xu Zhao, Wenchao Ding, Yongqi An, Yinglong Du, Tao Yu, Min Li, Ming Tang, and
  Jinqiao Wang.
\newblock Fast segment anything.
\newblock \emph{arXiv:2306.12156}, 2023.

\bibitem[Zhou et~al.(2023)Zhou, Kang, and Ming]{zhou2023vine}
Pei Zhou, Xuejing Kang, and Anlong Ming.
\newblock Vine spread for superpixel segmentation.
\newblock \emph{IEEE TIP}, 32:\penalty0 878--891, 2023.

\bibitem[Zhu et~al.(2021)Zhu, She, Zhang, Lu, Lu, Li, and Hu]{zhu2021learning}
Lei Zhu, Qi She, Bin Zhang, Yanye Lu, Zhilin Lu, Duo Li, and Jie Hu.
\newblock Learning the superpixel in a non-iterative and lifelong manner.
\newblock In \emph{CVPR}, 2021.

\end{thebibliography}
